\documentclass{article}

\usepackage{microtype}
\usepackage{graphicx}
\usepackage[footnotesize]{subfigure}
\usepackage{booktabs} 

\usepackage[final]{neurips_2019}

\usepackage[utf8]{inputenc} 
\usepackage{url}
\usepackage{amsfonts}       
\usepackage{nicefrac}       
\usepackage{microtype}      
\usepackage{amsmath,amssymb}
\usepackage{mathtools}
\usepackage{bm}
\usepackage{paralist}

\usepackage{color}
\usepackage{float}
\usepackage{graphicx} 
\graphicspath{{figures/}}
\usepackage[compact]{titlesec}

\titlespacing{\section}{0pt}{2ex}{1ex}
\titlespacing{\subsection}{0pt}{1ex}{0ex}
\titlespacing{\subsubsection}{0pt}{0.5ex}{0ex}

\usepackage{enumitem}
\setlist[itemize]{leftmargin=15pt}%
\setlist[enumerate]{leftmargin=15pt}%

\usepackage{amsmath,amsfonts,bm}

\def\eqref#1{equation~\ref{#1}}

\def\1{\bm{1}}

\def\vtheta{{\bm{\theta}}}

\def\vx{{\bm{x}}}

\DeclareMathAlphabet{\mathsfit}{\encodingdefault}{\sfdefault}{m}{sl}
\SetMathAlphabet{\mathsfit}{bold}{\encodingdefault}{\sfdefault}{bx}{n}

\newcommand{\myvspace}[1]{\vspace{#1}} 
\renewcommand{\myvspace}[1]{} 

\newcommand{\D}{\mathcal{D}}

\newcommand{\Reals}{\mathbb{R}}

\newcommand\iid{i.i.d.}

\definecolor{mydarkblue}{rgb}{0,0.08,0.45}
\usepackage[colorlinks,citecolor=mydarkblue,urlcolor=mydarkblue,linkcolor=mydarkblue,filecolor=mydarkblue]{hyperref} 

\title{
Can You Trust Your Model's Uncertainty? Evaluating Predictive Uncertainty Under Dataset Shift
}

\author{%
 Yaniv Ovadia\thanks{Equal contribution}\\
  Google Research \\
  \texttt{yovadia@google.com} 
  \And
  Emily Fertig\footnotemark[1]\, \thanks{AI Resident} \\
  Google Research \\
  \texttt{emilyaf@google.com}
  \And
     Jie Ren\footnotemark[2]\\
  Google Research \\
  \texttt{jjren@google.com} 
  \And
 Zachary Nado \\
  Google Research \\
  \texttt{znado@google.com} 
  \And
 D Sculley \\
  Google Research \\
  \texttt{dsculley@google.com} 
  \And
  Sebastian Nowozin  \\
  Google Research  \\
  \texttt{nowozin@google.com} \\
  \And
  Joshua V. Dillon \\
  Google Research  \\
  \texttt{jvdillon@google.com} \\
  \And
  Balaji Lakshminarayanan\thanks{Corresponding authors}
  \\
  DeepMind  \\
  \texttt{balajiln@google.com} \\
    \And
  Jasper Snoek\footnotemark[3] \\
  Google Research \\
  \texttt{jsnoek@google.com}
}

\begin{document}
\maketitle

\vskip 0.1in

\begin{abstract}
Modern machine learning methods including deep learning have achieved great success in predictive accuracy for supervised learning tasks, but may still fall short in giving useful estimates of their predictive {\em uncertainty}. Quantifying uncertainty is especially critical in real-world settings, which often involve input distributions that are shifted from the training distribution due to a variety of factors including sample bias and non-stationarity.  In such settings, well calibrated uncertainty estimates convey information about when a model's output should (or should not) be trusted.  Many probabilistic deep learning methods, including Bayesian-and non-Bayesian methods, have been proposed in the literature for quantifying predictive uncertainty, but to our knowledge there has not previously been a rigorous large-scale empirical comparison of these methods under dataset shift. We present a large-scale benchmark of existing state-of-the-art methods on classification problems and investigate the effect of dataset shift on accuracy and calibration.  We find that traditional post-hoc calibration does indeed fall short, as do several other previous methods.  However, some methods that marginalize over models give surprisingly strong results across a broad spectrum of tasks.
\end{abstract}

\section{Introduction}

Recent successes across a variety of domains have led to the widespread deployment of deep neural networks (DNNs) in practice.  Consequently, the predictive distributions of these models are increasingly being used to make decisions in important applications ranging from machine-learning aided medical diagnoses from imaging~\citep{esteva2017} to self-driving cars~\citep{bojarski16}.  Such high-stakes applications require not only point predictions but also accurate quantification of predictive uncertainty, i.e.\  meaningful confidence values in addition to class predictions. With sufficient independent labeled samples from a target data distribution, one can estimate how well a model's confidence aligns with its accuracy and adjust the predictions accordingly.  However, in practice, once a model is deployed the distribution over observed data may shift and eventually be very different from the original training data distribution.  Consider, e.g., online services for which the data distribution may change with the time of day, seasonality or popular trends. Indeed, robustness under conditions of distributional shift and out-of-distribution (OOD) inputs is necessary for the safe deployment of machine learning~\citep{amodei2016concrete}.  For such settings, calibrated predictive uncertainty is important because it enables accurate assessment of risk, allows practitioners to know how accuracy may degrade, and allows a system to abstain from decisions due to low confidence.  

A variety of methods have been developed for quantifying predictive uncertainty in DNNs. Probabilistic neural networks such as mixture density networks \citep{mackay1999density} capture the inherent ambiguity in outputs for a given input, also referred to as \emph{aleatoric uncertainty} \citep{kendall2017uncertainties}. Bayesian neural networks learn a posterior distribution over parameters that quantifies parameter uncertainty, a type of \emph{epistemic uncertainty} that can be reduced through the collection of additional data. Popular approximate Bayesian approaches include Laplace approximation  \citep{mackay1992bayesian}, variational inference \citep{graves,BBB}, dropout-based  variational inference \citep{gal,Kingma15}, expectation propagation~\cite{hernandez2015probabilistic} and stochastic gradient MCMC \citep{Welling2011}. Non-Bayesian methods include training multiple probabilistic neural networks with bootstrap or ensembling~\citep{bootstrapdqn,deepensembles}. 
Another popular non-Bayesian approach involves re-calibration of probabilities on a held-out validation set through temperature scaling \citep{platt99}, which was shown by \citet{guo2017calibration} to lead to well-calibrated predictions on the i.i.d.~test set.

\textbf{Using Distributional Shift to Evaluate Predictive Uncertainty}
While previous work has evaluated the quality of predictive uncertainty on OOD inputs \citep{deepensembles},
there has not to our knowledge been a comprehensive evaluation of uncertainty estimates from different methods under dataset shift.  Indeed, we suggest that effective evaluation of predictive uncertainty is most meaningful under conditions of distributional shift.  One reason for this is that post-hoc calibration gives good results in independent and identically distributed (\iid) regimes, but can fail under even a mild shift in the input data.  And in real world applications, as described above, distributional shift is widely prevalent.  Understanding questions of risk, uncertainty, and trust in a model's output becomes increasingly critical as shift from the original training data grows larger.

\textbf{Contributions}  In the spirit of calls for more rigorous understanding of existing methods \citep{lipton2018troubling,sculley2018winner,rahimi2017addendum}, this paper provides a benchmark for evaluating uncertainty that focuses not only on the i.i.d.~setting but also \emph{uncertainty under distributional shift.}  We present a large-scale evaluation of popular approaches in probabilistic deep learning, focusing on methods that operate well in large-scale settings, and evaluate them on a diverse range of classification benchmarks across image, text, and categorical modalities. We use these experiments to evaluate the following questions:
\begin{itemize}
    \item How trustworthy are the uncertainty estimates of different methods under dataset shift?
    \item Does calibration in the i.i.d.~setting translate to calibration under dataset shift? 
    \item How do uncertainty and accuracy of different methods co-vary under dataset shift?  Are there methods that consistently do well in this regime?
\end{itemize}

In addition to answering the questions above, our code is made available open-source along with our model predictions such that researchers can easily evaluate their approaches on these benchmarks~\footnote{\url{https://github.com/google-research/google-research/tree/master/uq_benchmark_2019}}.

\section{Background} 
\textbf{Notation and Problem Setup} 
Let $\vx\in\Reals^d$ represent a set of $d$-dimensional features and $y\in\{1,\ldots,k\}$ denote corresponding labels (targets) for $k$-class classification.   
We assume that a training dataset $\D$ consists of $N$ \iid samples $\D=\{(\vx_n,y_n)\}_{n=1}^N$. 

Let $p^*(\vx,y)$ denote the true distribution (unknown, observed only through the samples $\D$), also referred to as the \emph{data generating process}. 
We focus on classification problems, in which the true distribution is assumed to be a discrete distribution over $k$ classes, and the observed $y\in\{1,\ldots,k\}$ is a sample from the conditional distribution $p^*(y|\vx)$. 
We use a neural network to model $p_\vtheta(y|\vx)$ and estimate the parameters $\vtheta$ using the training dataset. 
At test time, we evaluate the model predictions against a test set, sampled from the same distribution as the training dataset.
However, here we also evaluate the model against OOD inputs sampled from $q(\vx,y) \neq p^*(\vx,y) $.
In particular, we consider two kinds of shifts:
\begin{itemize}[leftmargin=*]
    \item \emph{shifted versions} of the test inputs where the ground truth label belongs to one of the $k$ classes. We use shifts such as corruptions and perturbations proposed by \citet{hendrycks2018benchmarking}, and ideally would like the model predictions to become more uncertain with increased shift, assuming shift degrades accuracy.  This is also referred to as \emph{covariate shift}~\citep{sugiyama2017dataset}.
    \item \emph{a completely different OOD dataset}, where the ground truth label is not one of the $k$ classes. Here we check if the model exhibits higher predictive uncertainty for those new instances and to this end report diagnostics that rely only on predictions and not ground truth labels.
\end{itemize}

\textbf{High-level overview of existing methods} A large variety of methods have been developed to either provide higher quality uncertainty estimates or perform OOD detection to inform model confidence.  These can roughly be divided into:
\begin{enumerate}[leftmargin=*]{
\item Methods which deal with $p(y|\vx)$ only, we discuss these in more detail in Section~\ref{sec:methods}.
\item Methods which model the joint distribution $p(y, \vx)$, e.g. deep hybrid models~\citep{kingma2014semi,uncertaintyvib,nalisnickhybrid,behrmann2018invertible}.
\item Methods with an OOD-detection component in addition to $p(y|\vx)$ \citep{bishop1994novelty,lee2018simple,odin}, and related work on selective classification \citep{geifman2017selective}.
}\end{enumerate}
 We refer to \citet{shafaei2018does} for a recent summary of these methods.
Due to the differences in modeling assumptions, a fair comparison between these different classes of methods is challenging; for instance, some OOD detection methods rely on knowledge of a known OOD set, or train using a none-of-the-above class, and it may not always be meaningful to compare predictions from these methods with those obtained from a Bayesian DNN. We focus on methods described by (1) above, as this allows us to focus on methods which make the same modeling assumptions about data and differ only in how they quantify predictive uncertainty.

\section{Methods and Metrics}\label{sec:methods}
We select a subset of methods from the probabilistic deep learning literature for their prevalence, scalability and practical applicability\footnote{The methods used scale well for training and prediction (see \textbf{in Appendix~\ref{sec:costs}.)}.  We also explored methods such as scalable extensions of Gaussian Processes~\citep{hensman2015scalable}, but they were challenging to train on the 37M example Criteo dataset or the 1000 classes of ImageNet.}.  These include (see also references within):
\begin{itemize}[leftmargin=*]

\item (\emph{Vanilla}) Maximum softmax probability \citep{hendrycks2016baseline} 
\item (\emph{Temp Scaling}) Post-hoc calibration by temperature scaling using a validation set \citep{guo2017calibration} 
\item (\emph{Dropout})  Monte-Carlo Dropout \citep{gal,srivastava2015training} with rate $p$ 
\item (\emph{Ensembles}) Ensembles of $M$ networks trained independently on the entire dataset using random initialization \citep{deepensembles} (we set $M=10$ in experiments below)
\item (\emph{SVI}) Stochastic Variational Bayesian Inference for deep learning
\citep{BBB,graves,louizos2017multiplicative,louizos2016structured,flipout}. We refer to Appendix~\ref{sec:svi} for details of our SVI implementation. 
\item (LL) Approx. Bayesian inference for the parameters of the last layer only \citep{riquelme2018deep}
\begin{itemize}

\item (\emph{LL SVI}) Mean field stochastic variational inference on the last layer only
\item (\emph{LL Dropout}) Dropout only on the activations before the last layer 
\end{itemize}
\end{itemize}

In addition to metrics (we use arrows to indicate which direction is better) that do not depend on predictive uncertainty, such as classification accuracy $\uparrow$,
the following metrics are commonly used:

\textbf{Negative Log-Likelihood (NLL) $\downarrow$}
Commonly used to evaluate the quality of model uncertainty on some held out set.
\textit{Drawbacks:} Although a proper scoring rule~\citep{gneiting2007strictly}, it can over-emphasize tail probabilities \citep{quinonero2006evaluating}. 

\textbf{Brier Score $\downarrow$}~\citep{brier1950verification}
Proper scoring rule for measuring the accuracy of predicted probabilities. It is computed as the squared error of a predicted probability \textit{vector}, $p(y|x_n, \vtheta)$, and the one-hot encoded true response, $y_n$. That is,
\begin{equation}
\mathrm{BS}
= |\mathcal{Y}|^{-1} \sum_{y\in\mathcal{Y}} (p(y|\vx_n,\vtheta) - \delta(y-y_n))^2
= |\mathcal{Y}|^{-1}\Big(1- 2 p(y_n|\vx_n,\vtheta)+\sum_{y\in\mathcal{Y}} p(y|\vx_n,\vtheta)^2\Big).
\end{equation}%
The Brier score has a convenient interpretation as $BS = \mathrm{uncertainty} - \mathrm{resolution} + \mathrm{reliability}$, where
$\mathrm{uncertainty}$ is the marginal uncertainty over labels,
$\mathrm{resolution}$ measures the deviation of individual predictions against the marginal, and
$\mathrm{reliability}$ measures calibration as the average violation of long-term true label frequencies.
We refer to \citet{degroot1983comparison} for the decomposition of Brier score into calibration and refinement for classification and to~\citep{brocker2009reliability} for the general decomposition for any proper scoring rule. 
\textit{Drawbacks:} Brier score is insensitive to predicted probabilities associated with in/frequent events.

Both the Brier score and the negative log-likelihood are proper scoring rules and therefore the optimum score corresponds to a perfect prediction.
In addition to these two metrics, we also evaluate two metrics---\emph{expected calibration error} and \emph{entropy}.
Neither of these is a proper scoring rule, and thus there exist trivial solutions which yield optimal scores; for example, returning the marginal probability $p(y)$ for every instance will yield perfectly calibrated but uninformative predictions.  Each proper scoring rule induces a calibration measure~\citep{brocker2009reliability}.  However, ECE is not the result of such decomposition and has no corresponding proper scoring rule; we instead include ECE because it is popularly used and intuitive. Each proper scoring rule is also associated with a corresponding entropy function and Shannon entropy is that for log probability~\citep{gneiting2007strictly}.

\textbf{Expected Calibration Error (ECE) $\downarrow$}
Measures the correspondence between predicted probabilities and empirical accuracy~\citep{naeini2015obtaining}. It is computed as the average gap between within bucket accuracy and within bucket predicted probability for $S$ buckets $B_s = \{n \in 1\ldots N : p(y_n|\vx_n,\vtheta) \in (\rho_s, \rho_{s+1}]\}$. That is, $
    \mathrm{ECE} = \sum_{s=1}^S \frac{|B_s|}{N} |\operatorname{acc}(B_s) - \operatorname{conf}(B_s)|,$
where $\operatorname{acc}(B_s) = |B_s|^{-1}\sum_{n\in B_s} [y_n = \hat{y}_n]$, $\operatorname{conf}(B_s)=|B_s|^{-1}\sum_{n\in B_s} p(\hat{y}_n|\vx_n,\vtheta)$, and $\hat{y}_n=\arg\max_y p(y|\vx_n,\vtheta)$ is the $n$-th prediction. When bins $\{\rho_s : s\in 1\ldots S\}$ are quantiles of the held-out predicted probabilities, $|B_s|\approx|B_k|$ and the estimation error is approximately constant. \textit{Drawbacks:} Due to binning, ECE does not monotonically increase as predictions approach ground truth. If $|B_s|\ne |B_k|$, the estimation error varies across bins.%

There is no ground truth label for fully OOD inputs. Thus we report histograms of \textbf{confidence} and predictive \textbf{entropy} on known and OOD inputs and \textbf{accuracy versus confidence plots}~\citep{deepensembles}:
Given the prediction $p(y=k|\vx_n,\vtheta)$, we define the predicted label as $\hat{y}_n=\arg\max_y p(y|\vx_n,\vtheta)$, and the confidence as $p(y=\hat{y}|\vx,\vtheta) = \max_k p(y=k|\vx_n,\vtheta)$. We filter out test examples corresponding to a particular confidence threshold $\tau\in[0, 1]$ and compute the accuracy on this set.

\section{Experiments and Results}
We evaluate the behavior of the predictive uncertainty of deep learning models on a variety of datasets across three different modalities: images, text and categorical (online ad) data. For each we follow standard training, validation and testing protocols, but we additionally evaluate results on increasingly shifted data and an OOD dataset.  We detail the models and implementations used in Appendix~\ref{sec:model_details}.  Hyperparameters were tuned for all methods using Bayesian optimization~\citep{vizier} (except on ImageNet) as detailed in Appendix~\ref{sec:hyperparameter_tuning}.%
\subsection{An illustrative example - MNIST}%
\begin{figure}[h]
    \centering%
    \begin{subfigure}[Rotated MNIST ]{
      \includegraphics[width=0.33\linewidth]{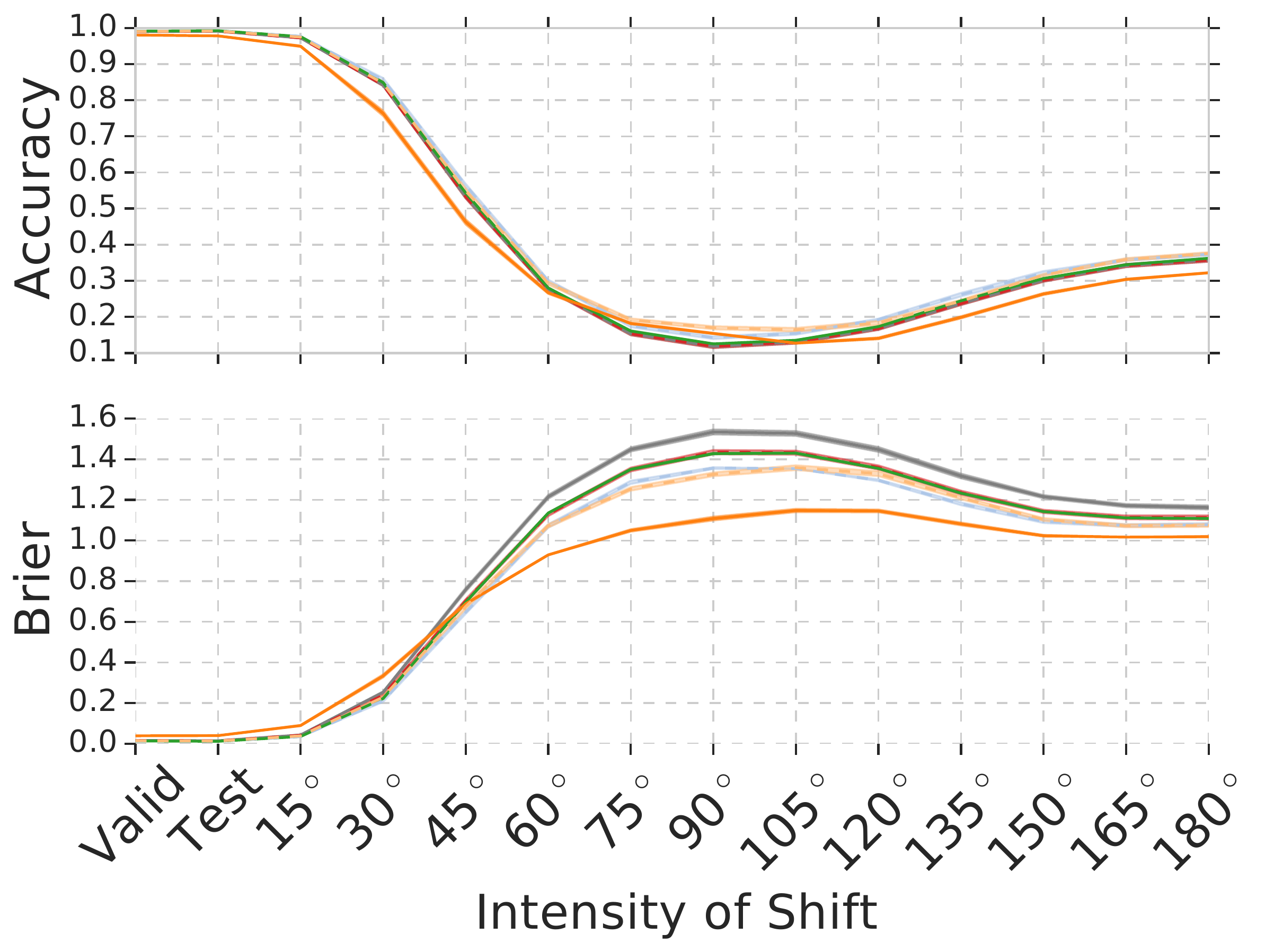}%
        \label{fig:mnist_rotations}%
    }\end{subfigure}%
    \begin{subfigure}[Translated MNIST]{%
      \includegraphics[width=0.33\linewidth]{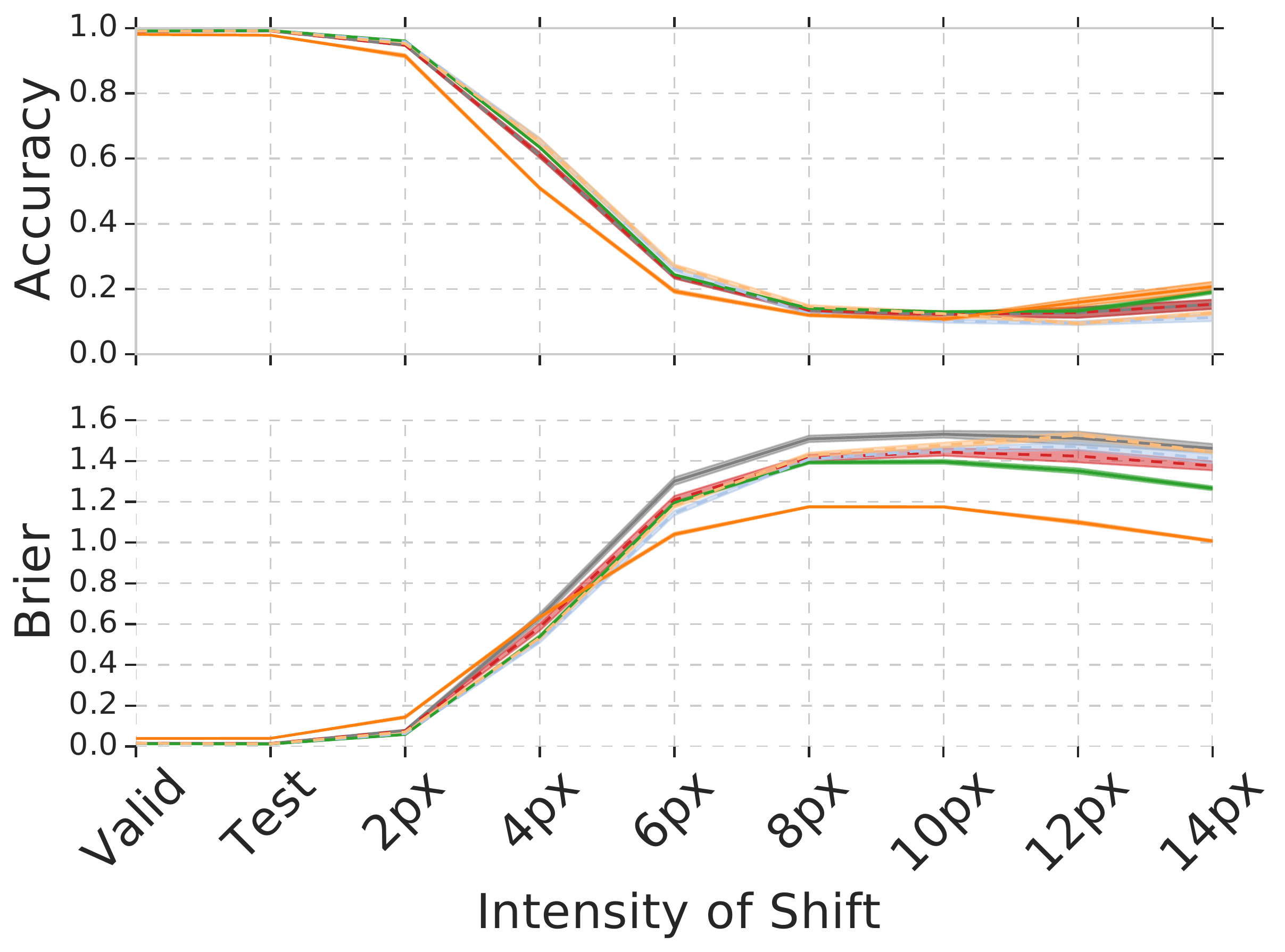}%
        \label{fig:mnist_rolls}}%
    \end{subfigure}%
    \begin{subfigure}[Confidence vs Acc Rotated 60$^\circ$]{
      \includegraphics[width=0.33\linewidth]{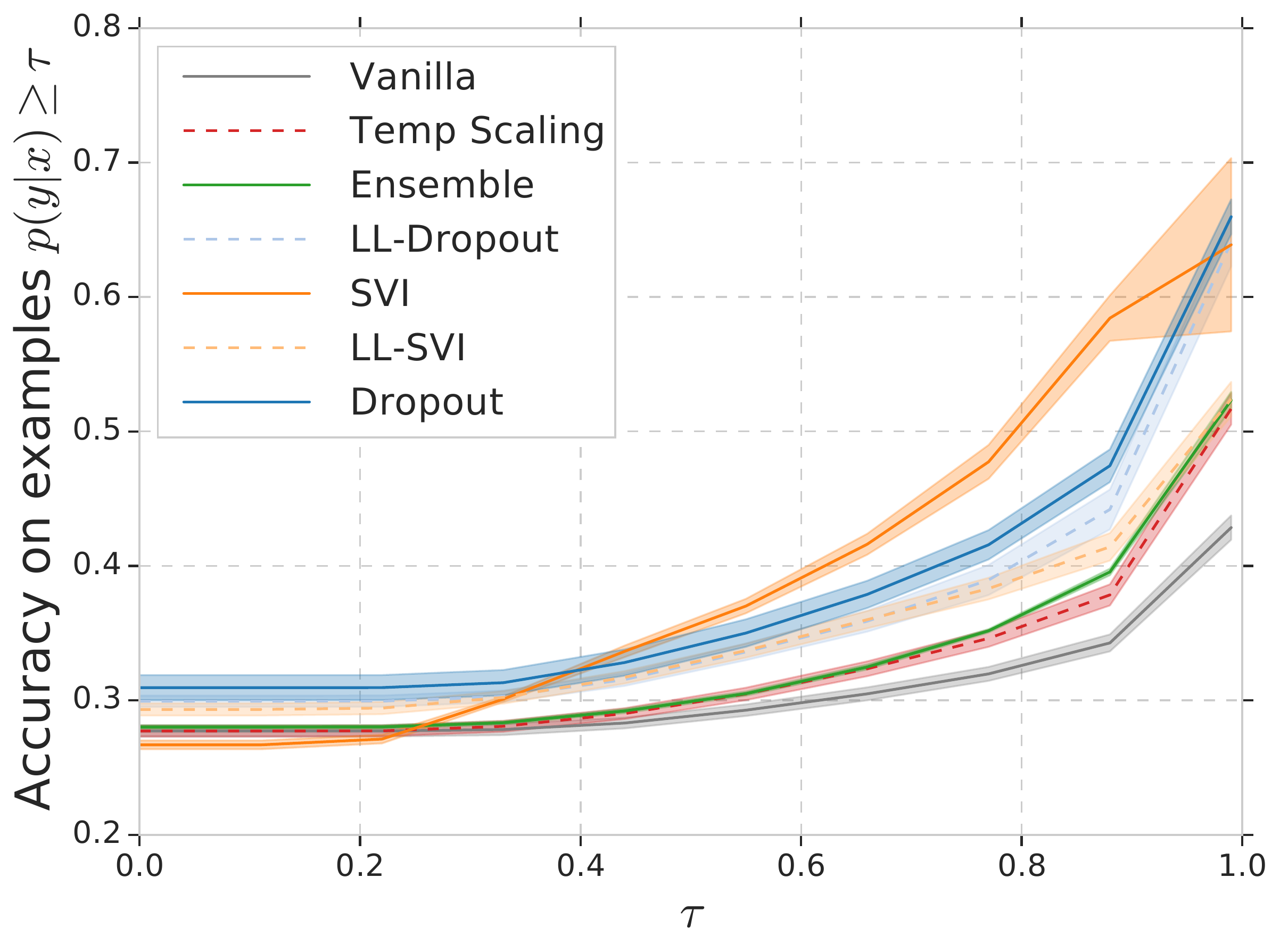}%
        \label{fig:mnist_conf_vs_accuracy_rot60}}%
    \end{subfigure}\\%
    \begin{subfigure}[Count vs Confidence Rotated 60$^\circ$]{
      \includegraphics[width=0.33\linewidth]{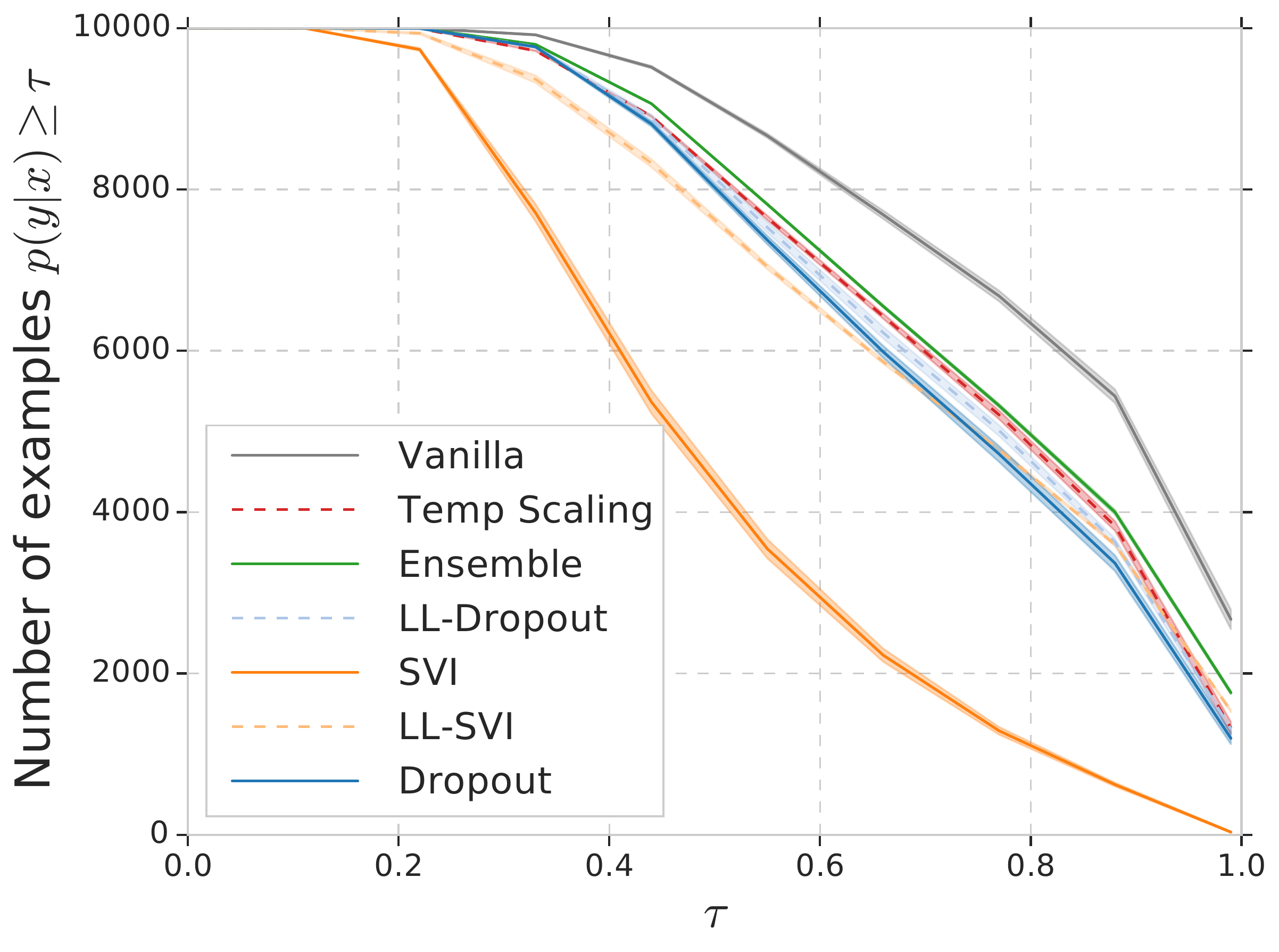}%
        \label{fig:mnist_count_vs_confidence_rot60}}%
    \end{subfigure}%
         \begin{subfigure}[Entropy on OOD]{
      \includegraphics[width=0.33\linewidth]{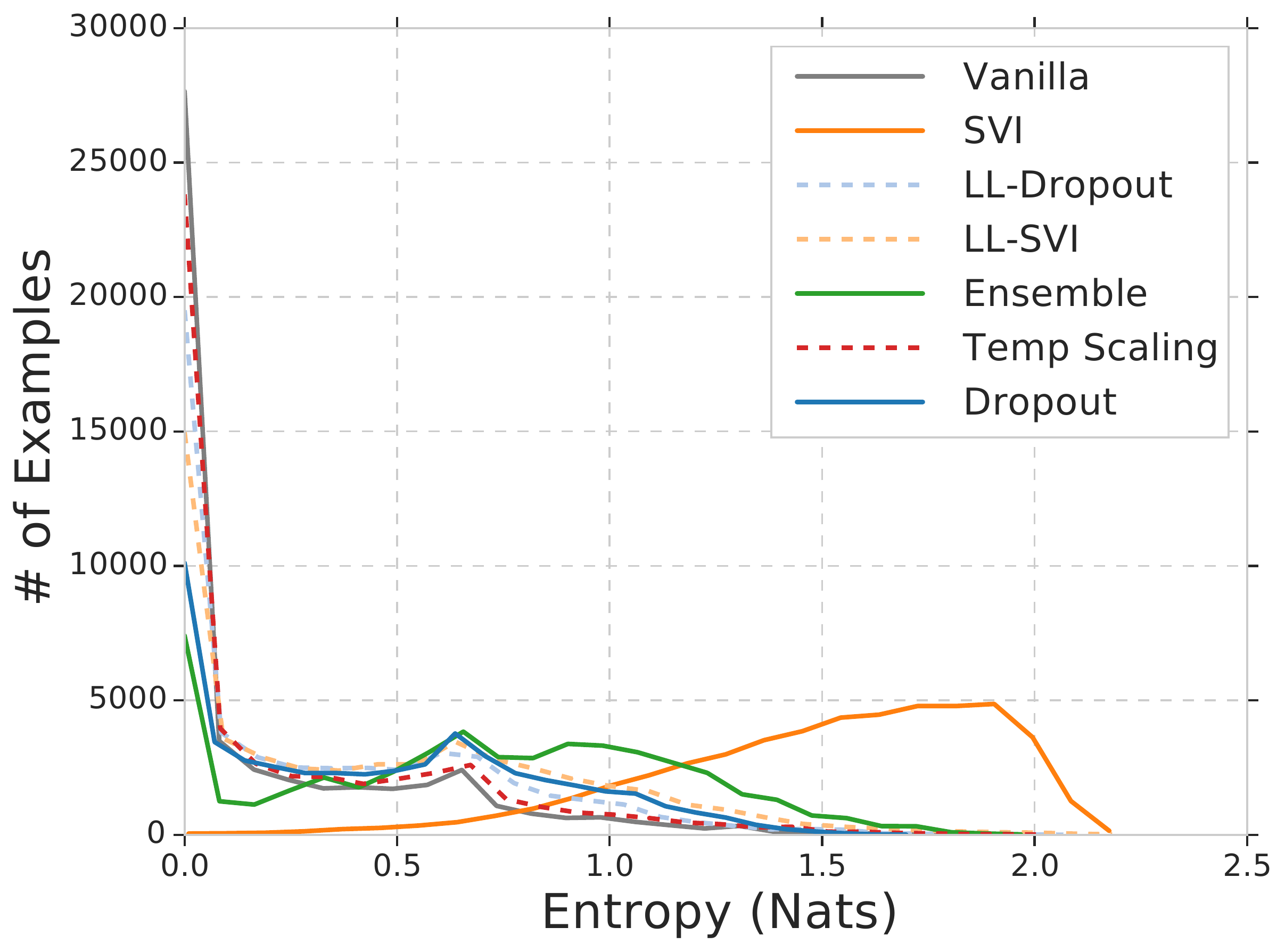}%
        \label{fig:MNIST_entropy_fashion}%
    }\end{subfigure}%
     \begin{subfigure}[Confidence on OOD]{
      \includegraphics[width=0.33\linewidth]{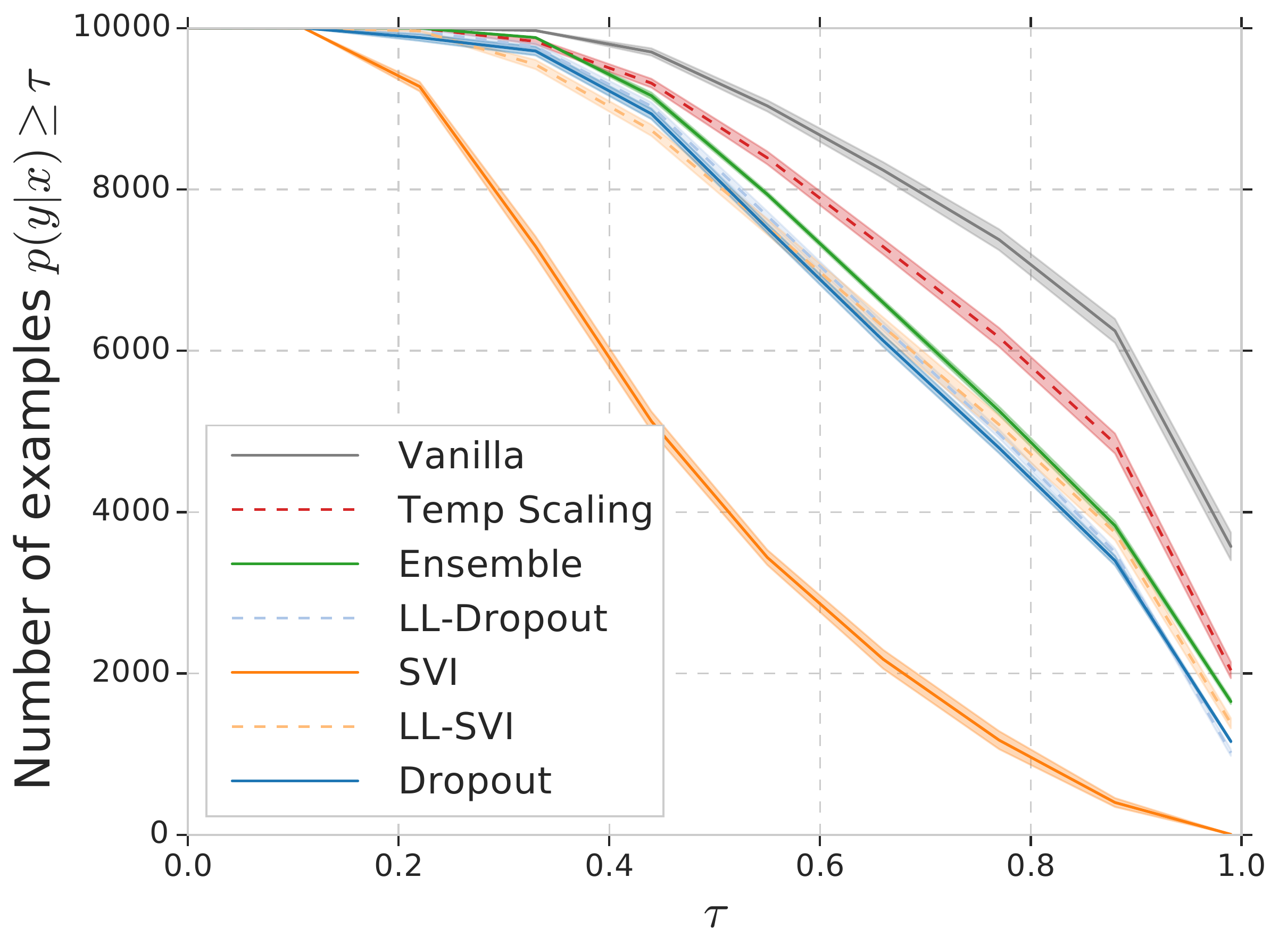}%
        \label{fig:MNIST_confidence_fashion}%
    }\end{subfigure}%
    \myvspace{-0.5em}%
    \caption{Results on MNIST:
       \ref{fig:mnist_rotations} and \ref{fig:mnist_rolls} show accuracy and Brier score as the data is increasingly shifted. Shaded regions represent standard error over 10 runs. To understand the discrepancy between accuracy and Brier score, we explore the predictive distributions of each method by looking at the confidence of the predictions in~\ref{fig:mnist_conf_vs_accuracy_rot60} and~\ref{fig:mnist_count_vs_confidence_rot60}.  We also explore the entropy and confidence of each method on entirely OOD data in \ref{fig:MNIST_entropy_fashion} and~\ref{fig:MNIST_confidence_fashion}. SVI has lower accuracy on the validation and test splits, but it is significantly more robust to dataset shift as evidenced by a lower Brier score, lower overall confidence~\ref{fig:mnist_count_vs_confidence_rot60} and higher predictive entropy under shift~(\ref{fig:mnist_conf_vs_accuracy_rot60}) and OOD data~(\ref{fig:MNIST_entropy_fashion},\ref{fig:MNIST_confidence_fashion}).}%
    \label{fig:mnist}%
    \myvspace{-2em}%
\end{figure}
\label{sec:mnist}
We first illustrate the problem setup and experiments using the MNIST dataset.  We used the LeNet~\citep{lecun-98} architecture, and, as with all our experiments, we follow standard training, validation, testing and hyperparameter tuning protocols.  However, we also compute predictions on increasingly shifted data (in this case increasingly rotated or horizontally translated images) and study the behavior of the predictive distributions of the models.  In addition, we predict on a completely OOD dataset, Not-MNIST \citep{notmnist},
and observe the entropy of the model's predictions.  We summarize some of our findings in Figure~\ref{fig:mnist} and discuss below.

\textbf{What we would like to see:} Naturally, we expect the accuracy of a model to degrade as it predicts on increasingly shifted data, and ideally this reduction in accuracy would coincide with increased forecaster entropy.  A model that was well-calibrated on the training and validation distributions would ideally remain so on shifted data. If calibration (ECE or Brier reliability) remained as consistent as possible, practitioners and downstream tasks could take into account that a model is becoming increasingly uncertain. On the completely OOD data, one would expect the predictive distributions to be of high entropy.  Essentially, we would like the predictions to indicate that a model ``knows what it does not know" due to the inputs 
straying away from the training data distribution.

\textbf{What we observe:} We see in Figures~\ref{fig:mnist_rotations} and~\ref{fig:mnist_rolls} that accuracy certainly degrades as a function of shift for all methods tested, and they are difficult to disambiguate on that metric. 
However, the Brier score paints a clearer picture and we see a significant difference between methods, i.e.\ prediction quality degrades more significantly for some methods than others.  An important observation is that \emph{while calibrating on the validation set leads to well-calibrated predictions on the test set, it does not guarantee calibration on shifted data}.  In fact, nearly all other methods (except vanilla) perform better than the state-of-the-art post-hoc calibration (Temperature scaling) in terms of Brier score under shift.  While SVI achieves the worst accuracy on the test set, it actually outperforms all other methods by a much larger margin when exposed to significant shift.  In Figures~\ref{fig:mnist_conf_vs_accuracy_rot60} and~\ref{fig:mnist_count_vs_confidence_rot60} we look at the distribution of confidences for each method to understand the discrepancy between metrics.  We see in Figure~\ref{fig:mnist_count_vs_confidence_rot60} that SVI has the lowest confidence in general but in Figure~\ref{fig:mnist_conf_vs_accuracy_rot60} we observe that SVI gives the highest accuracy at high confidence (or conversely is much less frequently confidently wrong), which can be important for high-stakes applications.  Most methods demonstrate very low entropy~(Figure~\ref{fig:MNIST_entropy_fashion}) and give high confidence predictions (Figure~\ref{fig:MNIST_confidence_fashion}) on data that is entirely OOD, i.e.\ they are confidently wrong about completely OOD data.

\begin{figure}[ht]
    \centering%
       \includegraphics[width=0.95\linewidth]{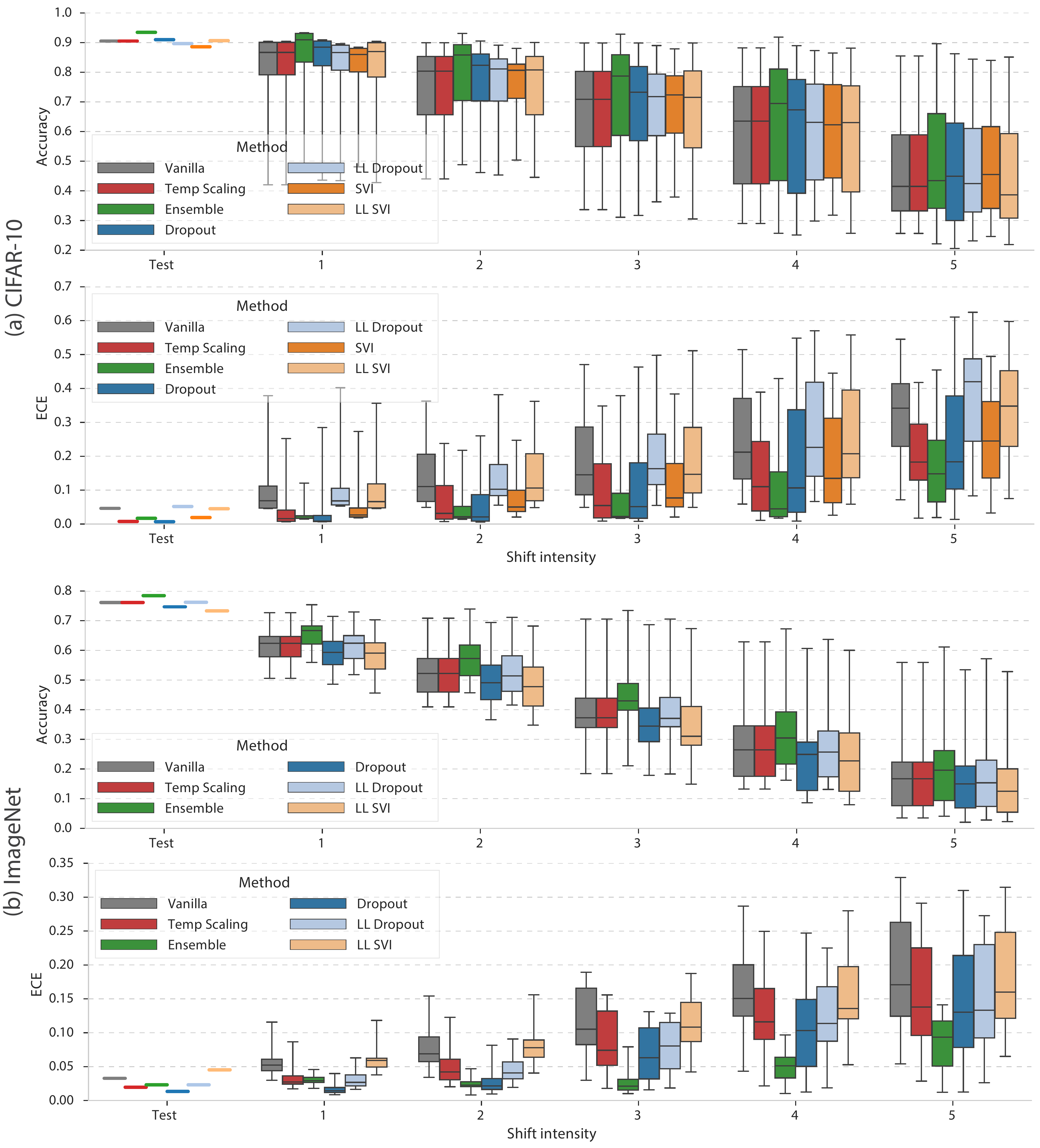} 
    \caption{Calibration under distributional shift: a detailed comparison of accuracy and ECE under all types of corruptions on
  (a) CIFAR-10 and (b) ImageNet.
  For each method we show the mean on the test set and summarize the results on each intensity of shift with a box plot.  Each box shows the quartiles summarizing the results across all (16) types of shift while the error bars indicate the min and max across different shift types.  Figures showing additional metrics are provided in Figures~\ref{fig:cifar_boxplots} (CIFAR-10) and \ref{fig:imagenet_boxplots} (ImageNet).  Tables for numerical comparisons are provided in Appendix~\ref{sec:tables}.}
  \vspace{-1em}
    \label{fig:cifar_and_imagenet_boxplots}%
\end{figure}%

\subsection{Image Models: CIFAR-10 and ImageNet}\label{sec:cifar_imagenet_results}

\begin{figure}[h]%
    \centering%
    \begin{subfigure}[CIFAR: Confidence vs Accuracy]{
      \includegraphics[width=0.33\linewidth]{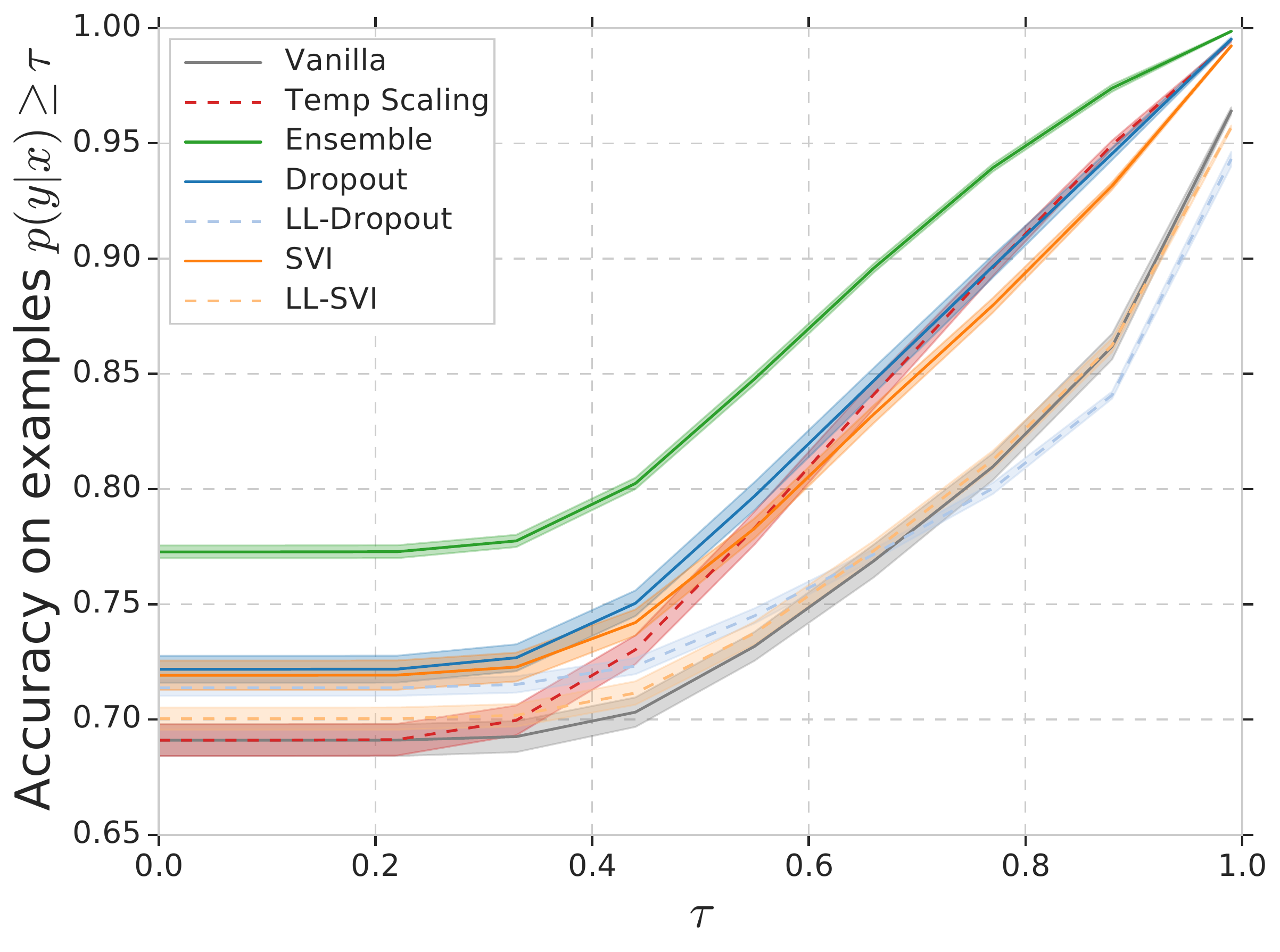}%
      \label{fig:cifar_conf_vs_accuracy}}%
    \end{subfigure}%
    \begin{subfigure}[CIFAR: Count vs Confidence]{
      \includegraphics[width=0.33\linewidth]{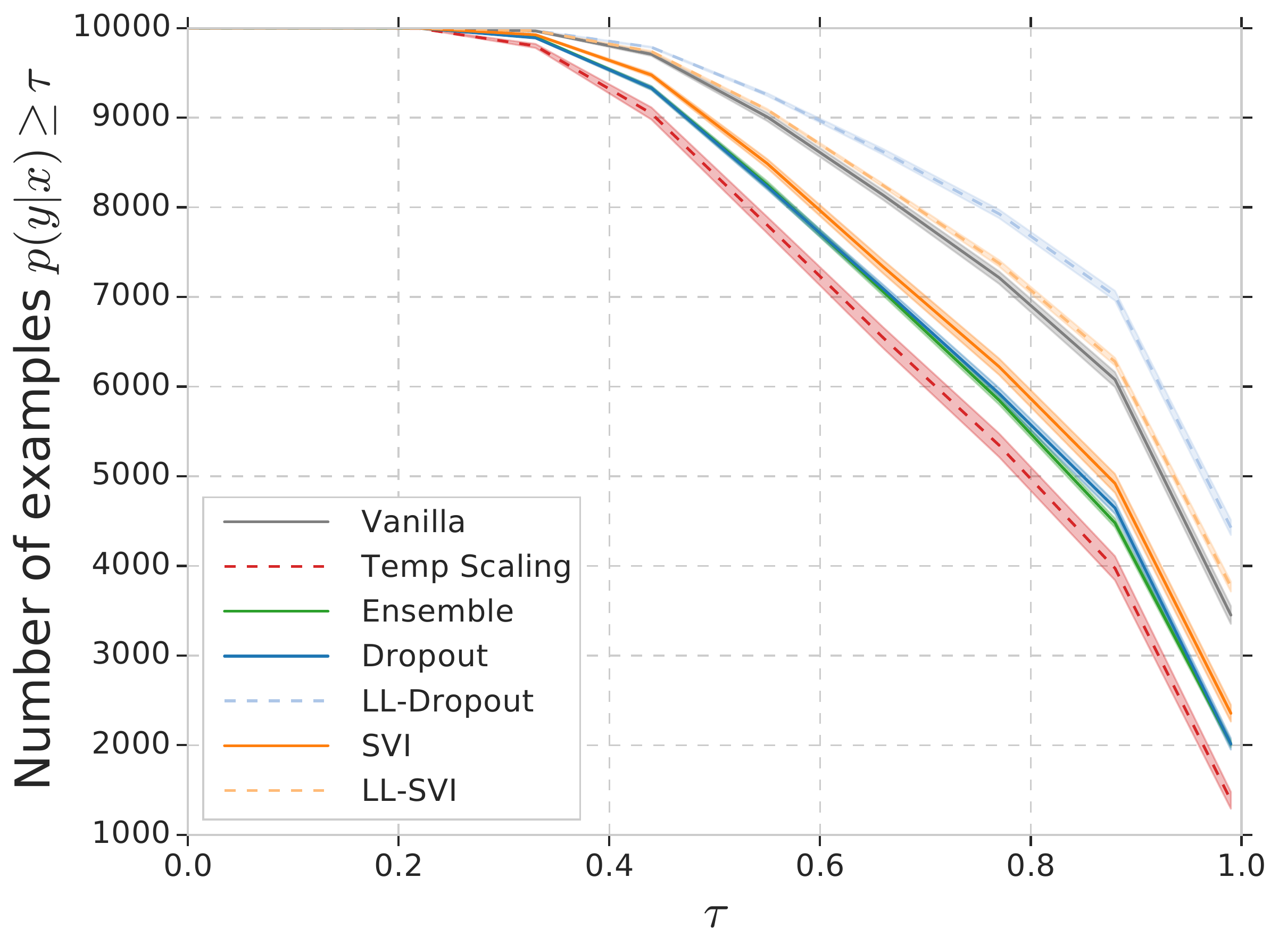}%
      \label{fig:cifar_count_vs_confidence}}%
    \end{subfigure}%
     \begin{subfigure}[CIFAR: Entropy on OOD]{
     \includegraphics[width=0.33\linewidth]{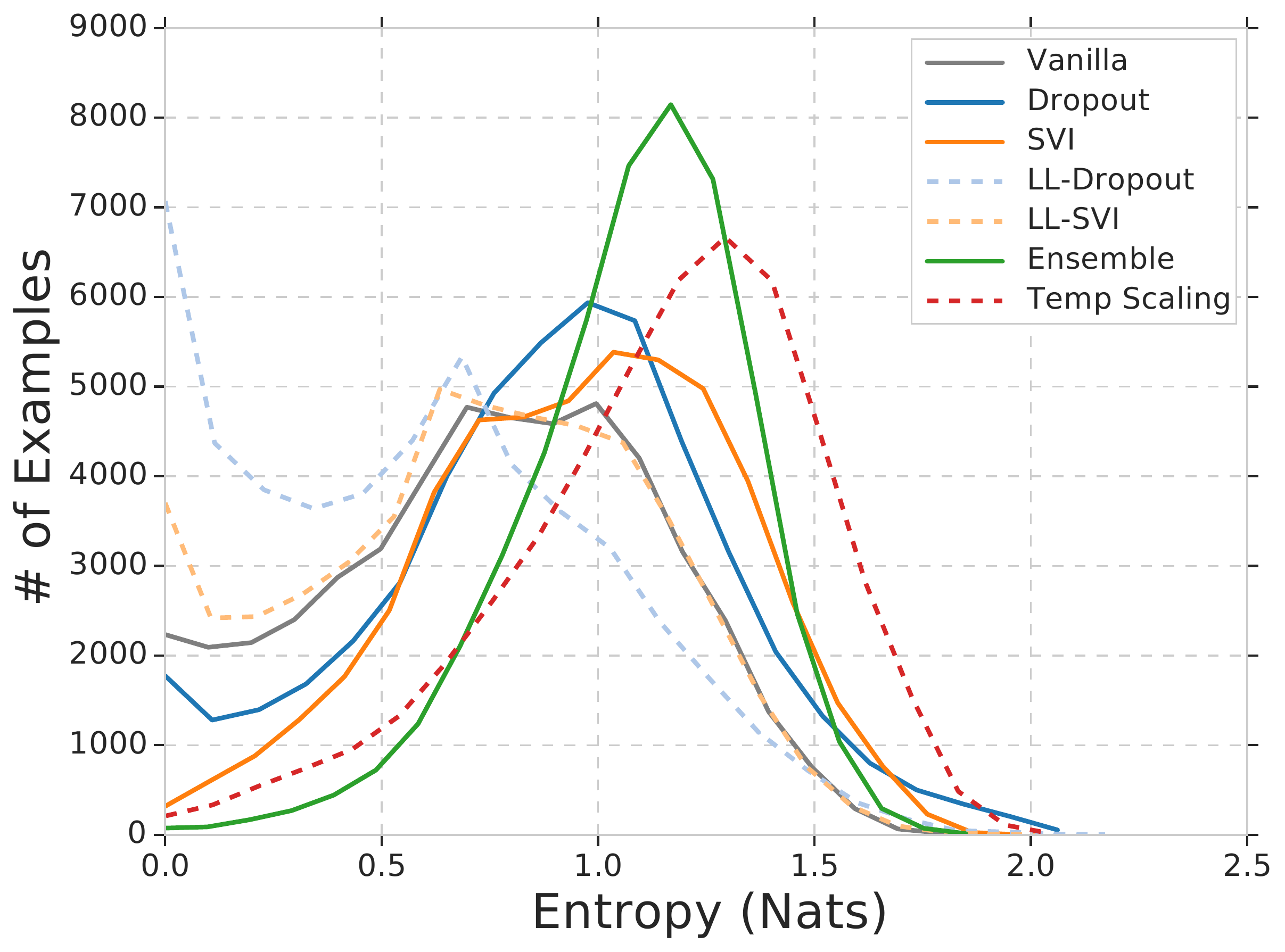}%
     \label{fig:CIFAR_entropy_svhn}
     }\end{subfigure}
     \\
    \begin{subfigure}[ImageNet: Confidence vs Acc]{
    \includegraphics[width=0.33\linewidth]{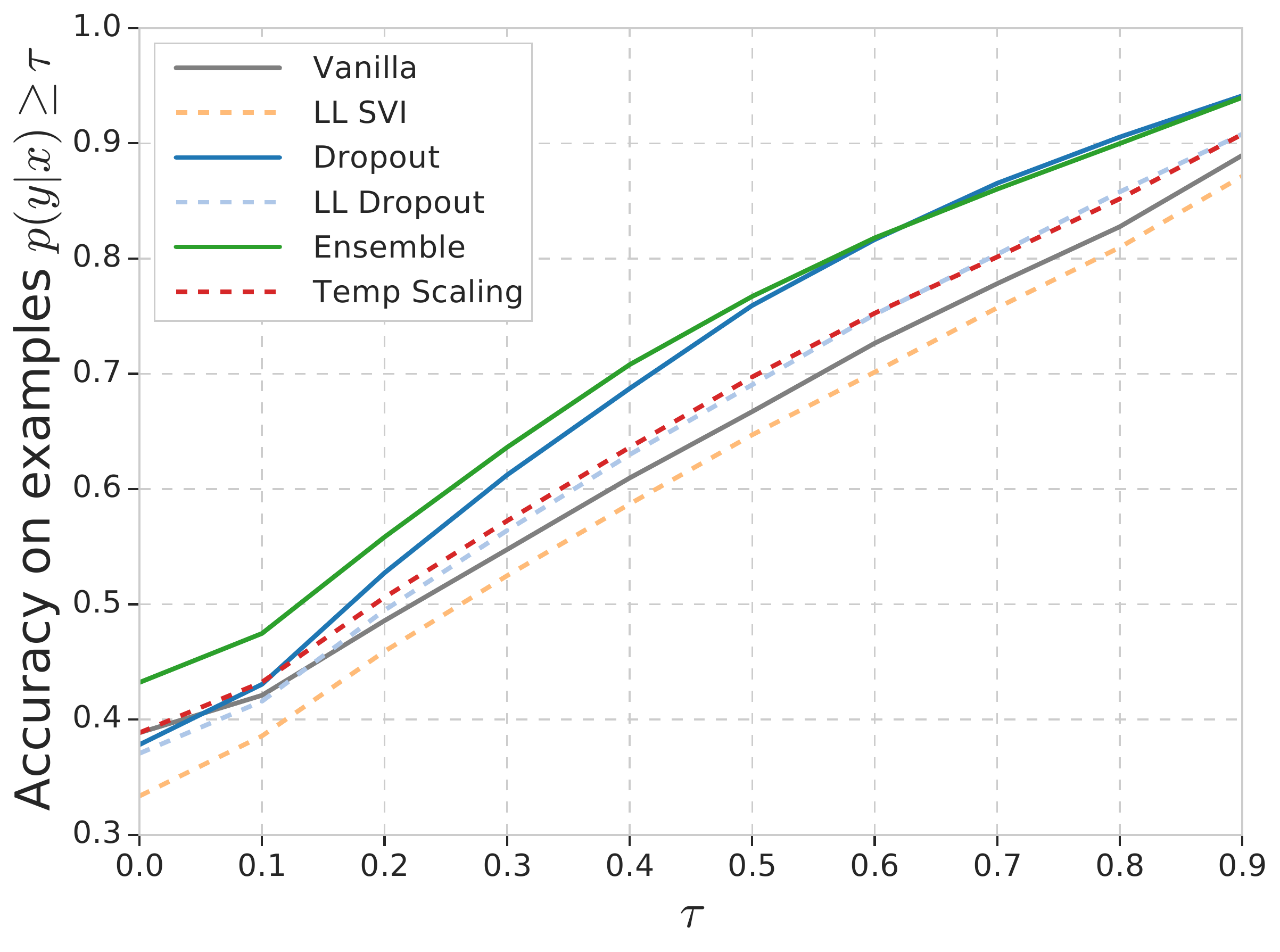}%
    \label{fig:imagenet_conf_vs_accuracy}}%
    \end{subfigure}%
    \begin{subfigure}[ImageNet: Count vs Confidence]{
    \includegraphics[width=0.33\linewidth]{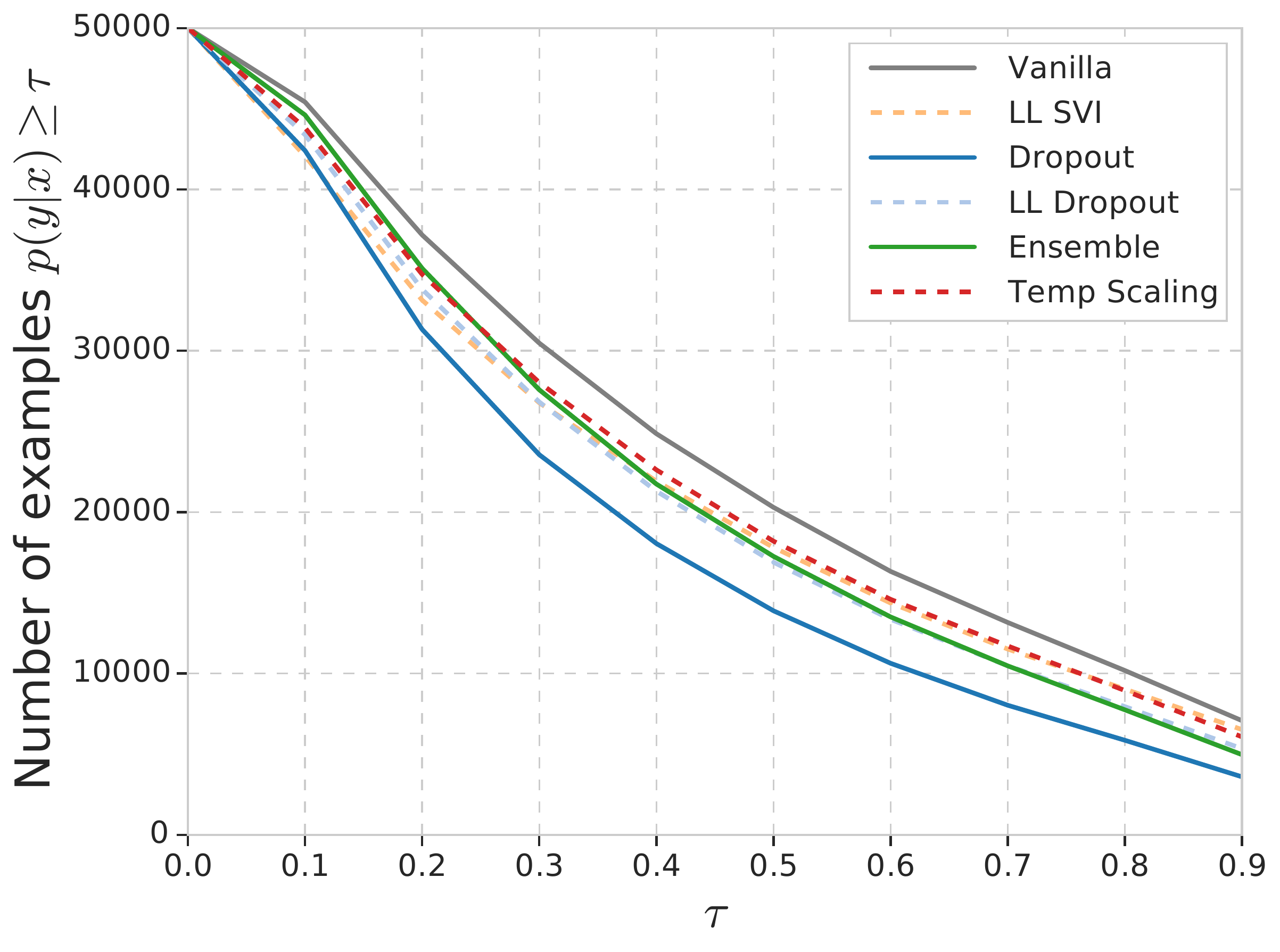}%
    \label{fig:imagenet_count_vs_confidence}}%
    \end{subfigure}%
         \begin{subfigure}[CIFAR: Confidence on OOD]{
       \includegraphics[width=0.33\linewidth]{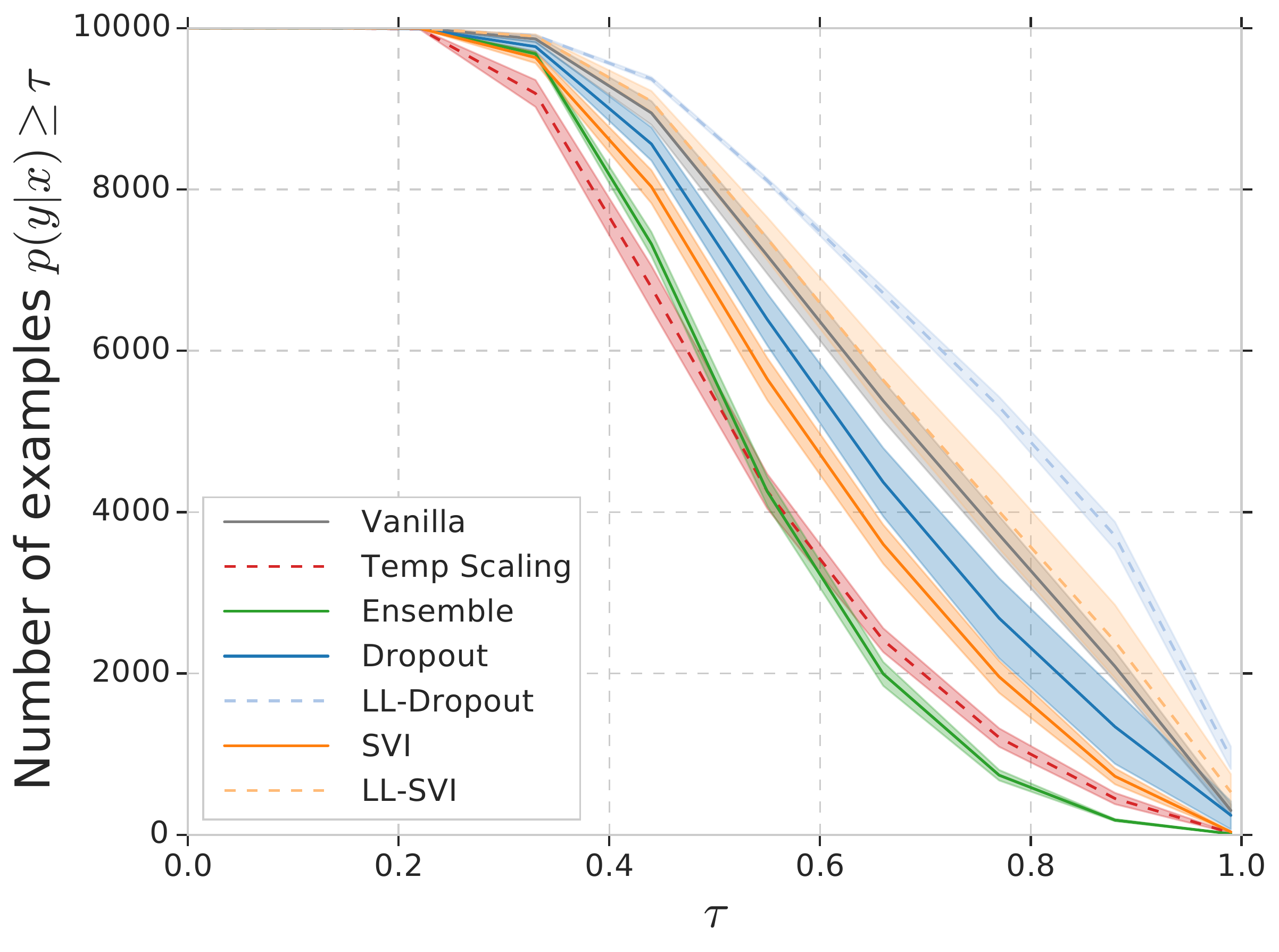}%
       \label{fig:CIFAR_confidence_svhn}%
    }\end{subfigure}
   \myvspace{-0.5em}%
  \caption{
Results on CIFAR-10 and ImageNet. 
Left column: \ref{fig:cifar_conf_vs_accuracy} and \ref{fig:imagenet_conf_vs_accuracy} show accuracy as a function of confidence. Middle column:  \ref{fig:cifar_count_vs_confidence} and \ref{fig:imagenet_count_vs_confidence}
show the number of examples greater than given confidence values for Gaussian blur of intensity 3. 
Right column: \ref{fig:CIFAR_entropy_svhn} and \ref{fig:CIFAR_confidence_svhn} show histogram of entropy and confidences from CIFAR-trained models on a completely different dataset (SVHN).
  }
  \label{fig:cifar_imagenet_conf_accuracy}%
\end{figure}%
We now study the predictive distributions of residual networks~\citep{resnet} trained  on two benchmark image datasets, CIFAR-10~\citep{cifar10} and ImageNet~\citep{imagenet_cvpr09}, under distributional shift.  We use 20-layer and  50-layer ResNets for CIFAR-10 and ImageNet respectively.  For shifted data we use 80 different distortions (16 different types with 5 levels of intensity each, see Appendix~\ref{sec:skew:examples} for illustrations) introduced by~\citet{hendrycks2018benchmarking}.
To evaluate predictions of CIFAR-10 models on entirely OOD data, we use the SVHN dataset \citep{svhn}.

Figure~\ref{fig:cifar_and_imagenet_boxplots} summarizes the accuracy and ECE for CIFAR-10 (top) and ImageNet (bottom) across all 80 combinations of corruptions and intensities from~\citep{hendrycks2018benchmarking}. Figure~\ref{fig:cifar_imagenet_conf_accuracy} inspects the predictive distributions of the models on CIFAR-10 (top) and ImageNet (bottom) for shifted (Gaussian blur) and OOD data.  Classifiers on both datasets show poorer accuracy and calibration with increasing shift.  Comparing accuracy for different methods, we see that ensembles achieve highest accuracy under distributional shift. Comparing the ECE for different methods, we observe that while the methods achieve comparable low values of ECE for small values of shift, ensembles outperform the other methods for larger values of shift. To test whether this result is due simply to the larger aggregate capacity of the ensemble, we trained models with double the number of filters for the Vanilla and Dropout methods. The higher-capacity models showed no better accuracy or calibration for medium- to high-shift than the corresponding lower-capacity models (see Appendix~\ref{sec:calibration:skew:additional}).  In Figures~\ref{fig:cifar_nsamples} and \ref{fig:ensemble_size} we also explore the effect of the number of samples used in dropout, SVI and last layer methods and size of the ensemble, on CIFAR-10. We found that while increasing ensemble size up to 50 did help, most of the gains of ensembling could be achieved with only 5 models.
Interestingly, \emph{while temperature scaling achieves low ECE for low values of shift, the ECE increases significantly as the shift increases, which indicates that calibration on the i.i.d.~validation dataset does not guarantee calibration under distributional shift}. (Note that for ImageNet, we found similar trends considering just the top-5 predicted classes, See Figure~\ref{fig:imagenet_boxplots}.)  Furthermore, the results show that while temperature scaling helps significantly over the vanilla method, ensembles and dropout tend to be better. In Figure~\ref{fig:cifar_imagenet_conf_accuracy}, we see that ensembles and dropout are more accurate at higher confidence.  However, in~\ref{fig:CIFAR_entropy_svhn} we see that temperature scaling gives the highest entropy on OOD data. Ensembles consistently have high accuracy but also high entropy on OOD data.  
We refer to Appendix~\ref{sec:calibration:skew:additional} for additional results; Figures~\ref{fig:cifar_boxplots} and \ref{fig:imagenet_boxplots} report additional metrics on CIFAR-10 and ImageNet, such as Brier score (and its component terms), as well as top-5 error for increasing values of shift. 

Overall, ensembles consistently perform best across metrics and dropout consistently performed better than temperature scaling and last layer methods. \emph{While the relative ordering of methods is consistent on both CIFAR-10 and ImageNet (ensembles perform best), the ordering is quite different from that on MNIST where SVI performs best.} Interestingly, LL-SVI and LL-Dropout perform worse than the vanilla method on shifted datasets as well as SVHN. We also evaluate a variational Gaussian process as a last layer method in Appendix~\ref{sec:vgp} but it did not outperform LL-SVI and LL-Dropout.%

\subsection{Text Models}%
\label{sec:results:text}%
 Following~\citet{hendrycks2016baseline}, we train an LSTM~\citep{hochreiter97} on the 20newsgroups dataset \citep{lang1995newsweeder} and assess the model's robustness under distributional shift and OOD text.
 We use the even-numbered classes (10 classes out of 20) as in-distribution and the 10 odd-numbered classes as shifted data. We provide additional details in Appendix~\ref{sec:20newsgroups_models}.%

\begin{figure}[t]%
    \centering%
    \begin{subfigure}{
      \includegraphics[width=0.16\linewidth]{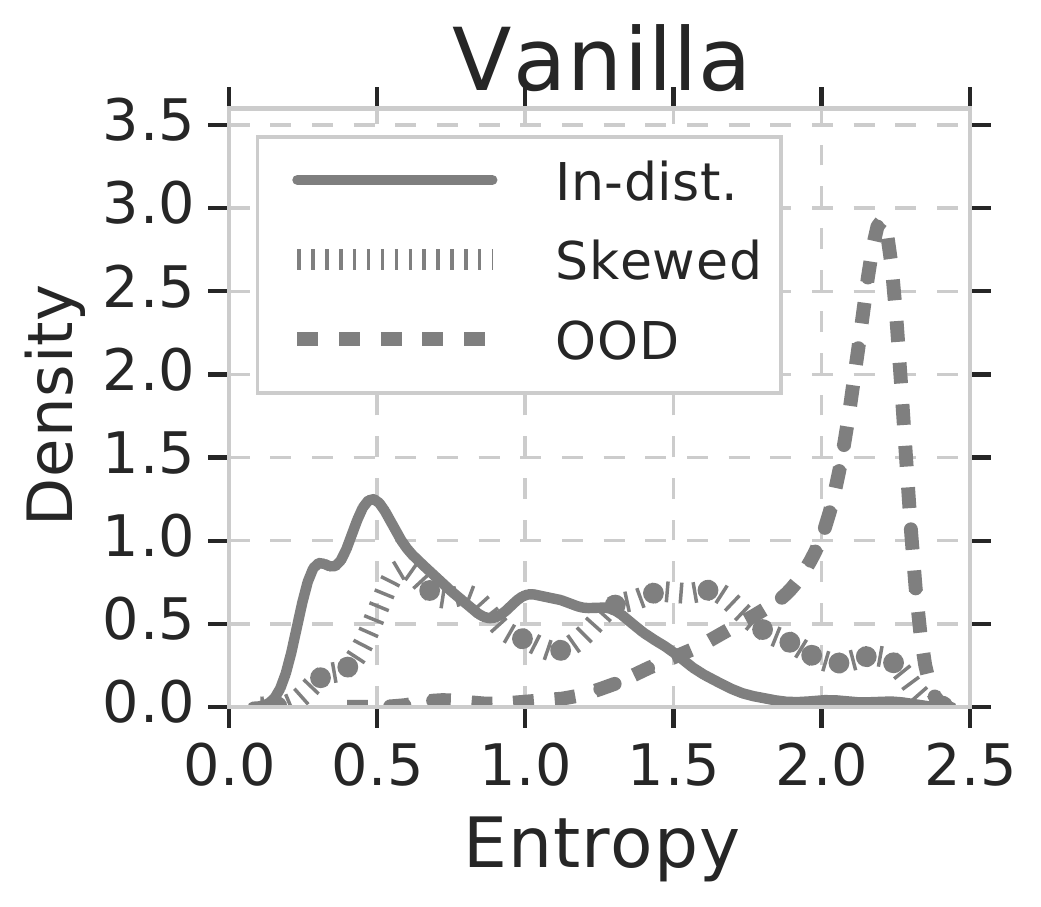}%
    }\end{subfigure}%
    \begin{subfigure}{
      \includegraphics[width=0.16\linewidth]{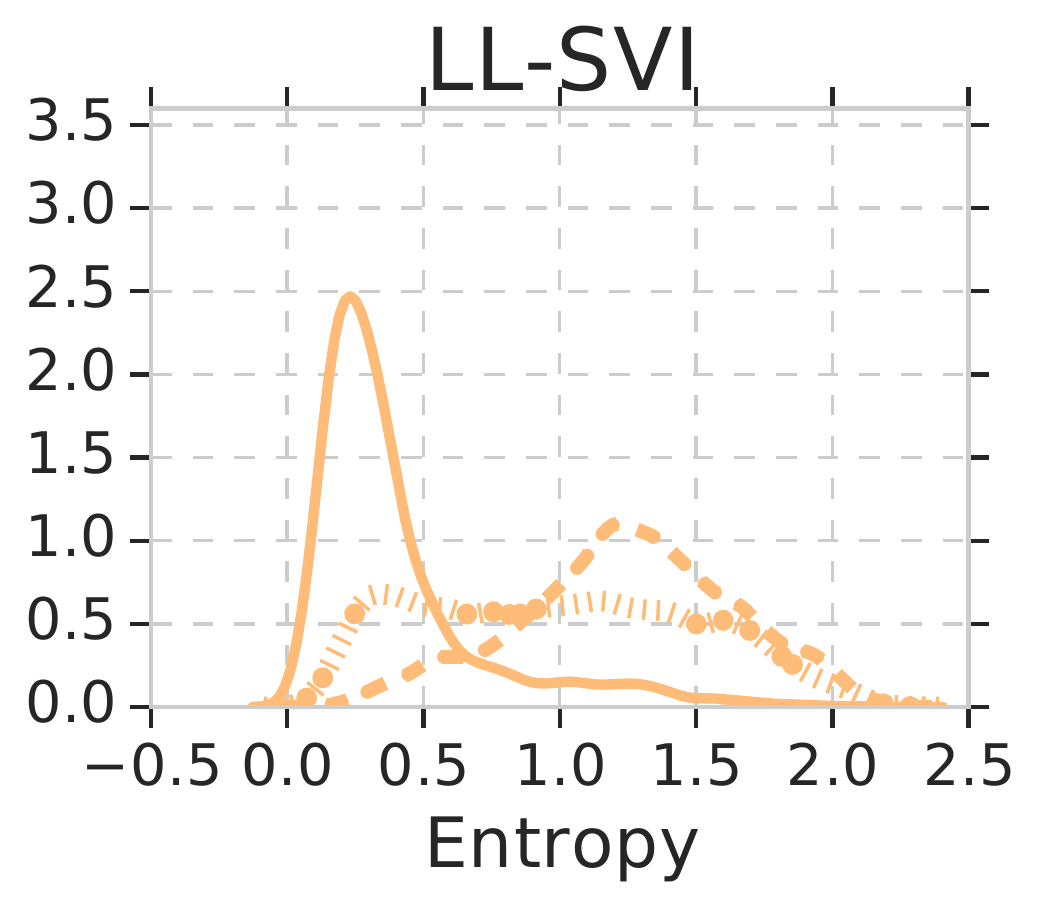}%
    }\end{subfigure}%
    \begin{subfigure}{
      \includegraphics[width=0.16\linewidth]{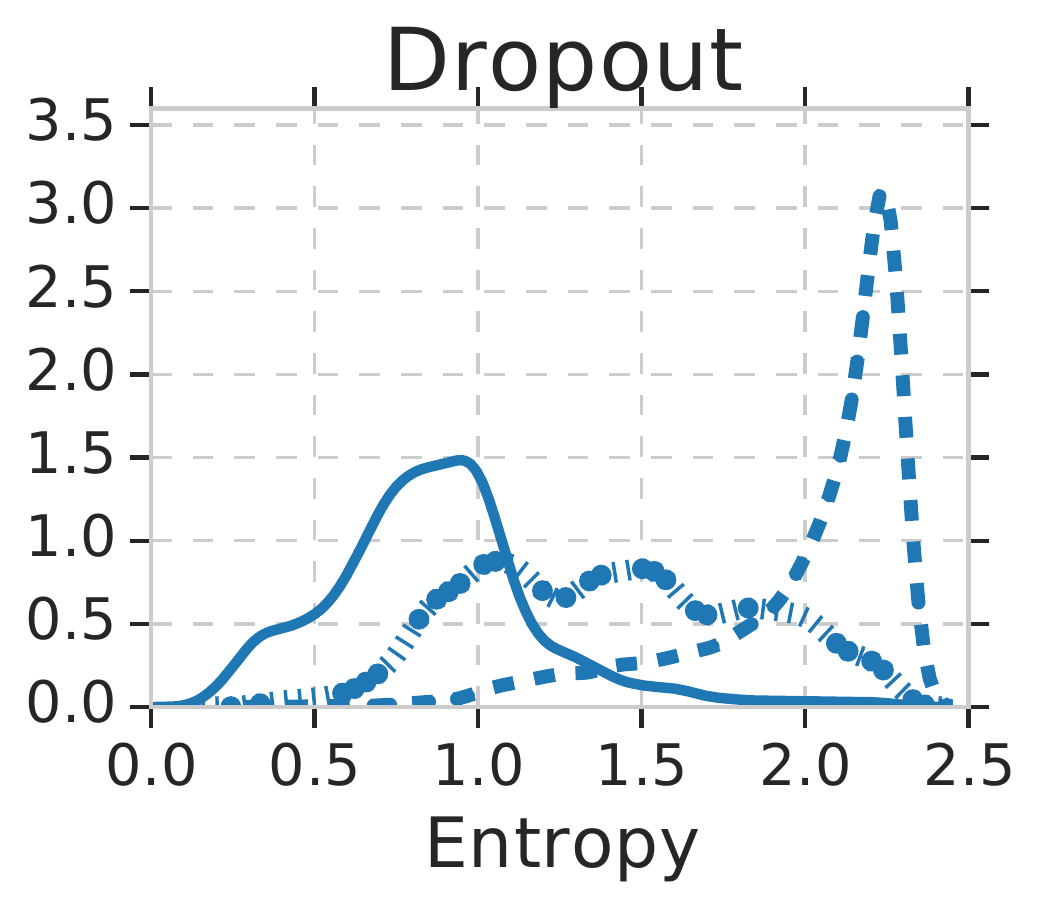}%
    }\end{subfigure}%
    \begin{subfigure}{
      \includegraphics[width=0.16\linewidth]{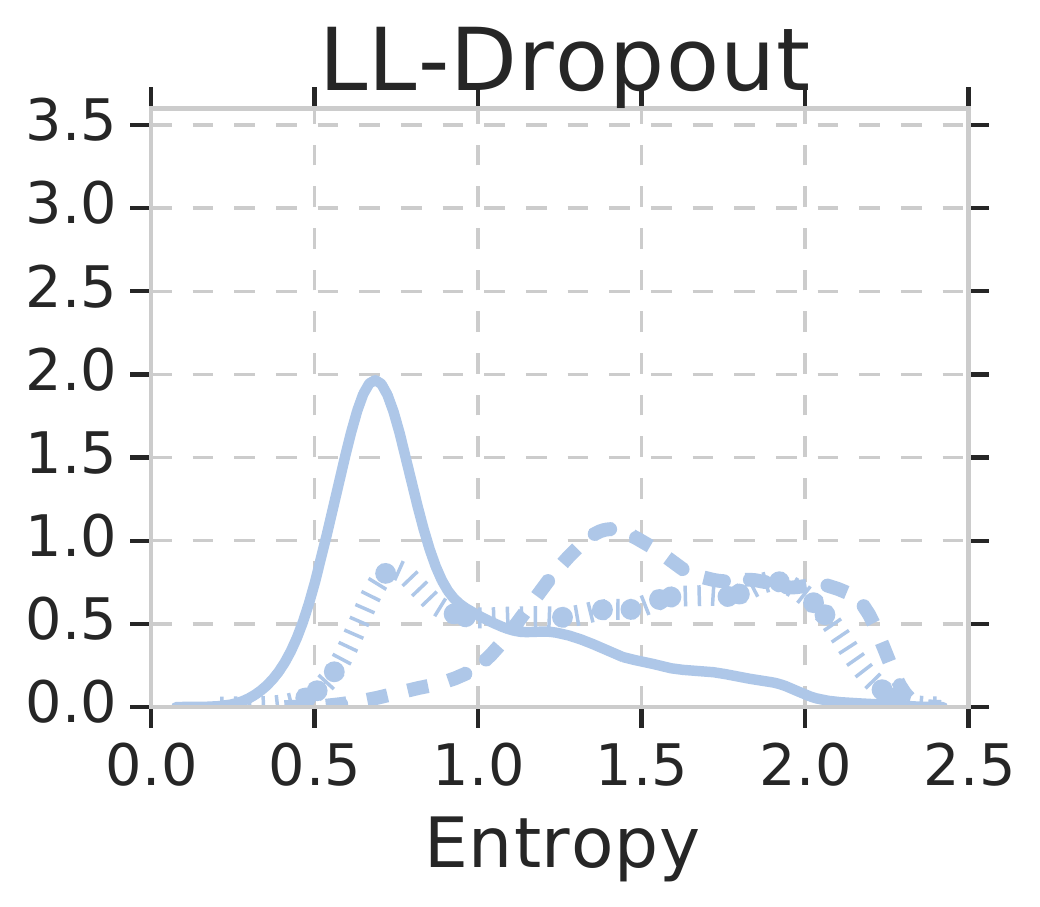}%
    }\end{subfigure}%
    \begin{subfigure}{
      \includegraphics[width=0.16\linewidth]{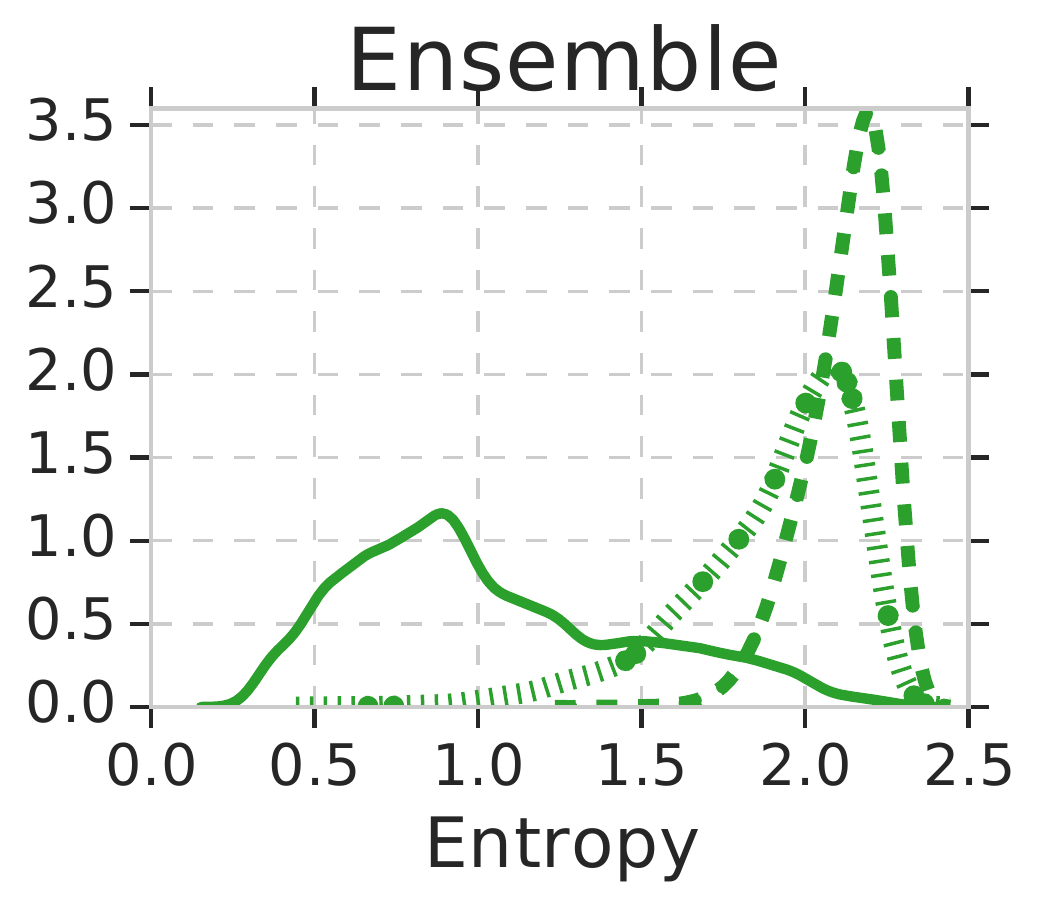}%
    }\end{subfigure}%
    \begin{subfigure}{
      \includegraphics[width=0.16\linewidth]{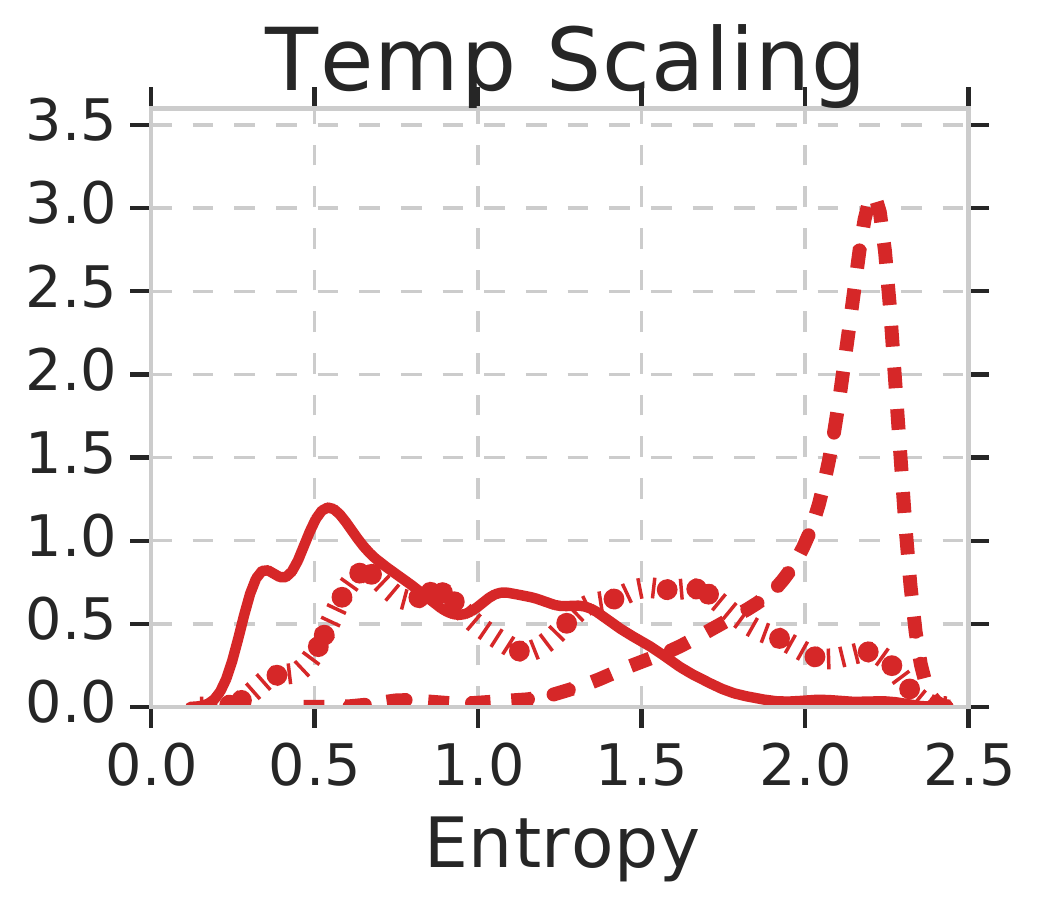}%
    }\end{subfigure}\\%
    \myvspace{-1em}%
    \addtocounter{subfigure}{-6}%
    \begin{subfigure}[Confidence vs Acc.]{
      \includegraphics[width=0.24\linewidth]{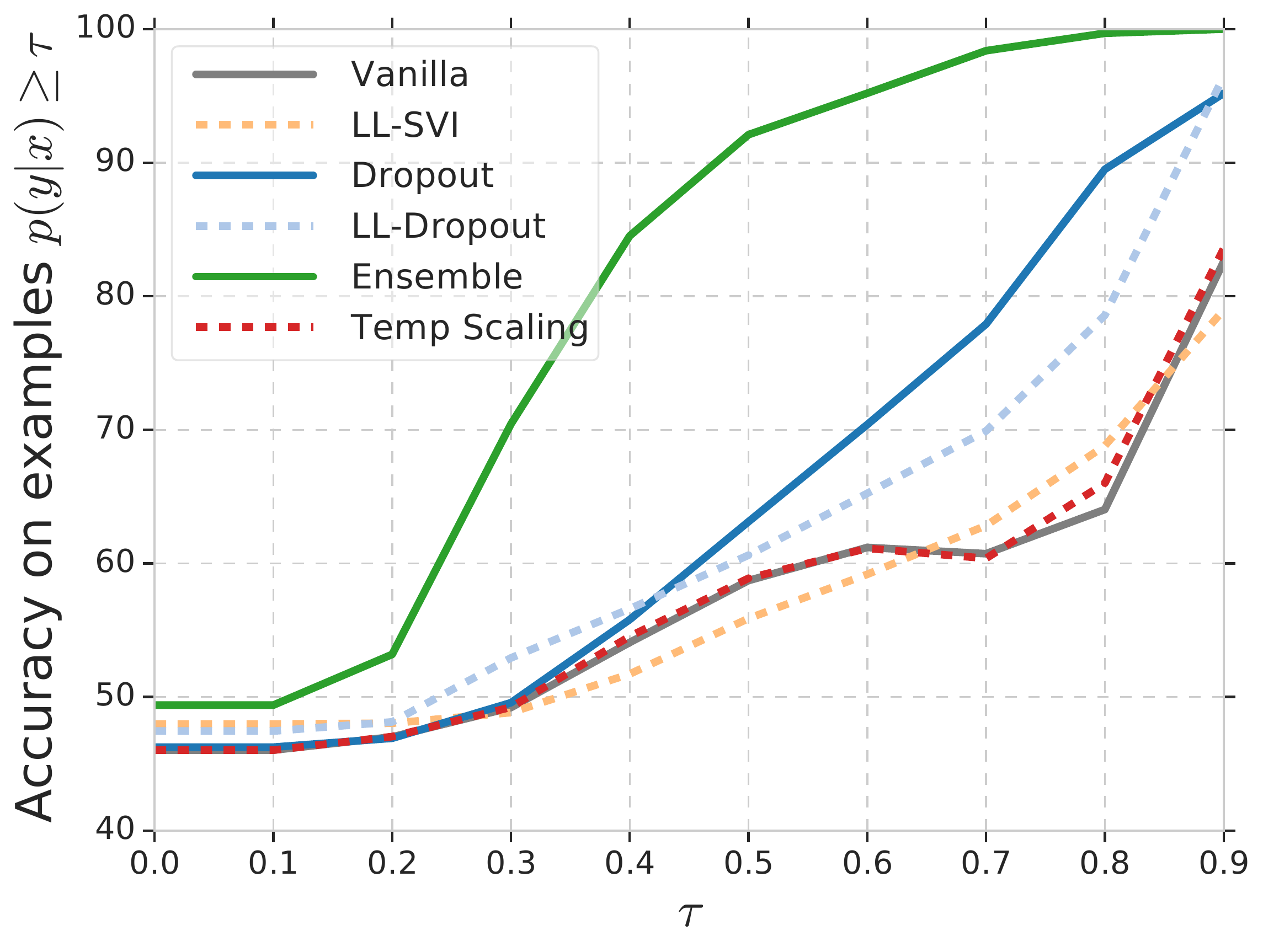}%
    }\end{subfigure}%
    \begin{subfigure}[Confidence vs Count]{
      \includegraphics[width=0.25\linewidth]{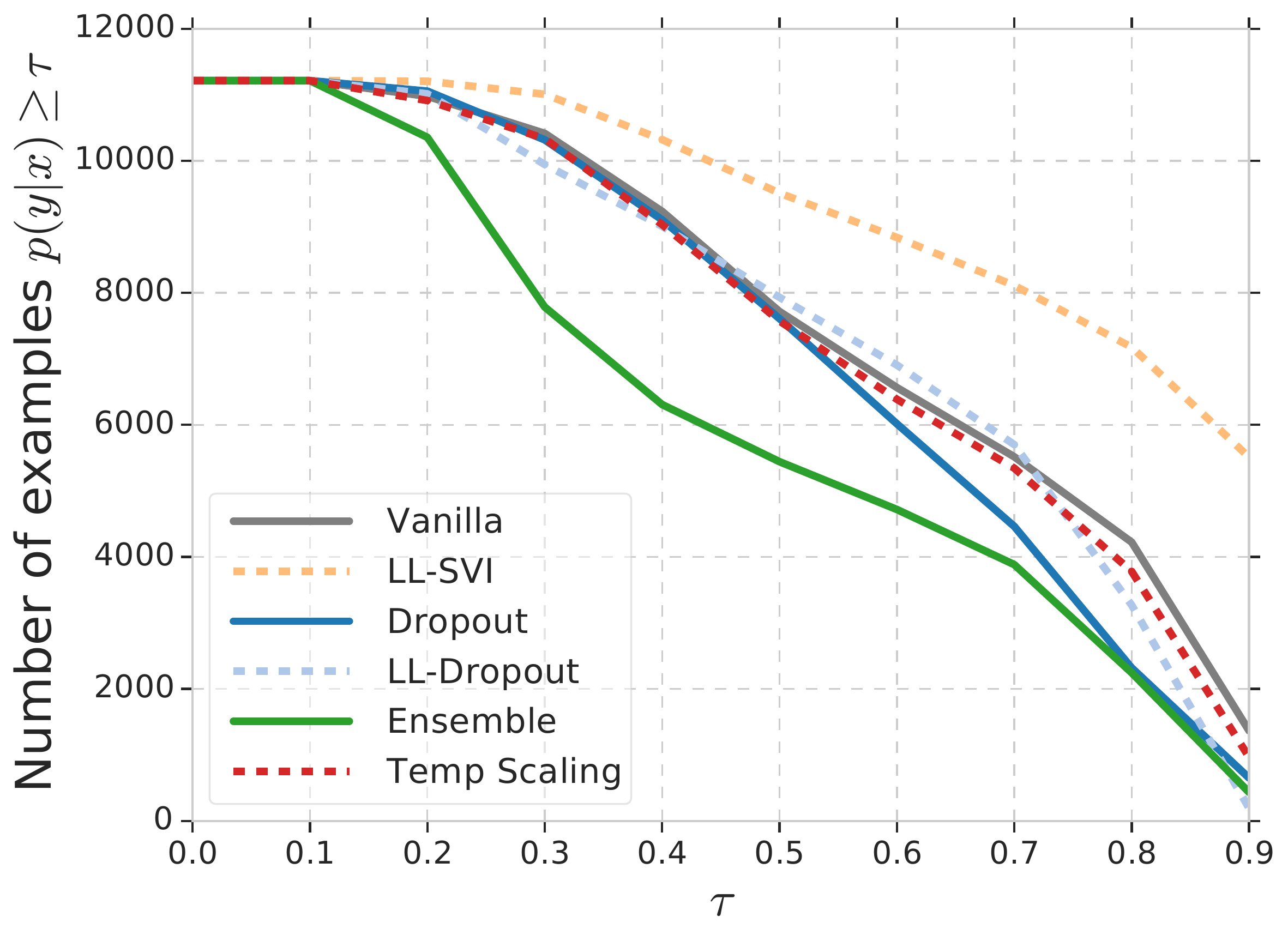}%
    }\end{subfigure}%
    \begin{subfigure}[Confidence vs Accuracy]{
      \includegraphics[width=0.25\linewidth]{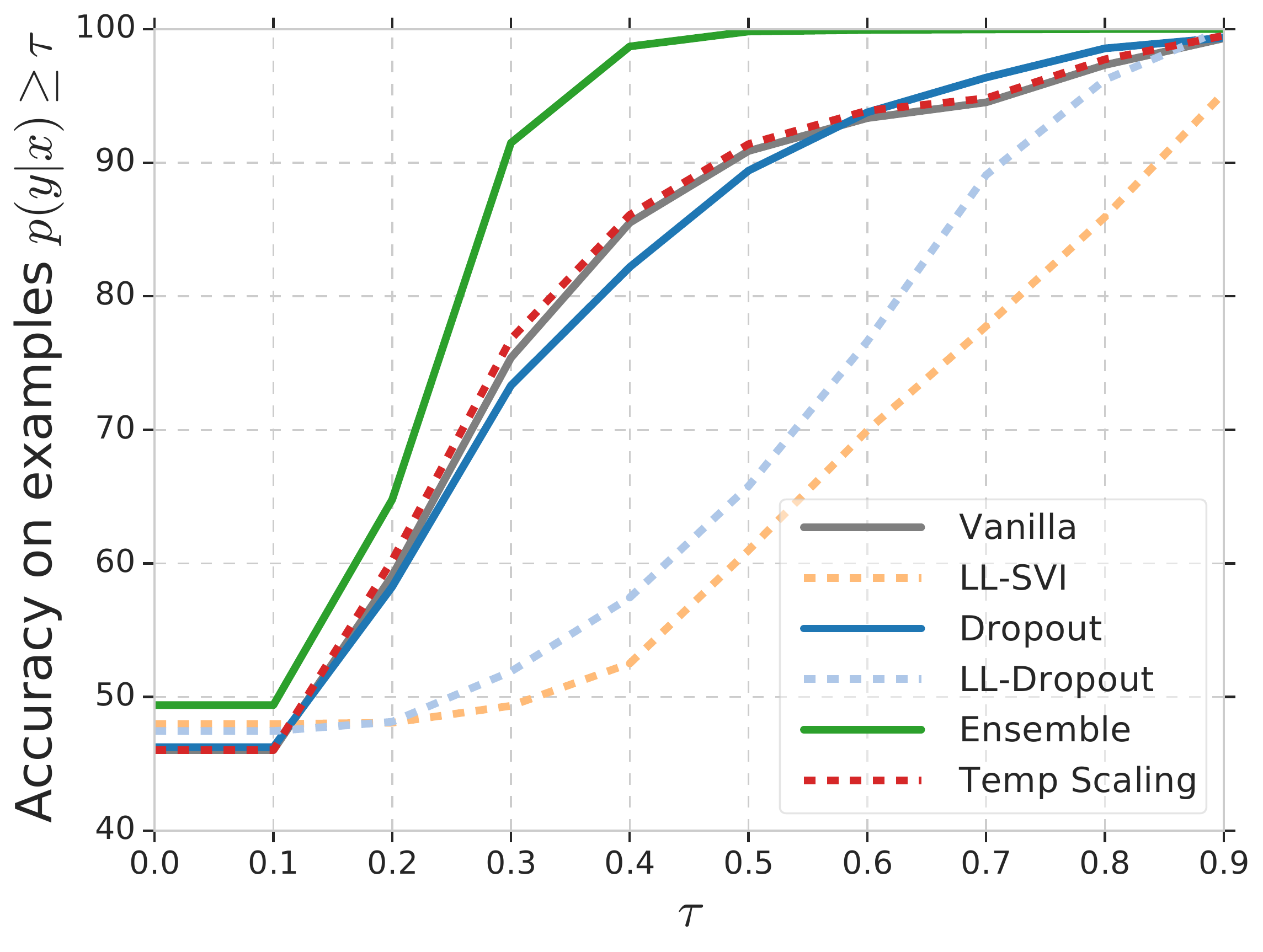}%
    }\end{subfigure}%
    \begin{subfigure}[Confidence vs Count]{
      \includegraphics[width=0.25\linewidth]{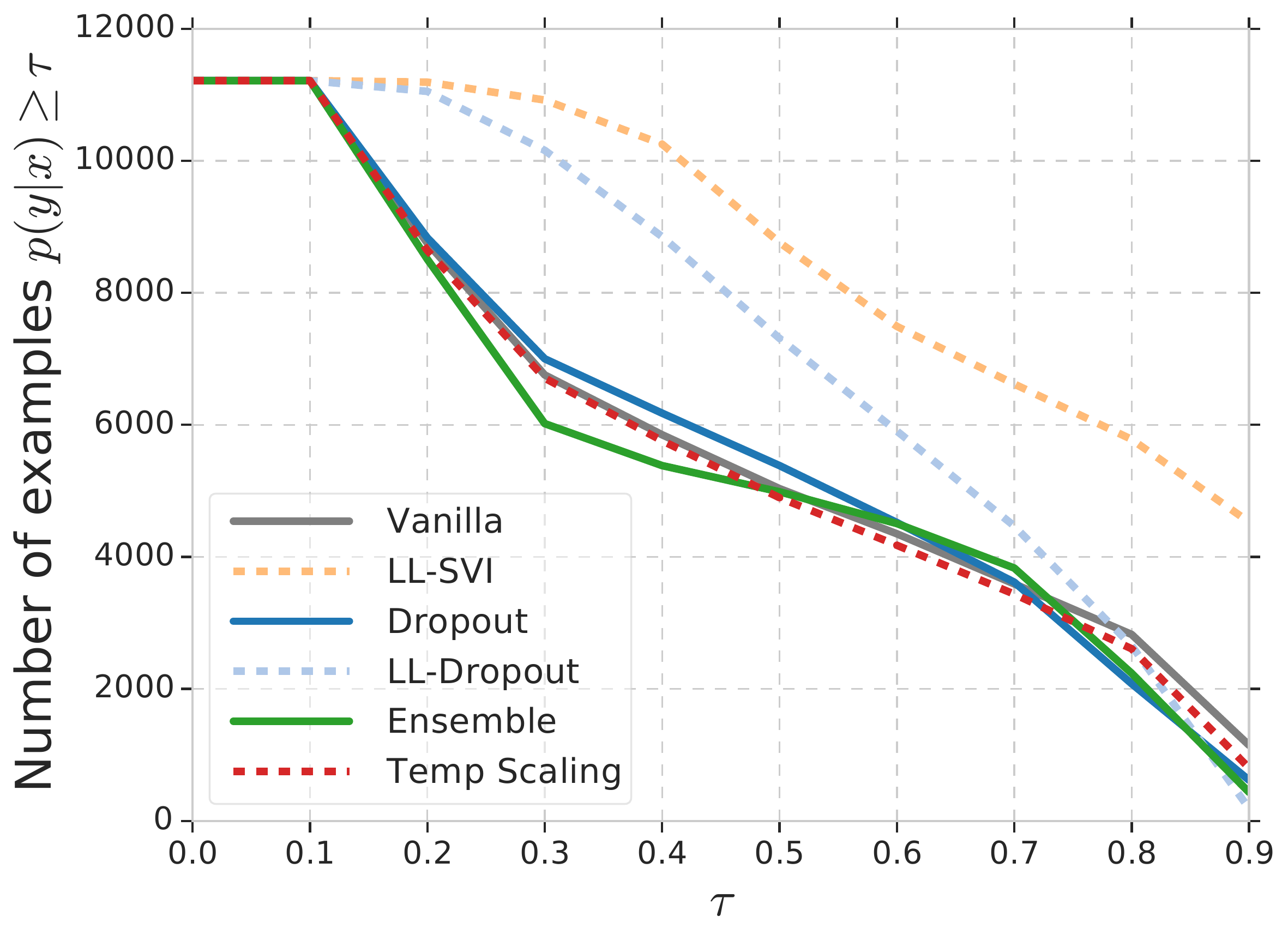}%
    }\end{subfigure}%
    \myvspace{-0.5em}%
    \caption{Top row: 
    Histograms of the entropy of the predictive distributions for in-distribution (solid lines), shifted (dotted lines), and completely different OOD (dashed lines) text examples. Bottom row: Confidence score vs accuracy and count respectively when evaluated for in-distribution and in-distribution shift text examples (a,b), and in-distribution and OOD text examples (c,d).}%
    \myvspace{-0.5em}%
    \label{fig:text_conf}%
\end{figure}

 We look at confidence vs accuracy when the test data consists of a mix of in-distribution and either shifted or completely OOD data, in this case the One Billion Word Benchmark (LM1B) \citep{chelba2013one}. Figure~\ref{fig:text_conf} (bottom row) shows the results. Ensembles significantly outperform all other methods, and achieve better trade-off between accuracy versus confidence.  Surprisingly, LL-Dropout and LL-SVI perform worse than the vanilla method, giving higher confidence incorrect predictions, especially when tested on fully OOD data. %

 \myvspace{-1em}%

Figure~\ref{fig:text_conf} reports histograms of predictive entropy on in-distribution data and compares them to those for the shifted and OOD datasets. This reflects how amenable each method is to abstaining from prediction by applying a threshold on the entropy. As expected, most methods achieve the highest predictive entropy on the completely OOD dataset, followed by the shifted dataset and then the in-distribution test dataset. Only ensembles have consistently higher entropy on the shifted data, which explains why they perform best on the confidence vs accuracy curves in the second row of Figure~\ref{fig:text_conf}. Compared with the vanilla model, Dropout and LL-SVI have more a distinct separation between in-distribution and shifted or OOD data.  While Dropout and LL-Dropout perform similarly on in-distribution, LL-Dropout exhibits less uncertainty than Dropout on shifted and OOD data. Temperature scaling does not appear to increase uncertainty significantly on the shifted data. 

\subsection{Ad-Click Model with Categorical Features}
\label{sec:results:criteo}%
Finally, we evaluate the performance of different methods on the \emph{Criteo Display Advertising Challenge}\footnote{\url{https://www.kaggle.com/c/criteo-display-ad-challenge}} dataset, a binary classification task consisting of 37M examples with 13 numerical and 26 categorical features per example.
We introduce shift by reassigning each categorical feature to a random new token with some fixed probability that controls the intensity of shift. 
{This coarsely simulates a type of shift observed in non-stationary categorical features as category tokens appear and disappear over time, for example due to hash collisions.}
The model consists of a 3-hidden-layer multi-layer-perceptron (MLP) with hashed and embedded categorical features and achieves a negative log-likelihood of approximately 0.5 (contest winners achieved 0.44). Due to class imbalance ($\sim25\%$ of examples are positive), we report AUC instead of classification accuracy.

\begin{figure}[ht]%
    \centering%
      \includegraphics[width=0.245\linewidth]{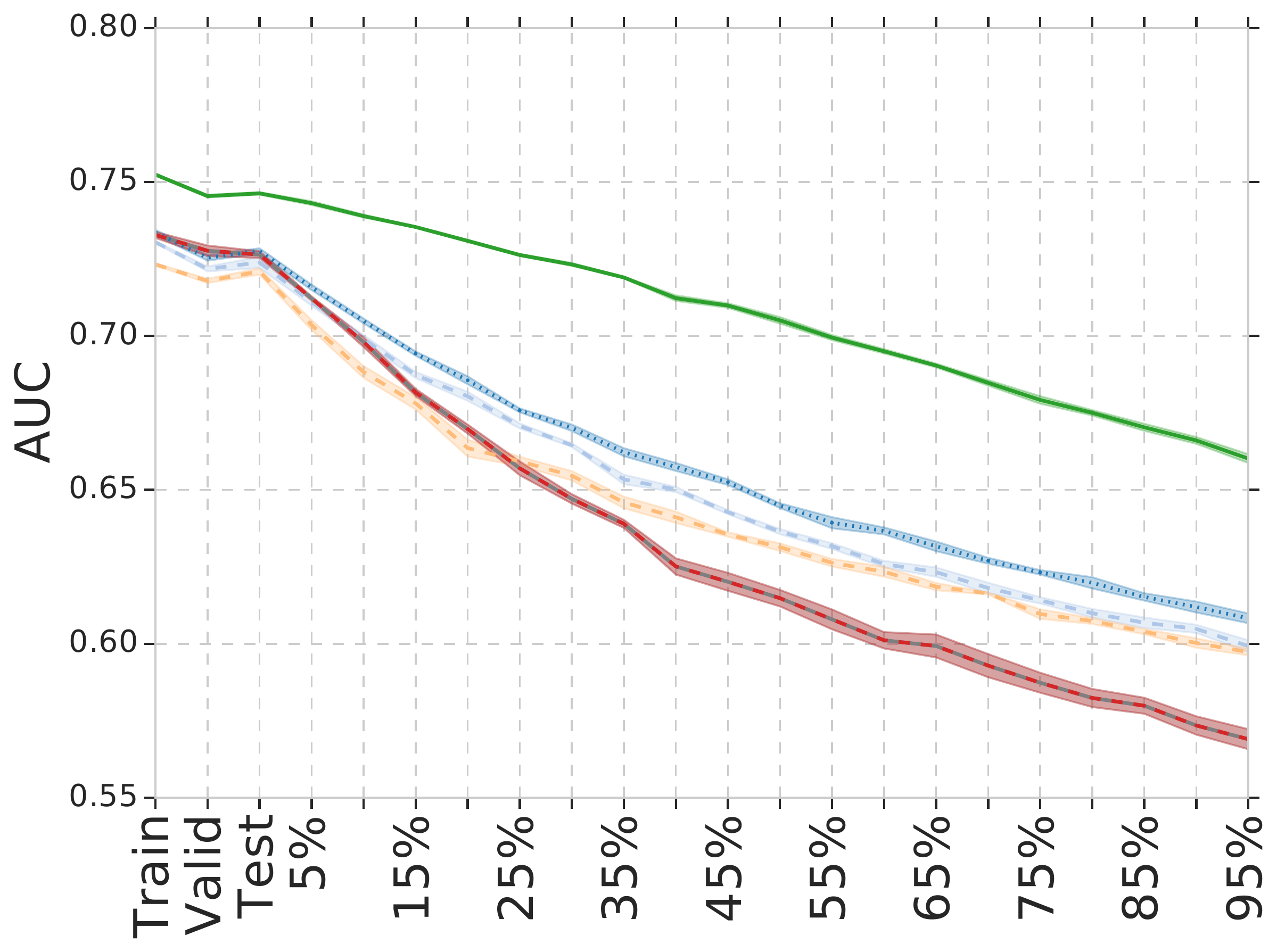}%
      \includegraphics[width=0.245\linewidth]{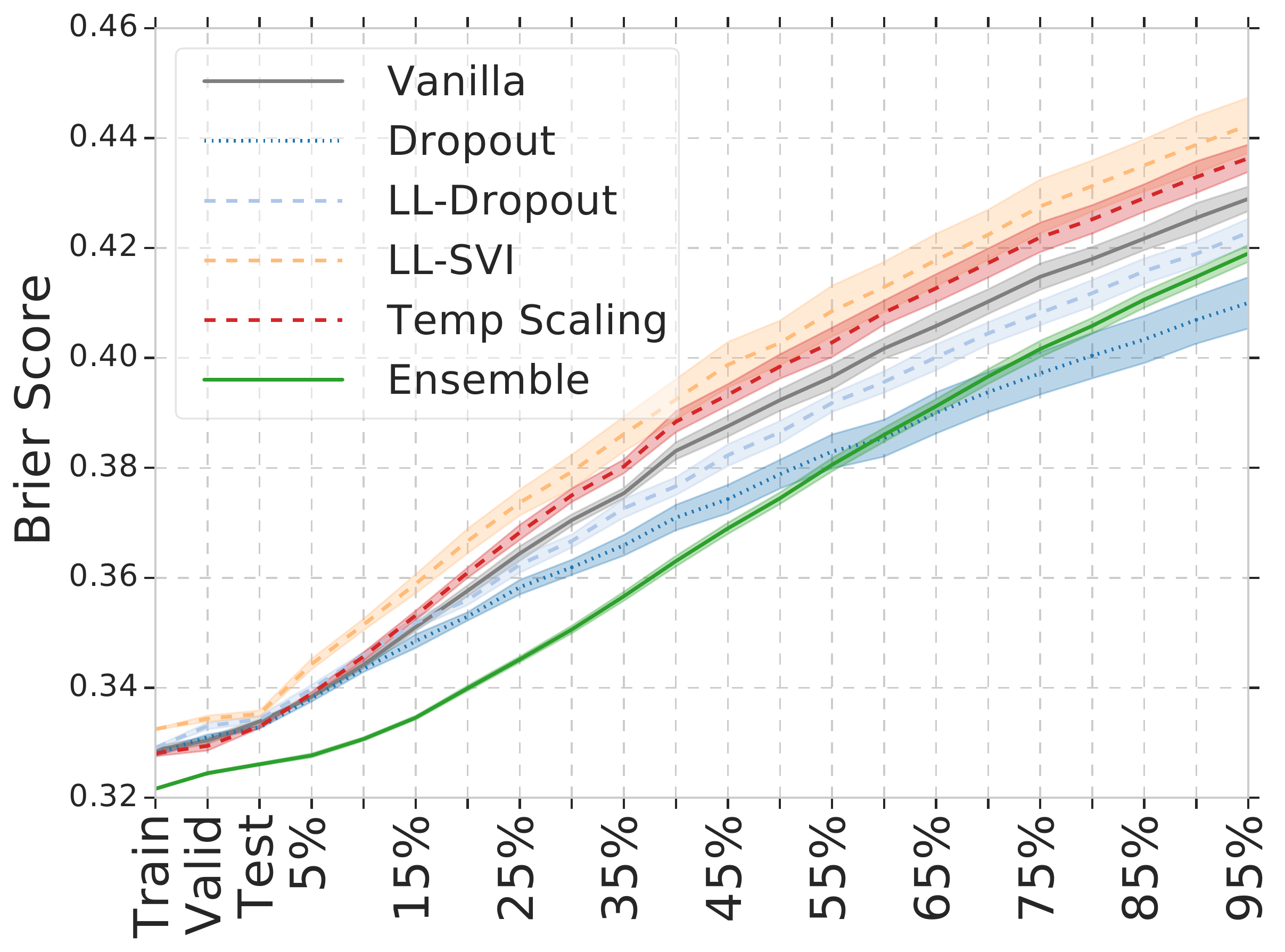}%
      \includegraphics[width=0.245\linewidth]{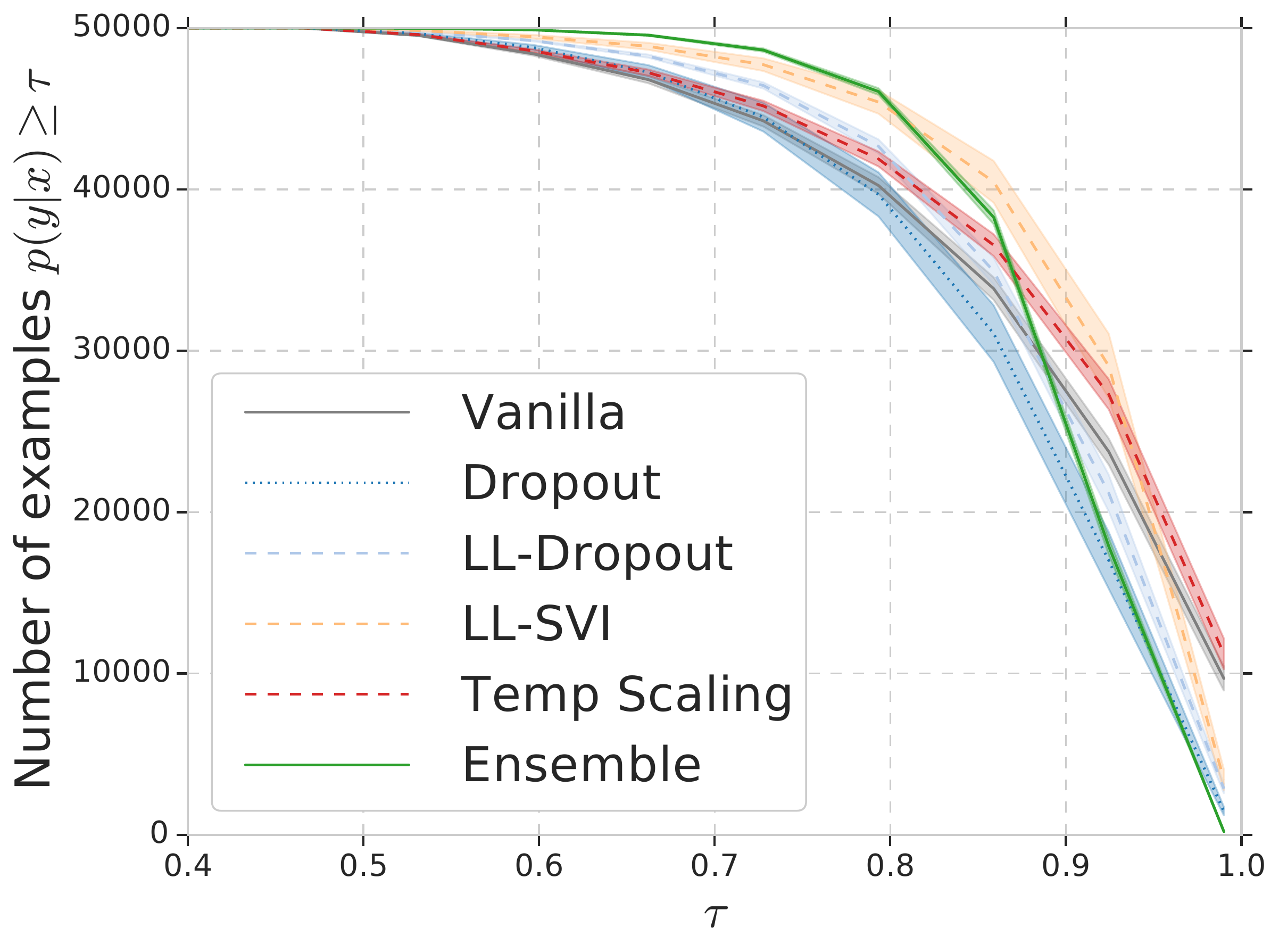}%
      \includegraphics[width=0.245\linewidth]{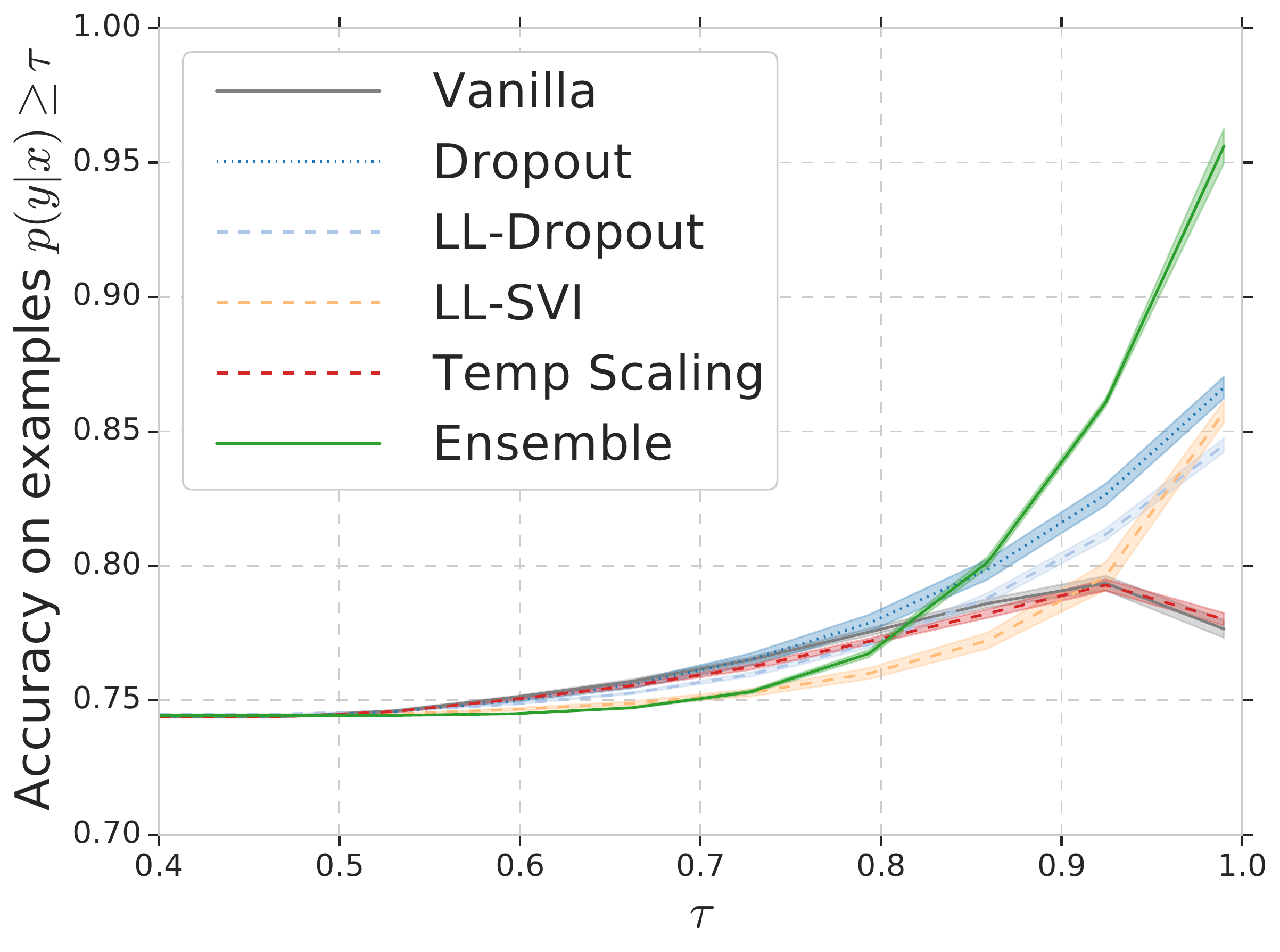}%
    \caption{Results on Criteo: The first two plots show degrading AUCs and Brier scores with increasing shift while the latter two depict the distribution of prediction confidences and their corresponding accuracies at 75\% randomization of categorical features.
    SVI is excluded as it performed too poorly.
    }
    \label{fig:criteo_summary_figures}%
\end{figure}%

Results from these experiments are depicted in Figure~\ref{fig:criteo_summary_figures}.  
(Figure~\ref{fig:criteo_addl} in Appendix~\ref{sec:calibration:skew:additional} shows additional results including ECE and Brier score decomposition.)  
We observe that ensembles are superior in terms of both AUC and  Brier score for most of the values of shift, with the performance gap between ensembles and other methods generally increasing as the shift increases. Both Dropout model variants yielded improved AUC on shifted data, and Dropout surpassed ensembles in Brier score at shift-randomization values above 60\%. SVI proved challenging to train, and the resulting model uniformly performed poorly; LL-SVI fared better but generally did not improve upon the vanilla model.  \emph{Strikingly,  temperature scaling has a worse Brier score than Vanilla indicating that post-hoc calibration on the validation set actually harms calibration under dataset shift.}

\section{Takeaways and Recommendations}
We presented a large-scale evaluation of different methods for quantifying predictive uncertainty under dataset shift, across different data modalities and architectures. Our take-home messages are the following:
\begin{itemize}
    \item Along with accuracy, the quality of uncertainty consistently degrades with increasing dataset shift regardless of method.
    \item Better calibration and accuracy on the i.i.d.~test dataset does not usually translate to better calibration under dataset shift (shifted versions as well as completely different OOD data). 
   \item Post-hoc calibration (on i.i.d~validation) with temperature scaling leads to well-calibrated uncertainty on the i.i.d.~test set and small values of shift, but is significantly outperformed by methods that take epistemic uncertainty into account as the shift increases.
         \item Last layer Dropout exhibits less uncertainty on shifted and OOD datasets than Dropout.
      \item SVI is very promising on MNIST/CIFAR but it is 
      difficult to get to work on larger datasets such as ImageNet and other architectures such as LSTMs.
      \item The relative ordering of methods is mostly consistent (except for 
      MNIST) across our experiments.  
     The relative ordering of  methods on MNIST is not reflective of their ordering on other datasets. 
   \item  Deep 
   ensembles 
   seem to perform the best across most metrics and be more robust to dataset shift. 
   We found that relatively small ensemble size (e.g. $M=5$) may be sufficient (Appendix~\ref{sec:sample:size}).
   \item We also compared the set of methods on a real-world challenging genomics problem from \cite{ren2019likelihood}. Our observations were consistent with the other experiments in the paper. Deep ensembles performed best, but there remains significant room for improvement, as with the other experiments in the paper.  See Section \ref{sec:gene} for details.  
\end{itemize}
We hope that this benchmark is useful to the community and inspires more research on uncertainty under dataset shift, which seems challenging for existing methods. 
While we focused only on the quality of predictive uncertainty, applications may also need to consider computational and memory costs of the methods; Table~\ref{table:costs} in Appendix~\ref{sec:costs} discusses these costs, and the best performing methods tend to be more expensive. Reducing the computational and memory costs, while retaining the same performance under dataset shift, would also be a key research challenge.%

\subsubsection*{Acknowledgements}
We thank Alexander D'Amour, Jakub Świ\c atkowski and our reviewers for helpful feedback that improved the manuscript.

{ \bibliography{main}}

\bibliographystyle{icml2019}
\clearpage
\newpage
\appendix

\setcounter{figure}{0}
\setcounter{table}{0}
\makeatletter 
\renewcommand{\thefigure}{S\@arabic\c@figure}
\renewcommand{\thetable}{S\@arabic\c@table}
\makeatother

{\Large{\textbf{
\begin{center}
Can You Trust Your Model's Uncertainty? Evaluating Predictive Uncertainty Under Dataset Shift: Appendix  
\end{center}
}}}

\section{Model Details}
\label{sec:model_details}
\subsection{MNIST}
We evaluated both LeNet and a fully-connected neural network (MLP) under shift on MNIST.  We observed similar trends across metrics for both models, so we report results only for LeNet in Section~\ref{sec:mnist}.  LeNet and MLP were trained for 20 epochs using the Adam optimizer \citep{kingma2014adam} and used ReLU activation functions. For stochastic methods, we averaged 300 sample predictions to yield a predictive distribution, and the ensemble model used 10 instances trained from independent random initializations.
The MLP architecture consists of two hidden layers of 200 units each with dropout applied before every dense layer. The LeNet architecture \citep{lecun-98} applies two convolutional layers 3x3 kernels of 32 and 64 filters respectively) followed by two fully-connected layers with one hidden layer of 128 activations; dropout was applied before each fully-connected layer.
We employed hyperparameter tuning (See Section~\ref{sec:hyperparameter_tuning}) to select the training batch size, learning rate, and dropout rate.

\subsection{CIFAR-10}
Our CIFAR model used the ResNet-20 V1 architecture with ReLU activations. Model parameters were trained for 200 epochs using the Adam optimizer and employed a learning rate schedule that multiplied an initial learning rate by 0.1, 0.01, 0.001, and 0.0005 at steps 80, 120, 160, and 180 respectively. Training inputs were randomly distorted using horizontal flips and random crops preceded by 4-pixel padding as described in~\citep{resnet}. For relevant methods, dropout was applied before each convolutional and dense layer (excluding the raw inputs), and stochastic methods sampled 128 predictions per sample. Hyperparameter tuning was used to select the initial learning rate, training batch size, and the dropout rate.

\subsection{ImageNet 2012}
Our ImageNet model used the ResNet-50 V1 architecture with ReLU activations and was trained for 90 epochs using SGD with Nesterov momentum. The learning rate schedule linearly ramps up to a base rate in 5 epochs and scales down by a factor of 10 at each of epochs 30, 60, and 80. As with the CIFAR-10 model, stochastic methods used a sample-size of 128. Training images were distorted with random horizontal flips and random crops.

\subsection{20 Newsgroups}
\label{sec:20newsgroups_models}
We use a pre-processing strategy similar to the one proposed by \citet{hendrycks2016baseline} for 20 Newsgroups.
We build a vocabulary of size 30,000 words and words are indexed based on the word frequencies.
The rare words are encoded as unknown words.  
We fix the length of each text input by setting a limit of 250 words, and those longer than 250 words are truncated, and those shorter than 250 words are padded with zeros. 
Text in even-numbered classes are used as in-distribution inputs, and text from the odd-numbered of classes are used shifted OOD inputs. 
A dataset with the same number of  randomly selected text inputs from the LM1B dataset \citep{chelba2013one} is used as completely different OOD dataset. 
The classifier is trained and evaluated only using the text from the even-numbered in-distribution classes in the training dataset. The final test results are evaluated based on in-distribution test dataset, shift OOD test dataset, and LM1B dataset. 

The vanilla model uses a one-layer LSTM model of size 32 and a dense layer to predict the 10 class probabilities based on word embedding of size 128. A dropout rate of 0.1 is applied to both the LSTM layer and the dense layer for the Dropout model. The LL-SVI model replaces the last dense layer with a Bayesian layer, the ensemble model aggregates 10 vanilla models, and stochastic methods sample 5 predictions per example. The vanilla model accuracy for in-distribution test data is 0.955.

\subsection{Criteo}
Each categorical feature $x_k$ from the Criteo dataset was encoded by hashing the string token into a fixed number of buckets $N_k$ and either encoding the hash-bin as a one-hot vector if $N_k < 110$ or embedding each bucket as a $d_k$ dimensional vector otherwise. This dense feature vector, concatenated with 13 numerical features, feeds into a batch-norm layer followed by a 3-hidden-layer MLP. Each model was trained for one epoch using the Adam optimizer with a non-decaying learning rate.

Values of $N_k$ and $d_k$ were tuned to maximize log-likelihood for a vanilla model, and the resulting architectural parameters were applied to all methods.
This tuning yielded hidden-layers of size 2572, 1454, and 1596, and hash-bucket counts and embedding dimensions of sizes listed below:
\begin{align*}
  N_k = [&1373, 2148, 4847, 9781, 396, 28, 3591, 2798, 14, 7403, 2511, 5598, 9501, \\
         &46, 4753, 4056, 23, 3828, 5856, 12, 4226, 23, 61, 3098, 494, 5087] \\
  d_k = [&3, 9, 29, 11, 17, 0, 14, 4, 0, 12, 19, 24, 29, 0, 13, 25, 0, 8, 29, 0, 22, 0, 0, 31, 0, 29] \\
\end{align*}
Learning rate, batch size, and dropout rate were further tuned for each method.
Stochastic methods used 128 prediction samples per example.

\subsection{Stochastic Variational Inference Details}\label{sec:svi}
For MNIST we used Flipout~\citep{flipout}, where we replaced each dense layer and convolutional layer with mean-field variational dense and convolutional Flipout layers respectively.  Variational inference for deep ResNets~\citep{resnet} is non-trivial, so for CIFAR we replaced a single linear layer per residual branch with a Flipout layer, removed batch normalization, added Selu non-linearities~\citep{klambauer2017}, empirical Bayes for the prior standard deviations as in~\citet{wu2018} and careful tuning of the initialization via Bayesian optimization.

\subsection{Variational Gaussian Process Details}\label{sec:gaussian_processes}
For the experiments where Gaussian Processes were compared, we used Variational Gaussian Processes to fit the model logits as in~\cite{hensman2015scalable}. These were then passed through a Categorical distribution and numerically integrated over using Gauss-Hermite quadrature. Each class was treated as a separate Gaussian Process, with 100 inducing points used for each class. The inducing points were initialized with model outputs on random dataset examples for CIFAR, and with Gaussian noise for MNIST. Uniform noise inducing point initialization was also tested but there was negligible difference between the three methods. All zero inducing points initializations numerically failed early on in training. Exponentiated quadratic plus linear kernels were used for all experiments. 250 samples were drawn from the logit distribution during training time to get a better estimate of the ELBO to backpropagate through. 250 logit samples were drawn at test time. $10^{-5} * I$ was added to the diagonal of the covariance matrix to ensure positive definiteness.

We used 100 trials of random hyperparamter settings, selecting the configuration with the best final validation accuracy. The learning rate was tuned in $\lbrack 10^{-4}, 1.0\rbrack$ on a log scale; the initial kernel amplitude in $\lbrack -2.0, 2.0\rbrack$; the initial kernel length scale in $\lbrack -2.0, 2.0\rbrack$; the variational distribution covariance was initialized to $s * I$ where $s$ was tuned in $\lbrack 0.1, 2.0\rbrack$; $1 - \beta_1$ in Adam was tuned on $\lbrack 10^{-2}, 0.15\rbrack$ on a log scale.

The Adam optimizer with a batch size of 512 was used, training for the same number of epochs as other methods. The same learning rate schedule was as other methods for the model and kernel parameters, but the learning rate for the variational parameters also included a 5 epoch warmup in order to help with numerical stability.

\subsection{Hyperparameter Tuning}
\label{sec:hyperparameter_tuning}
Hyperparameters were optimized through Bayesian optimization using Google Vizier \citep{vizier}.  
We maximized the log-likelihood on a validation set that was held out from training (10K examples for MNIST and CIFAR-10, ~125K examples for ImageNet).  We optimized log-likelihood rather than accuracy since the former is a proper scoring rule.

\subsection{Computational and Memory Complexity of Different methods}\label{sec:costs}
   In addition to performance, applications may also need to consider computational and memory costs; Table~\ref{table:costs} 
   discusses them for each method. 

\begin{table}[ht]
\centering
\caption{Computational and memory costs for evaluated methods. Notation: $m$ represents flops or storage for the full model, $d$ represents flops or storage for the last layer, $k$ denotes replications, $z$ the number of inducing points for Gaussian Processes, $n$ denotes number of evaluated points, and $v$ denotes the validation set size. Serving/training compute is identical except that $v=0$ for serving. Implicit in this table is a memory/compute tradeoff for sampling. Sampled weights/masks need not be stored explicitly via PRNG seed reuse; we assume the computational cost of sampling is zero.}\label{table:costs}%
\begin{tabular}{rll}
    Method    & Compute/$n$    & Storage   \\ \hline
     Vanilla  &  $m$           &  $m$     \\
Temp Scaling  &  $m + v m/n$   &  $m$     \\
  LL-Dropout  &  $m + d(k-1)$  &  $m$     \\
      LL-SVI  &  $m + d(k-1)$  &  $m + d$ \\
         SVI  &  $m k$         &  $2 m$   \\
     Dropout  &  $m k$         &  $m$     \\
Gaussian Process  &  $m + z^3$         &  $m + z^2$     \\
    Ensemble  &  $m k$         &  $m k$   \\
\end{tabular}

\end{table}

\section{Shifted Images}\label{sec:skew:examples}
\begin{figure}[h]
    \centering
    \begin{subfigure}[Rotations]{
      \includegraphics[width=0.7\linewidth]{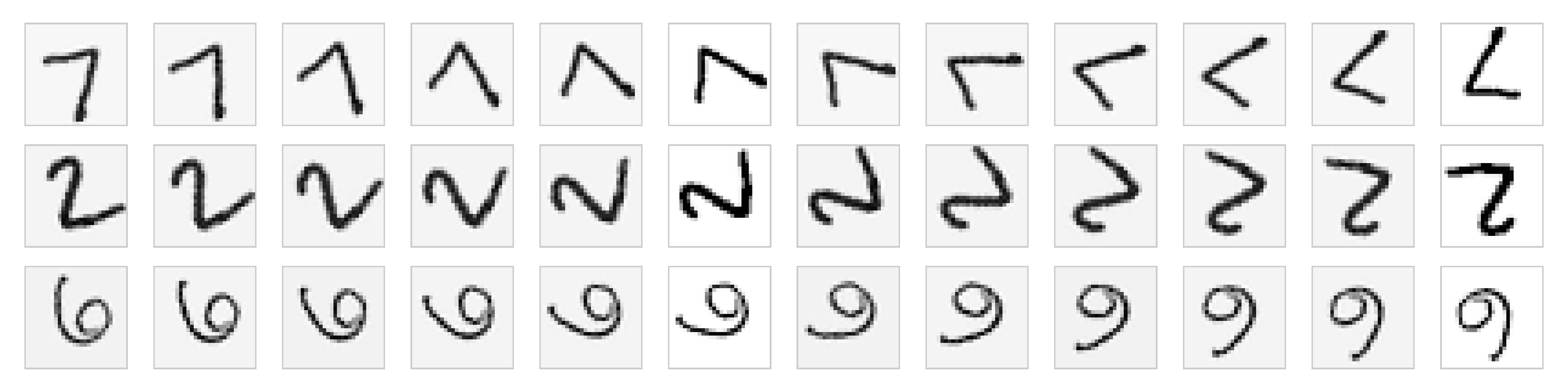}
    }\end{subfigure}
    \begin{subfigure}[Cyclic translations]{
      \includegraphics[width=0.7\linewidth]{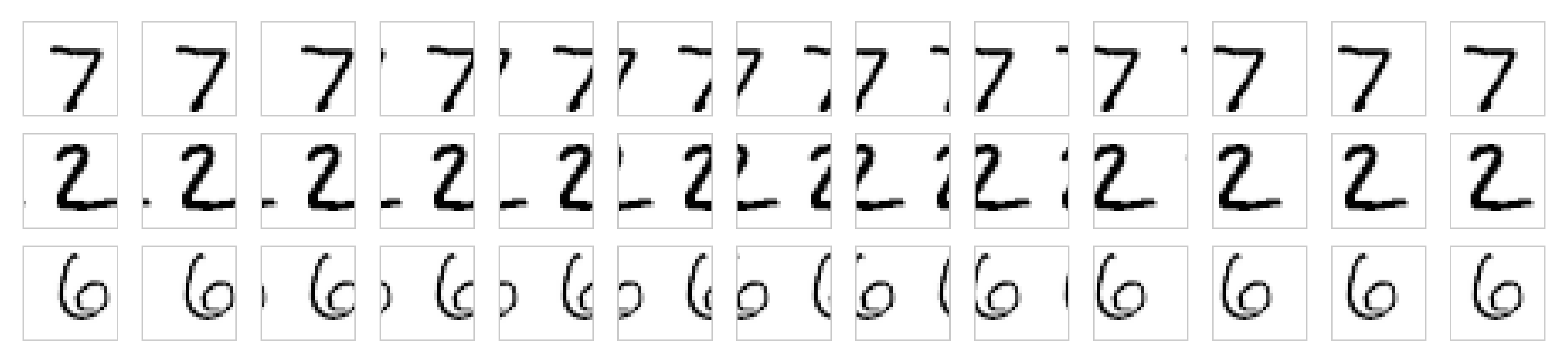}
    }\end{subfigure}
    \caption{Examples of rotated and cyclically translated MNIST digits. Results for accuracy and calibration on rotated/translated MNIST are shown in Figure~\ref{fig:mnist}.}
    \label{fig:mnist_image_samples}
\end{figure}

We distorted MNIST images using rotations with spline filter interpolation and cyclic translations as depicted in Figure~\ref{fig:mnist_image_samples}.

\begin{figure}[h]
    \centering
    \includegraphics[width=0.7\linewidth]{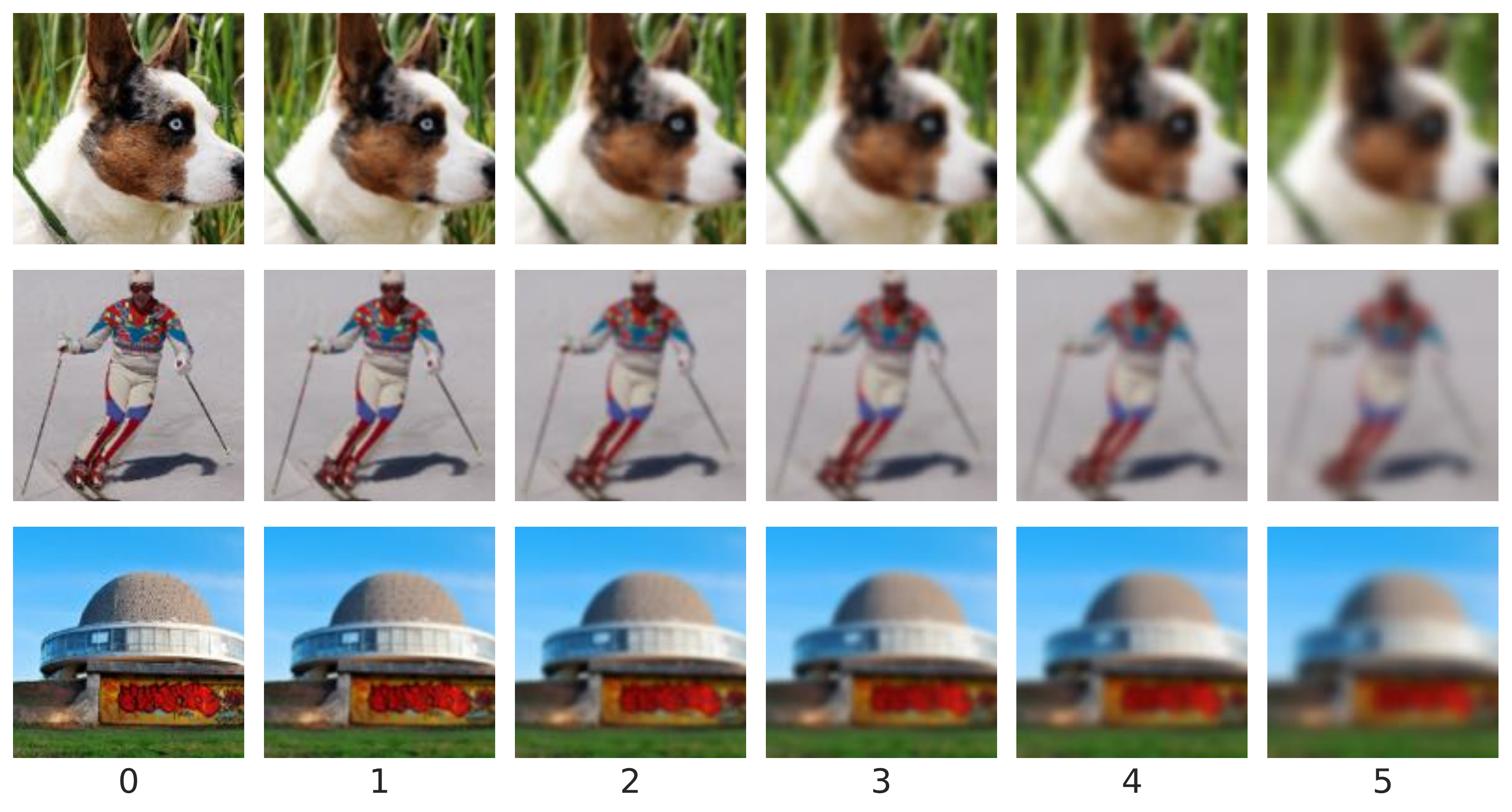}
    \caption{Examples of ImageNet images corrupted by Gaussian blur, at intensities of 0 (uncorrupted image) through 5 (maximum corruption included in ImageNet-C).
    }
    \label{fig:imagenet_gaussian_blur_samples}
\end{figure}

\begin{figure}[h]
    \centering
     \includegraphics[width=0.49\linewidth]{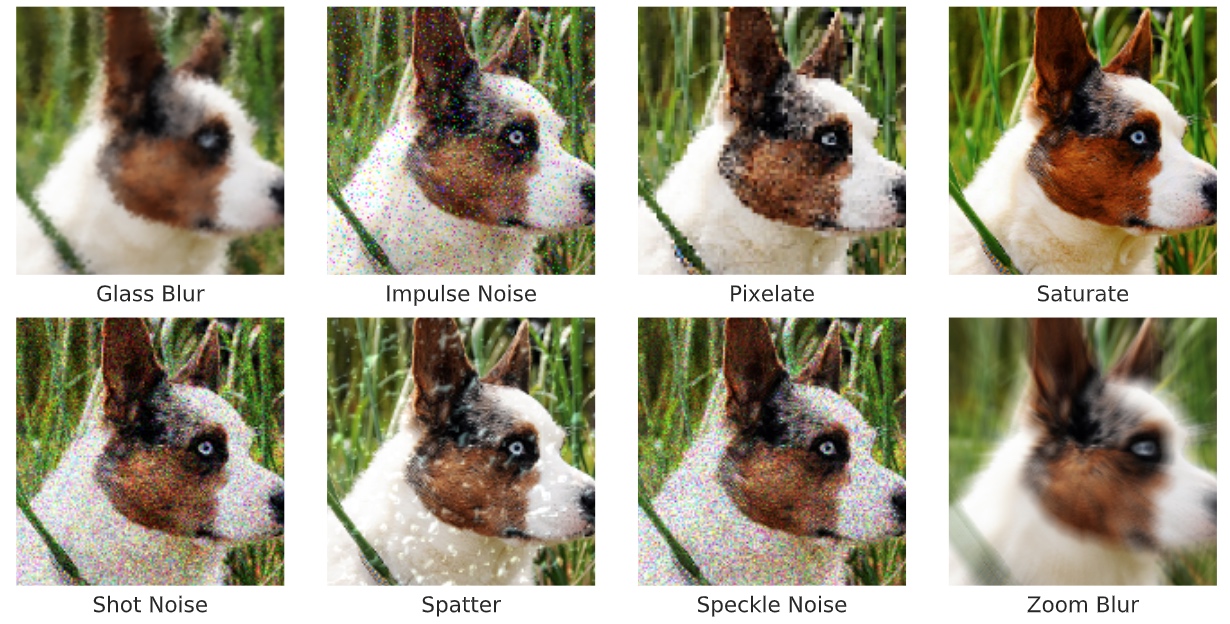}
      \includegraphics[width=0.49\linewidth]{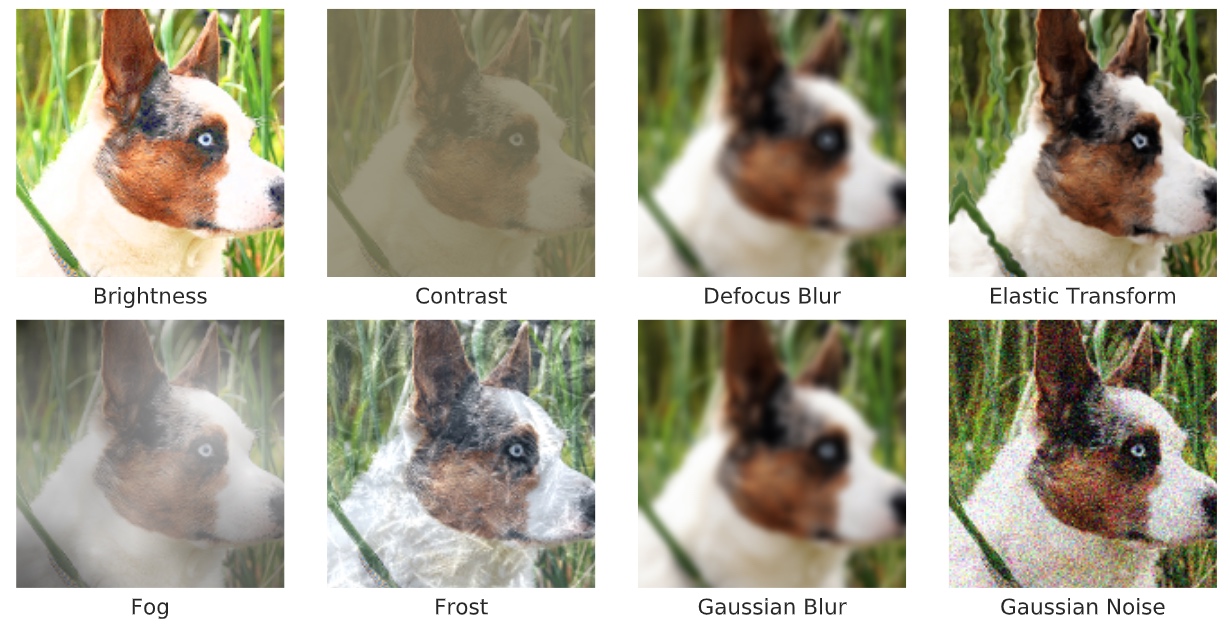}
    \caption{Examples of 16 corruption types in ImageNet-C images, at corruption intensity 3 (on a scale from 1--5). The same corruptions were applied to CIFAR-10. Figure~\ref{fig:cifar_and_imagenet_boxplots} and Section \ref{sec:calibration:skew:additional} show boxplots for each uncertainty method and corruption intensity, spanning all corruption types.}
    \label{fig:imagenet_corruptions_all}
\end{figure}

For the corrupted ImageNet dataset, we used ImageNet-C \citep{hendrycks2018benchmarking}. Figure~\ref{fig:imagenet_gaussian_blur_samples} shows examples of ImageNet-C images at varying corruption intensities. Figure~\ref{fig:imagenet_corruptions_all} shows ImageNet-C images with the 16 corruptions analyzed in this paper, at intensity 3 (on a scale of 1 to 5).

\clearpage\newpage
\section{Evaluating uncertainty under distributional shift: Additional Results}\label{sec:calibration:skew:additional}

\begin{figure}[h]
    \centering
      \includegraphics[width=0.85\linewidth]{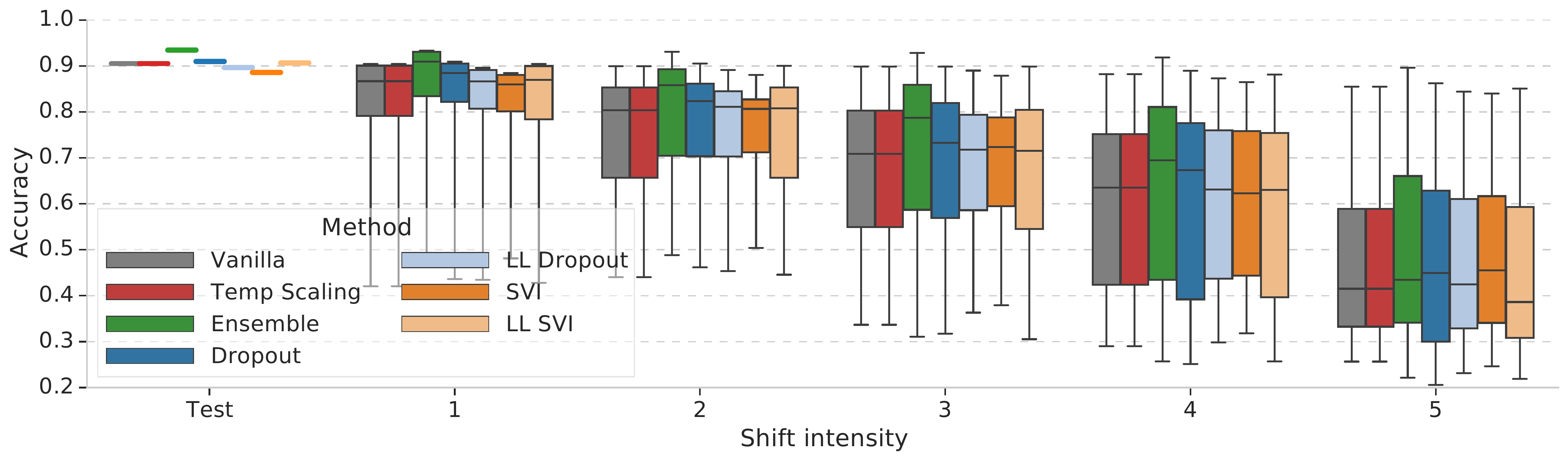}
      \includegraphics[width=0.85\linewidth]{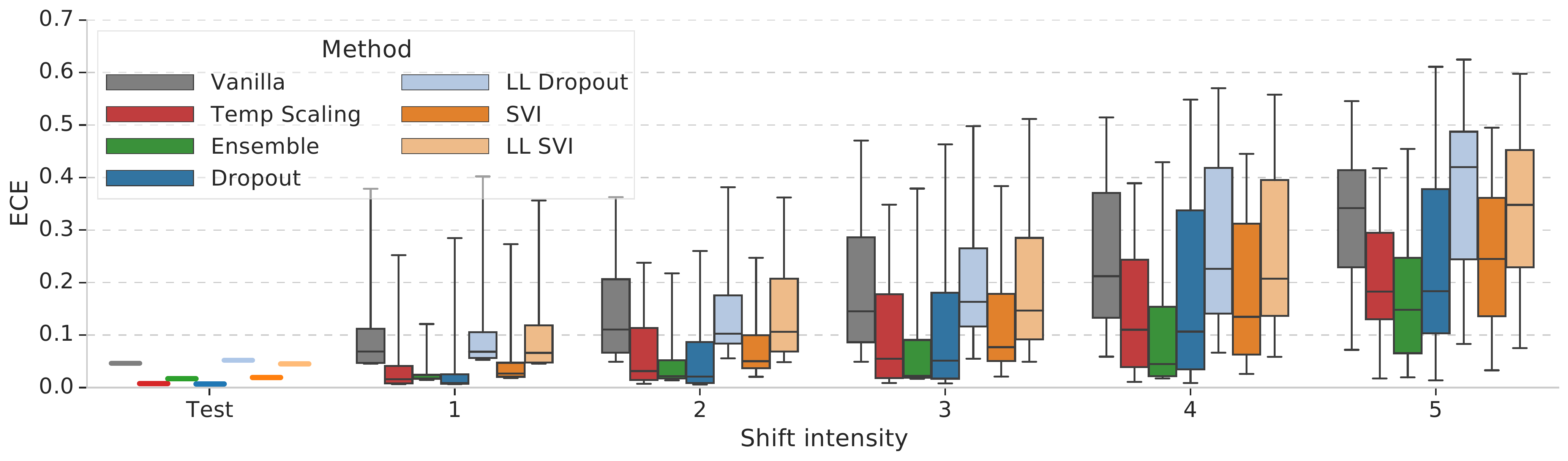}
      \includegraphics[width=0.85\linewidth]{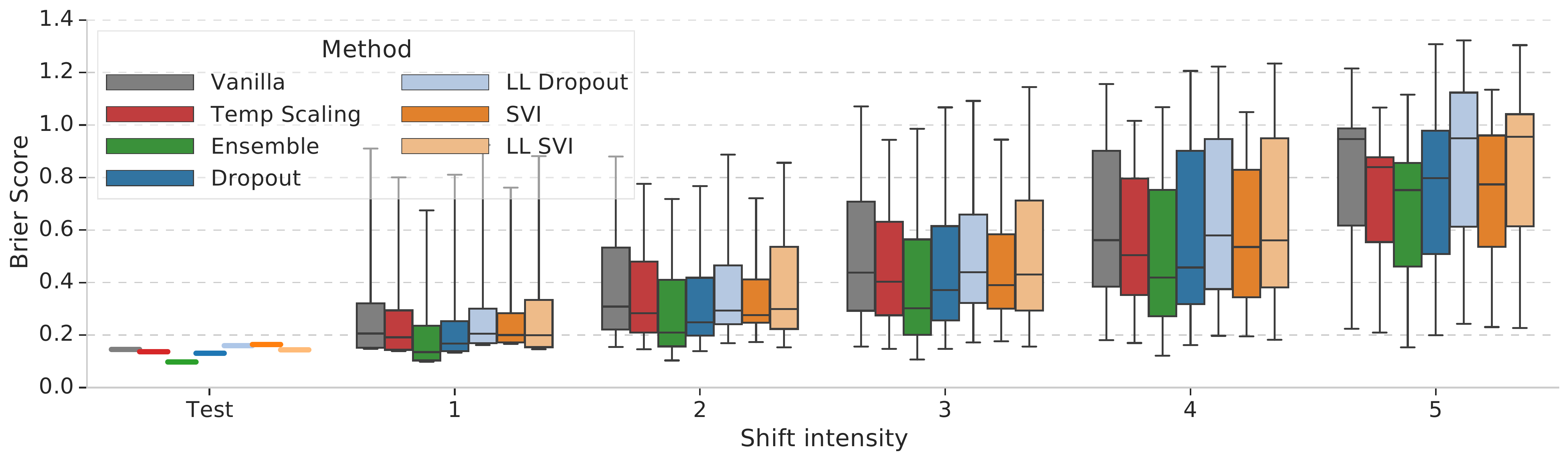}
      \includegraphics[width=0.85\linewidth]{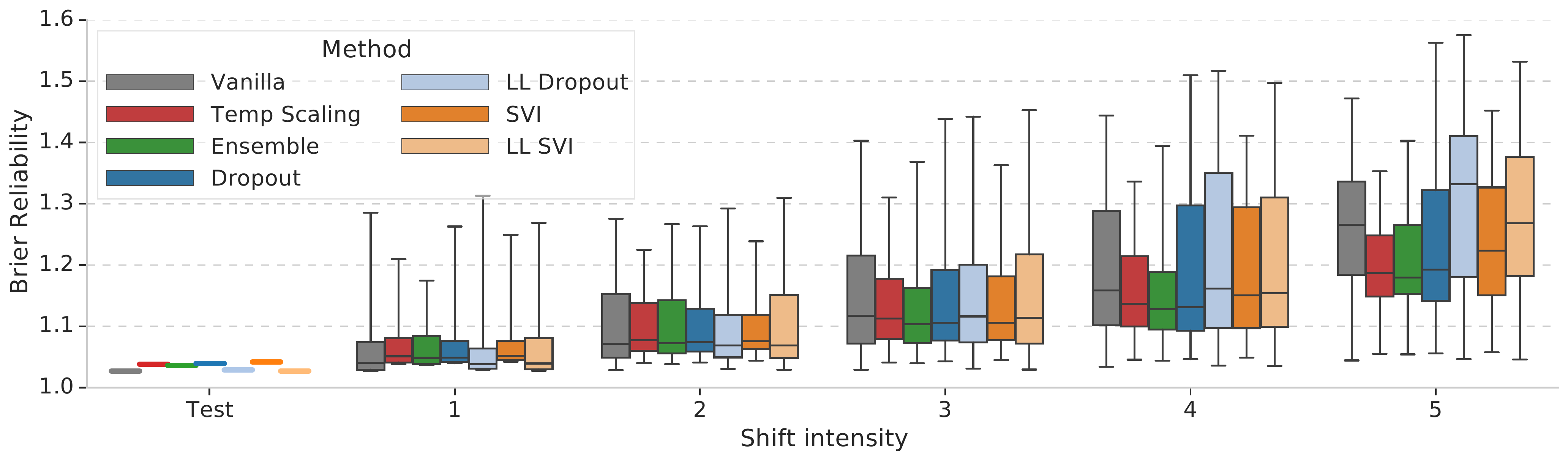}
      \includegraphics[width=0.85\linewidth]{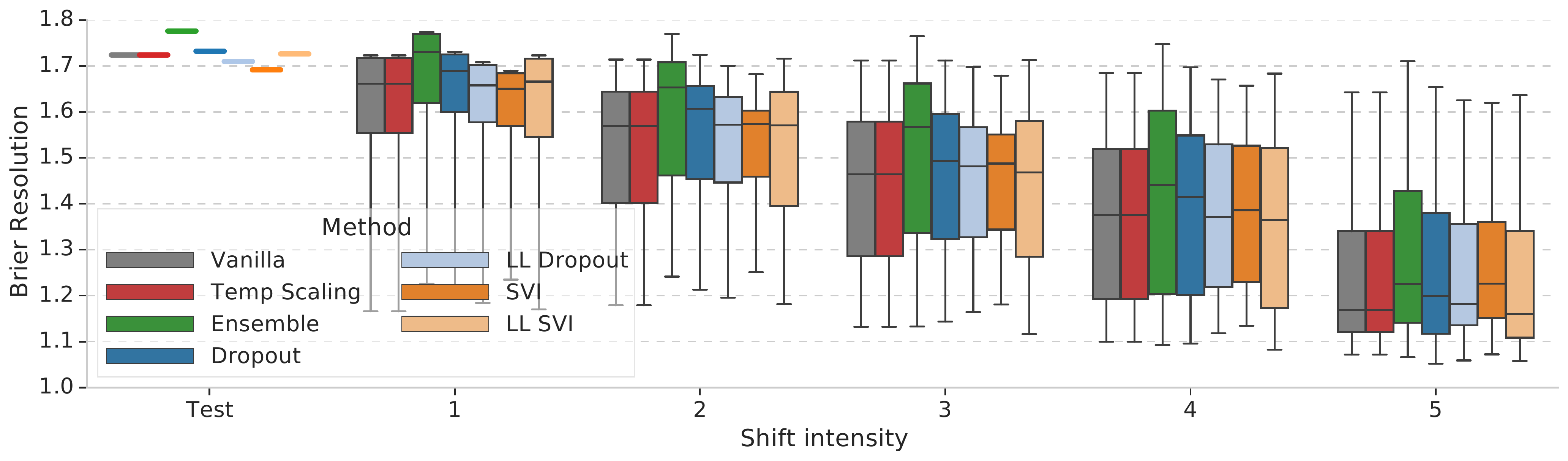}
    \caption{Boxplots facilitating comparison of methods for each shift level showing detailed comparisons of various metrics under all types of corruptions on CIFAR-10.  Each box shows the quartiles summarizing the results across all types of shift while the error bars indicate the min and max across different shift types.}
    \label{fig:cifar_boxplots}
\end{figure}

\begin{figure}[h]
    \centering
      \includegraphics[width=0.85\linewidth]{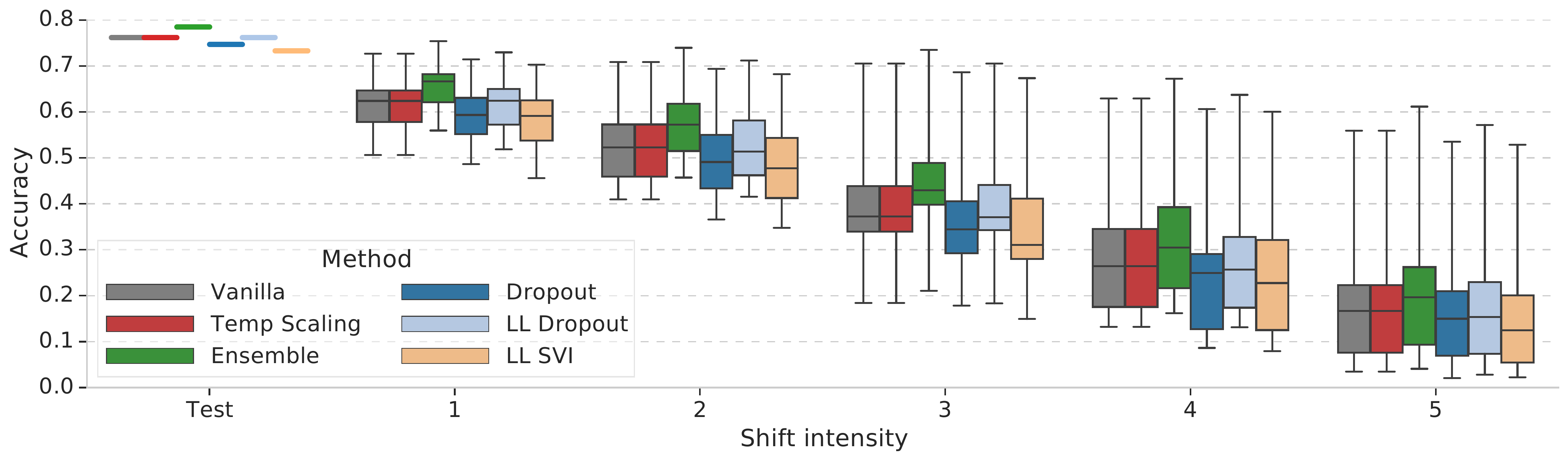}
      \includegraphics[width=0.85\linewidth]{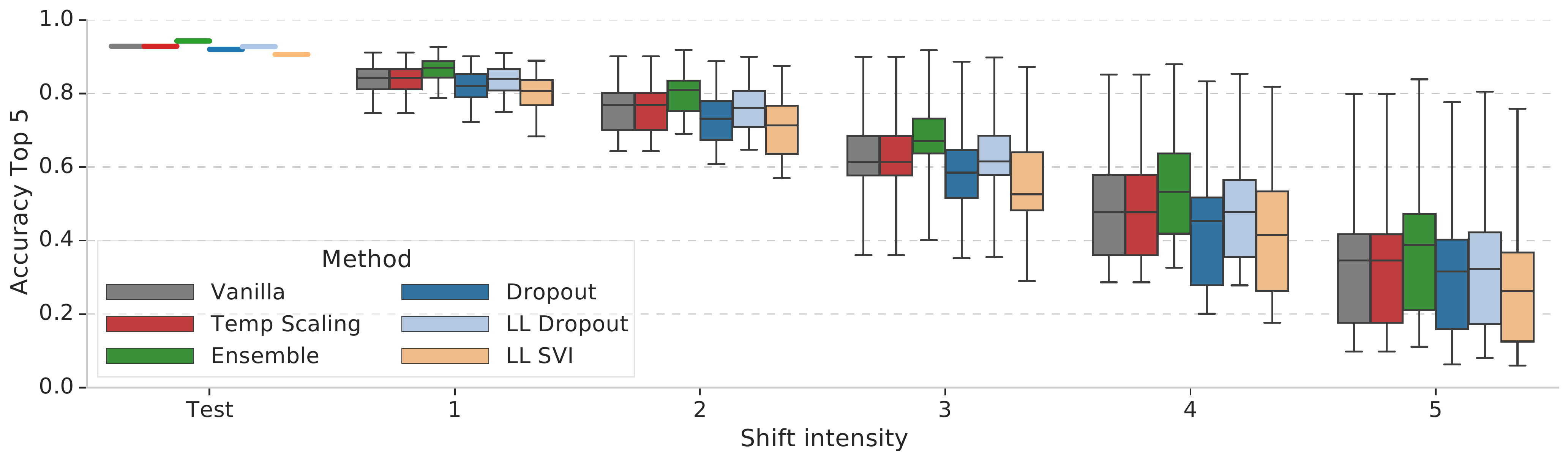}
      \includegraphics[width=0.85\linewidth]{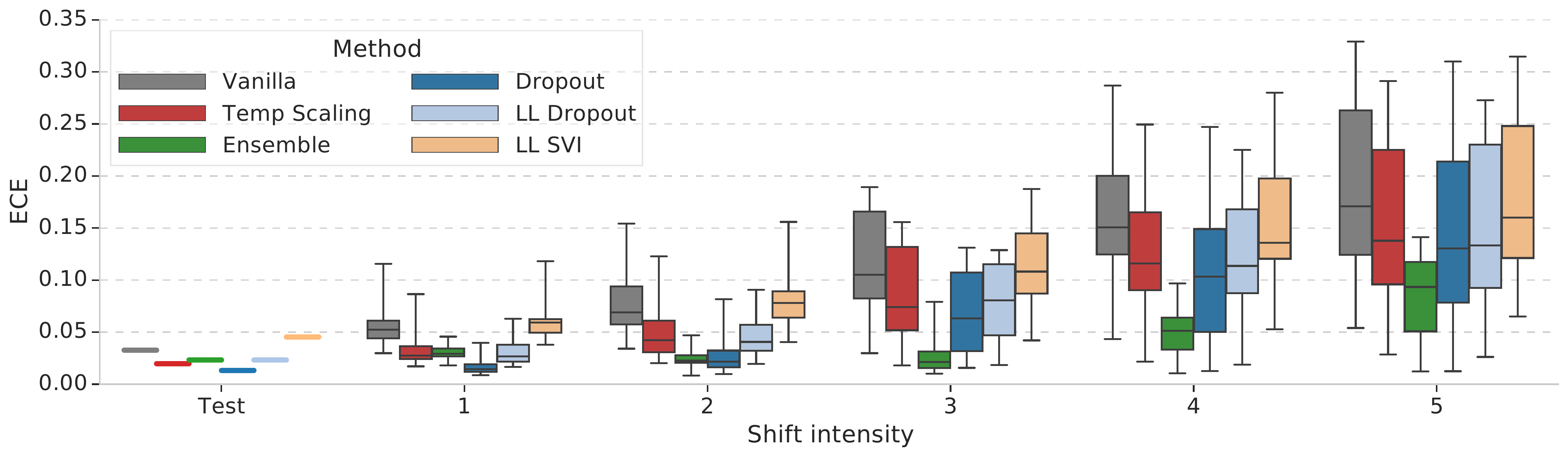}
      \includegraphics[width=0.85\linewidth]{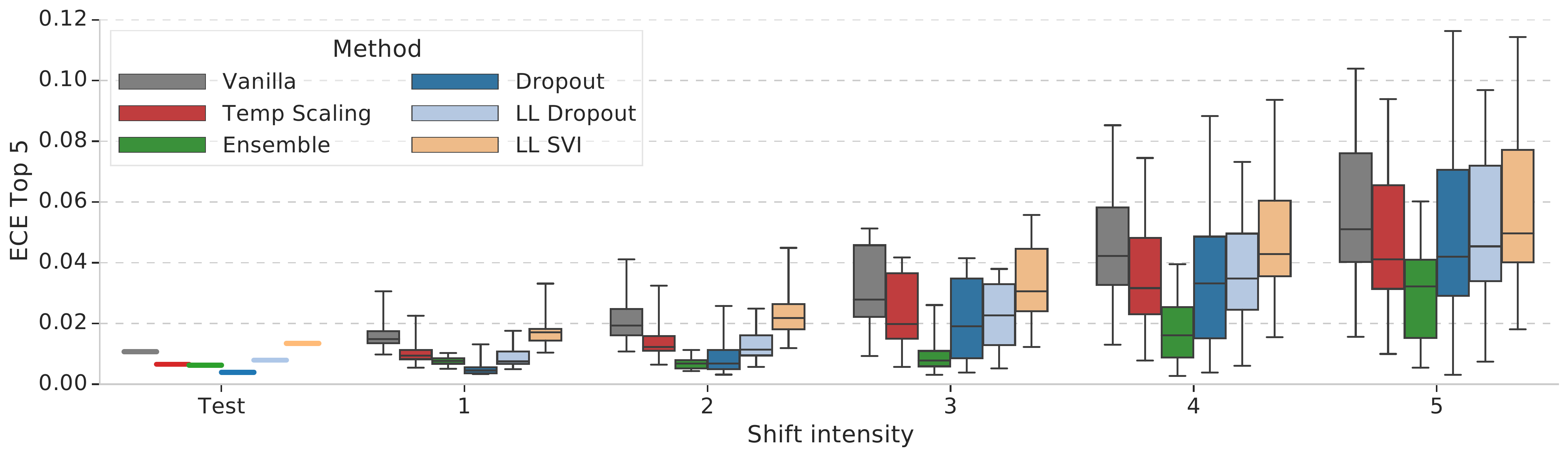}
      \includegraphics[width=0.85\linewidth]{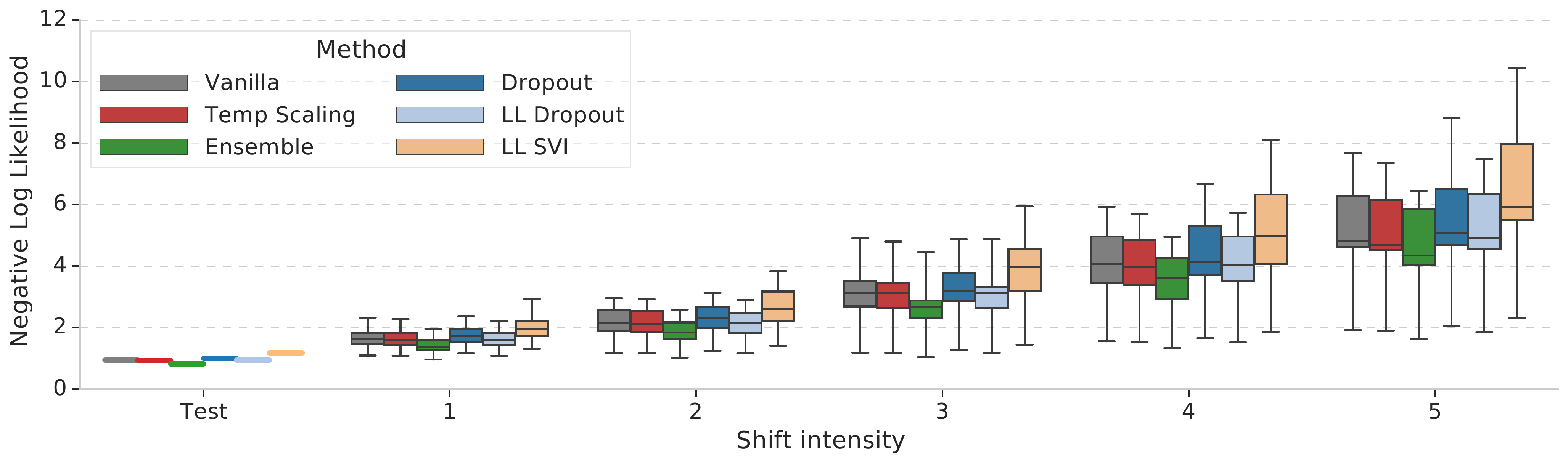}
      \includegraphics[width=0.85\linewidth]{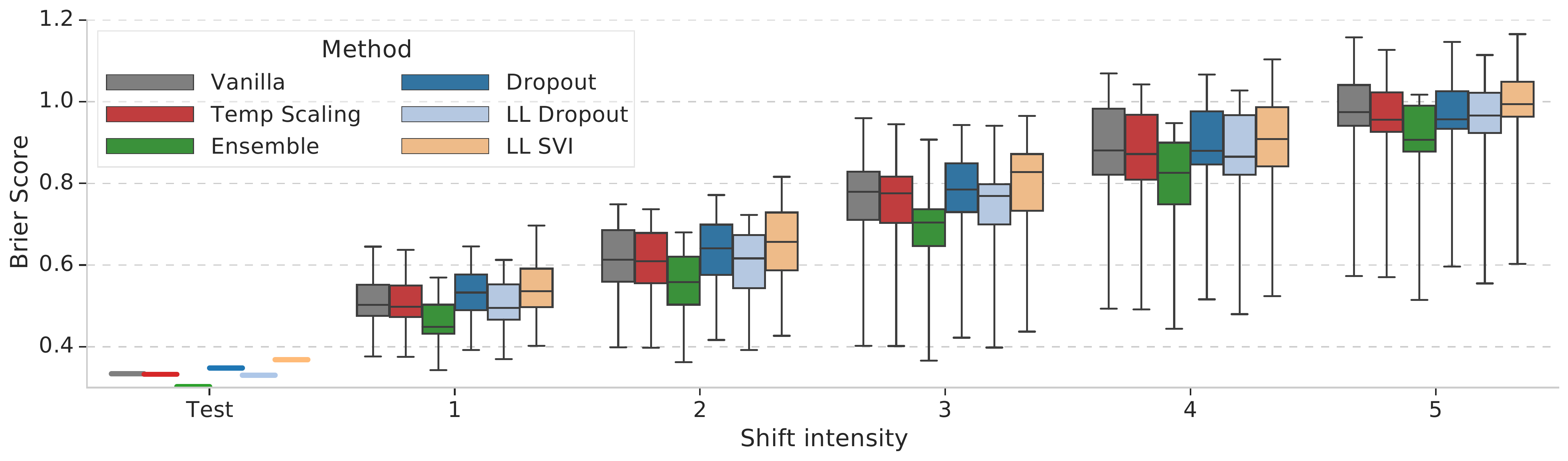}      
    \caption{Boxplots facilitating comparison of methods for each shift level showing detailed comparisons of various metrics under all types of corruptions on ImageNet.  Each box shows the quartiles summarizing the results across all types of shift while the error bars indicate the min and max across different shift types.}
    \label{fig:imagenet_boxplots}
\end{figure}

\begin{figure}[h]
    \centering
      \includegraphics[width=0.85\linewidth]{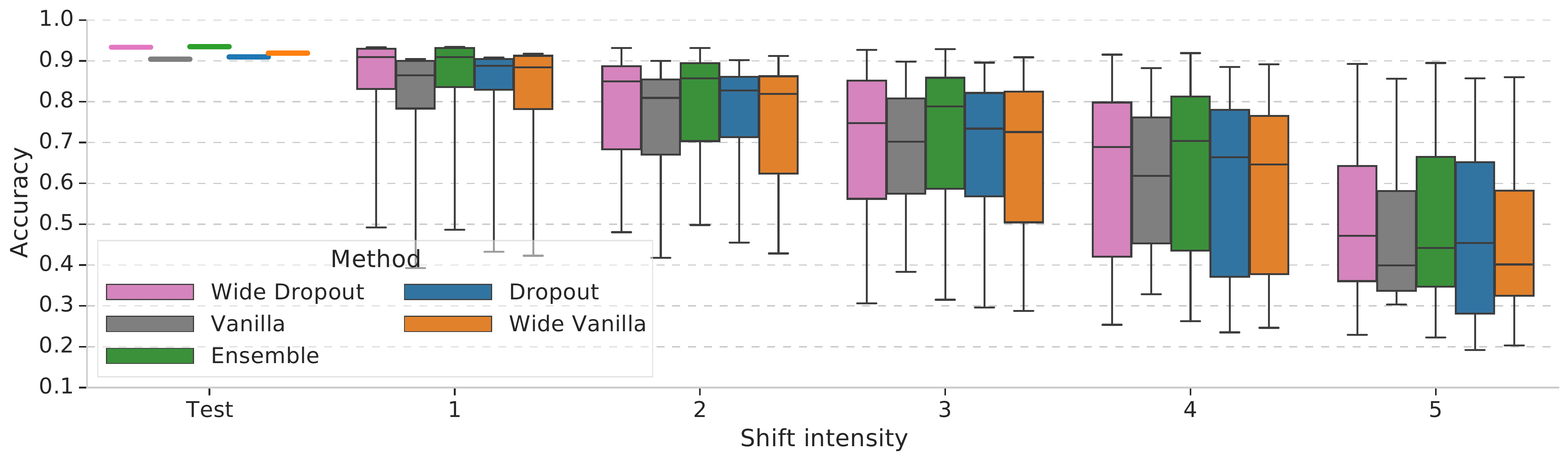}
      \includegraphics[width=0.85\linewidth]{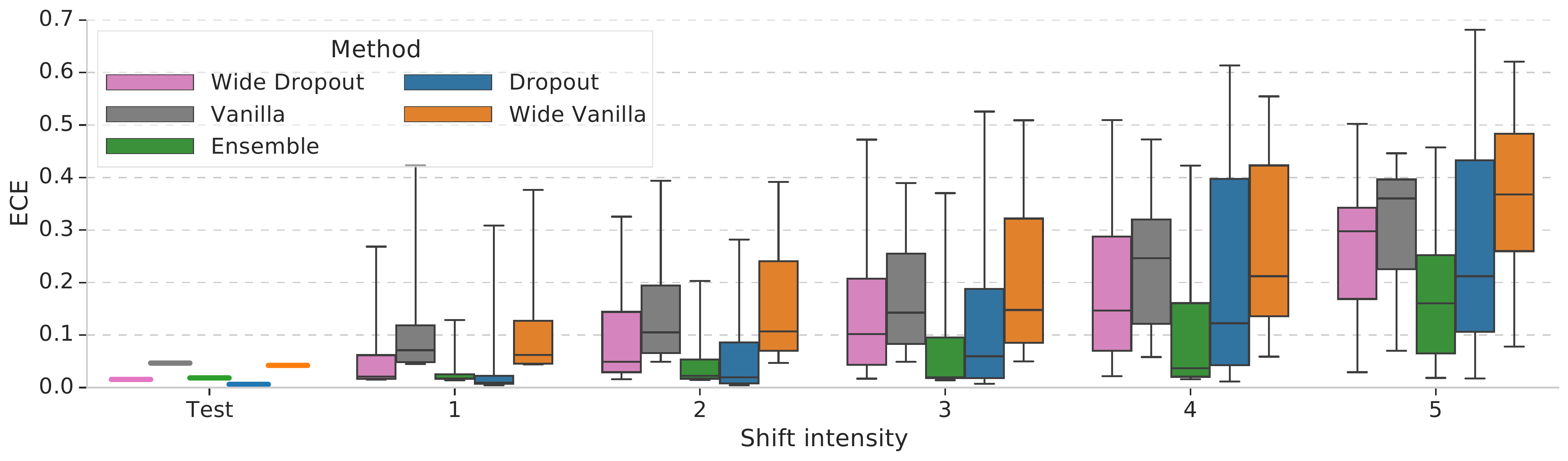}
    \caption{Boxplots facilitating comparison of results for higher-capacity models ('Wide Vanilla' and 'Wide Dropout') with their lower-capacity counterparts on CIFAR. Each box shows the quartiles summarizing the results across all types of shift while the error bars indicate the min and max across different shift types.}
    \label{fig:cifar_wide_boxplots}
\end{figure}

\begin{figure}[h]
    \centering
      \includegraphics[width=0.34\linewidth]{criteo_auc.pdf}
      \includegraphics[width=0.34\linewidth]{criteo_brier.pdf}
      \includegraphics[width=0.32\linewidth]{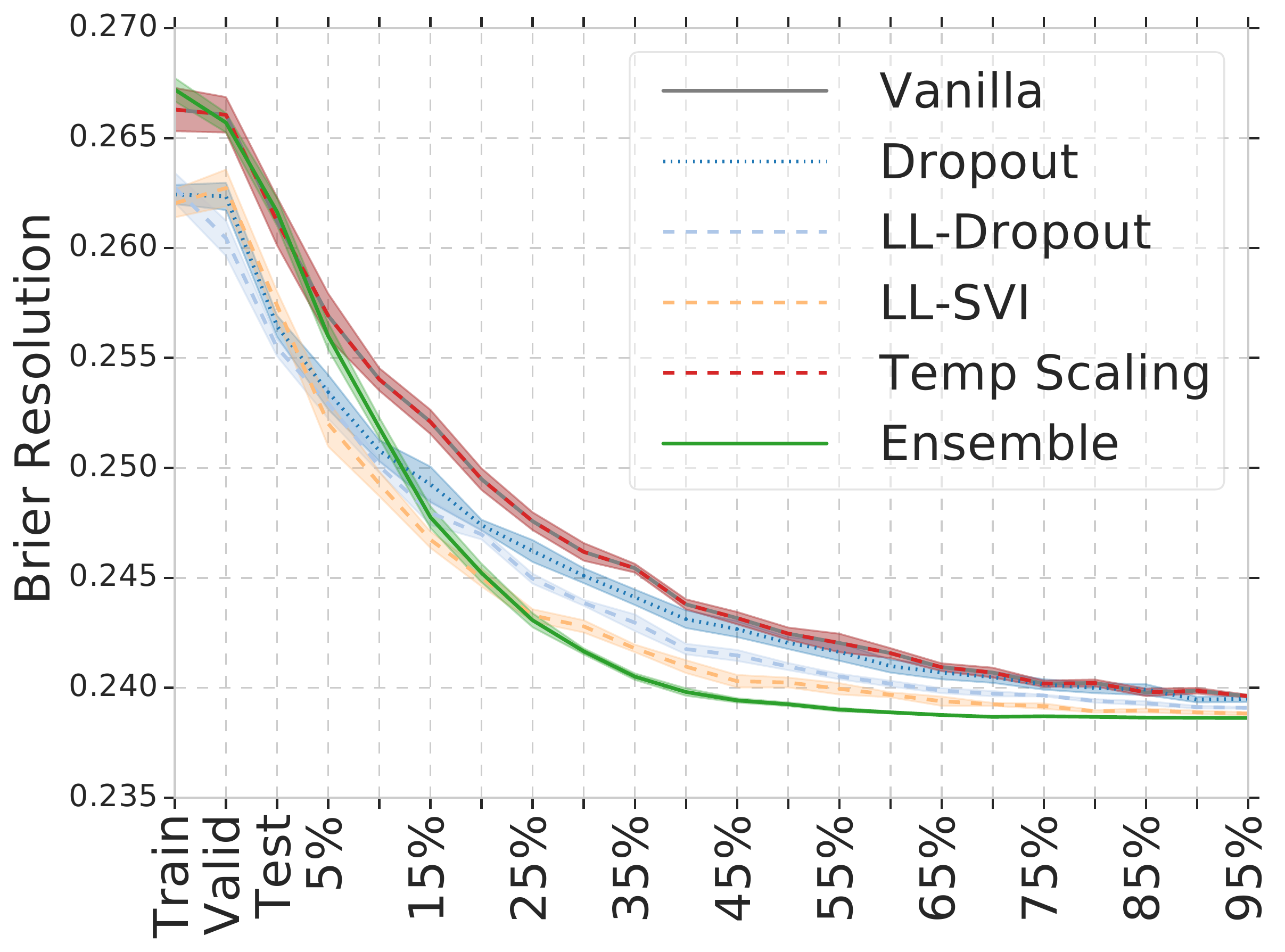}
      \includegraphics[width=0.32\linewidth]{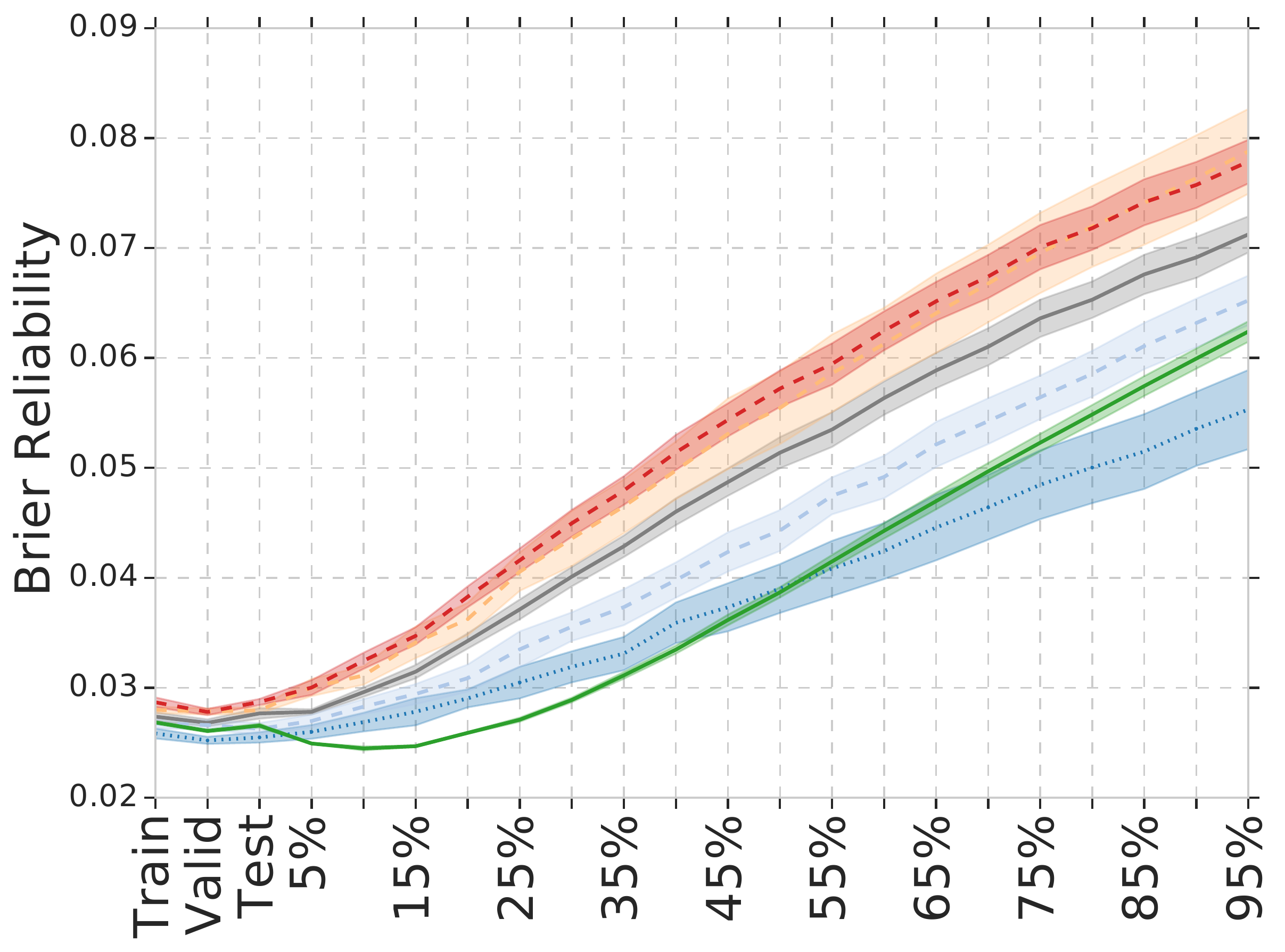}
      \includegraphics[width=0.32\linewidth]{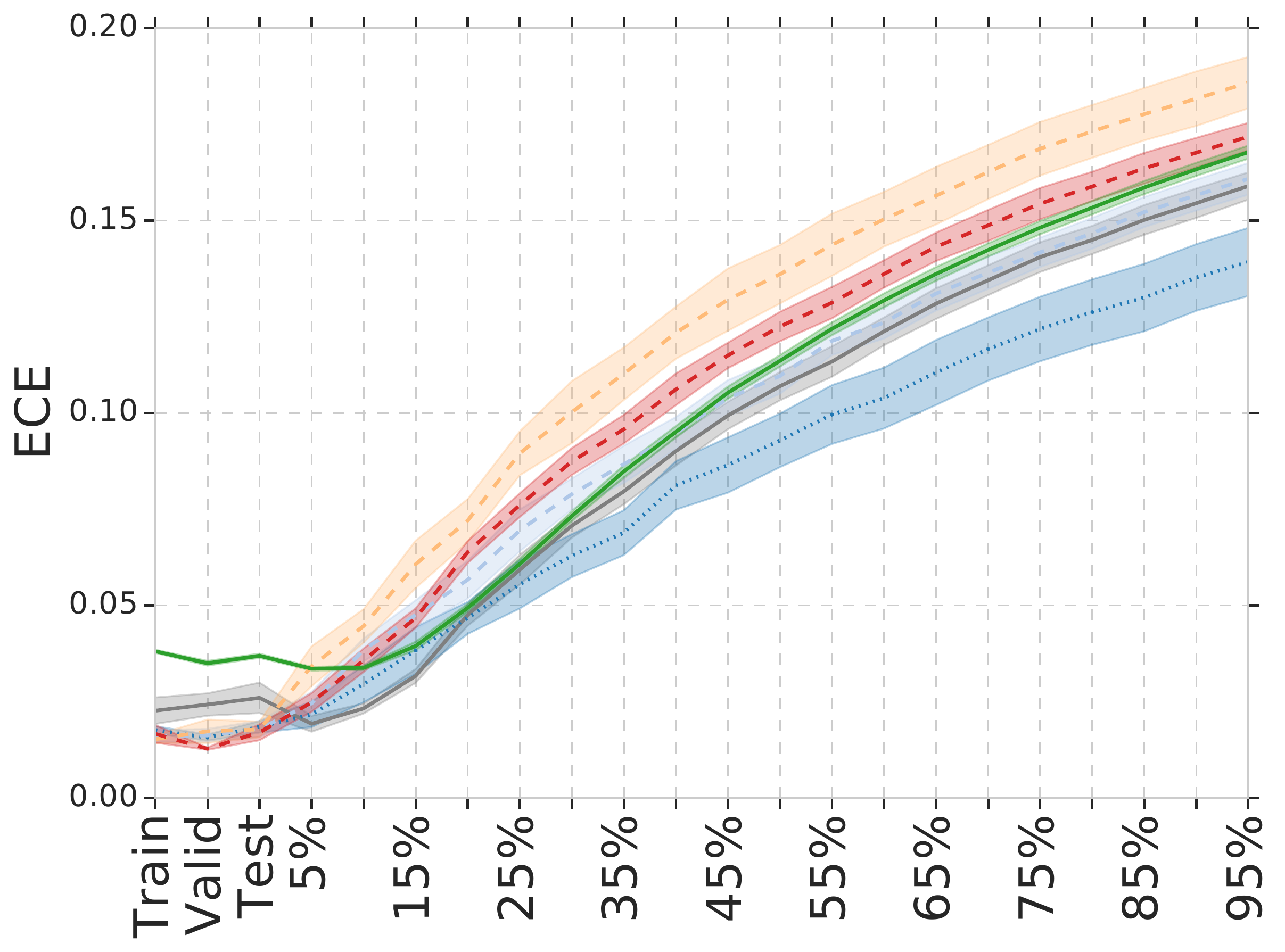}
    \caption{Comprehensive comparison of metrics on Criteo models.
    The Brier decomposition reveals that the majority of its degradation is due to worsening reliability, and this component alone appears to largely explain the ranking of methods in total Brier score.
    Ensemble notably degrades most rapidly in resolution but persists with better reliability compared other methods for most of the data-corruption range; on ECE it remains roughly in the middle among explored methods. 
    Dropout (and to a lesser extend LL-Dropout) perform best on ECE and experience slower degradation in both resolution and reliability leading it to surpass ensembles at the severe range of data corruption.
    Total Brier score and AUC results are discussed in detail in Section \ref{sec:results:criteo}.
    }
    \label{fig:criteo_addl}
\end{figure}

Figures~\ref{fig:cifar_boxplots}, \ref{fig:imagenet_boxplots} and \ref{fig:criteo_addl} show comprehensive results on CIFAR-10, ImageNet and Criteo respectively across various metrics including Brier score, along with the components of the Brier score : reliability (lower means better calibration) and resolution (higher values indicate better predictive quality). 
Ensembles and dropout outperform all other methods across corruptions, while LL SVI shows no improvement over the baseline model. Figure \ref{fig:cifar_wide_boxplots} shows accuracy and ECE for models with double the number of ResNet filters; the higher-capacity models are not better calibrated than their lower-capacity counterparts, suggesting that the good calibration performance of ensembles is not due simply to higher capacity.


\clearpage \newpage

\clearpage\newpage

\section{Effect of the number of samples on the quality of uncertainty}\label{sec:sample:size}

\begin{figure}[h]
    \centering
    \begin{subfigure}[Dropout]{
      \includegraphics[width=0.475\linewidth]{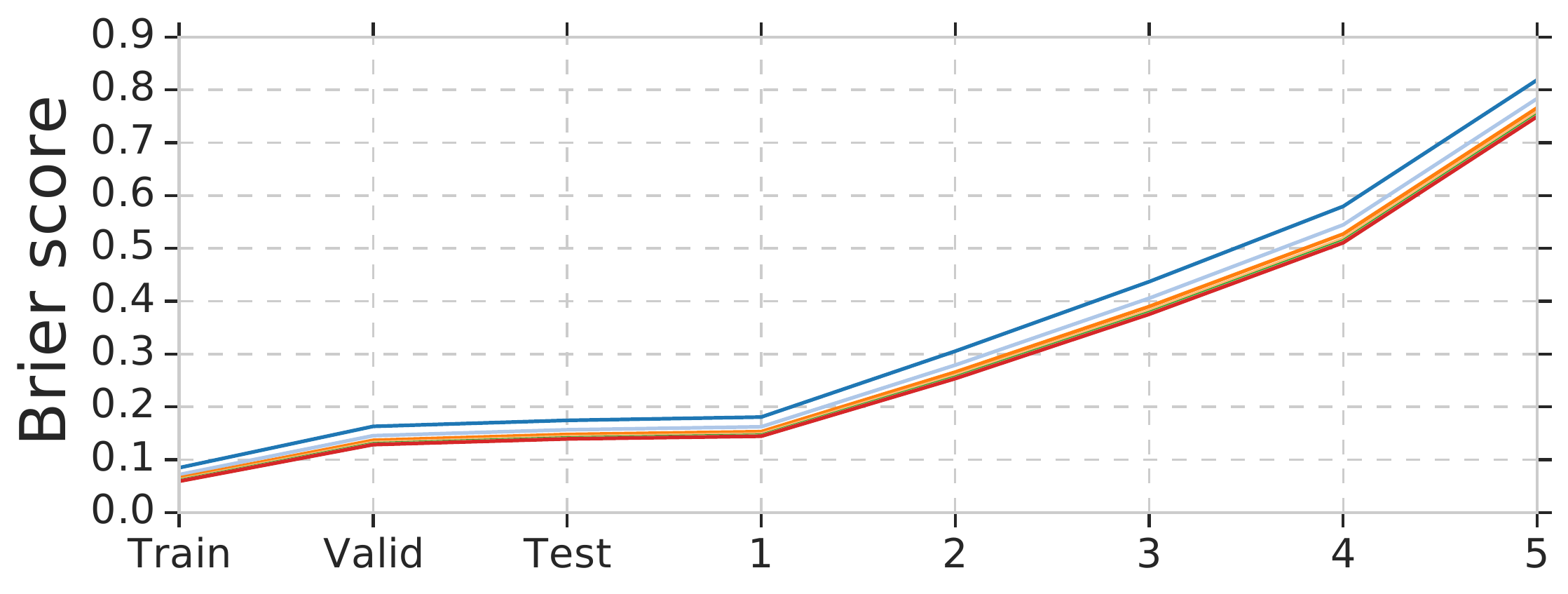}
    }\end{subfigure}
    \begin{subfigure}[LL-Dropout]{
      \includegraphics[width=0.475\linewidth]{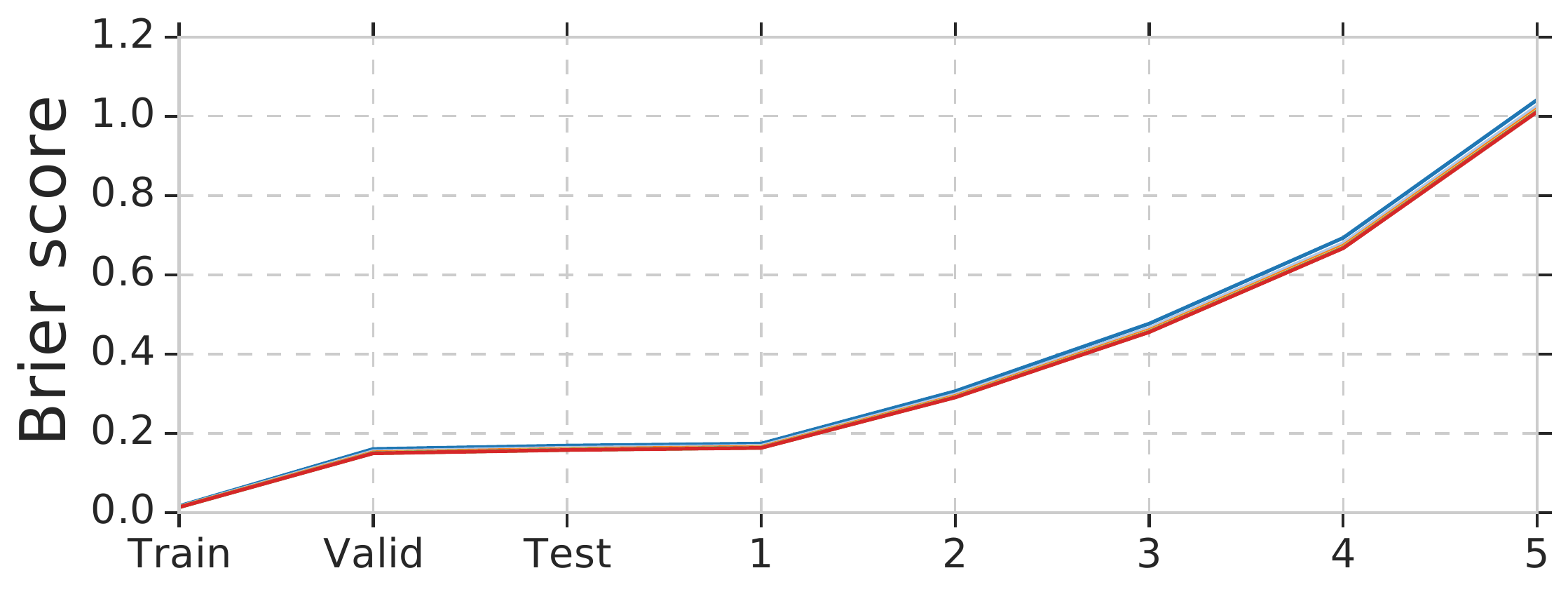}
    }\end{subfigure}
     \begin{subfigure}[SVI]{
      \includegraphics[width=0.475\linewidth]{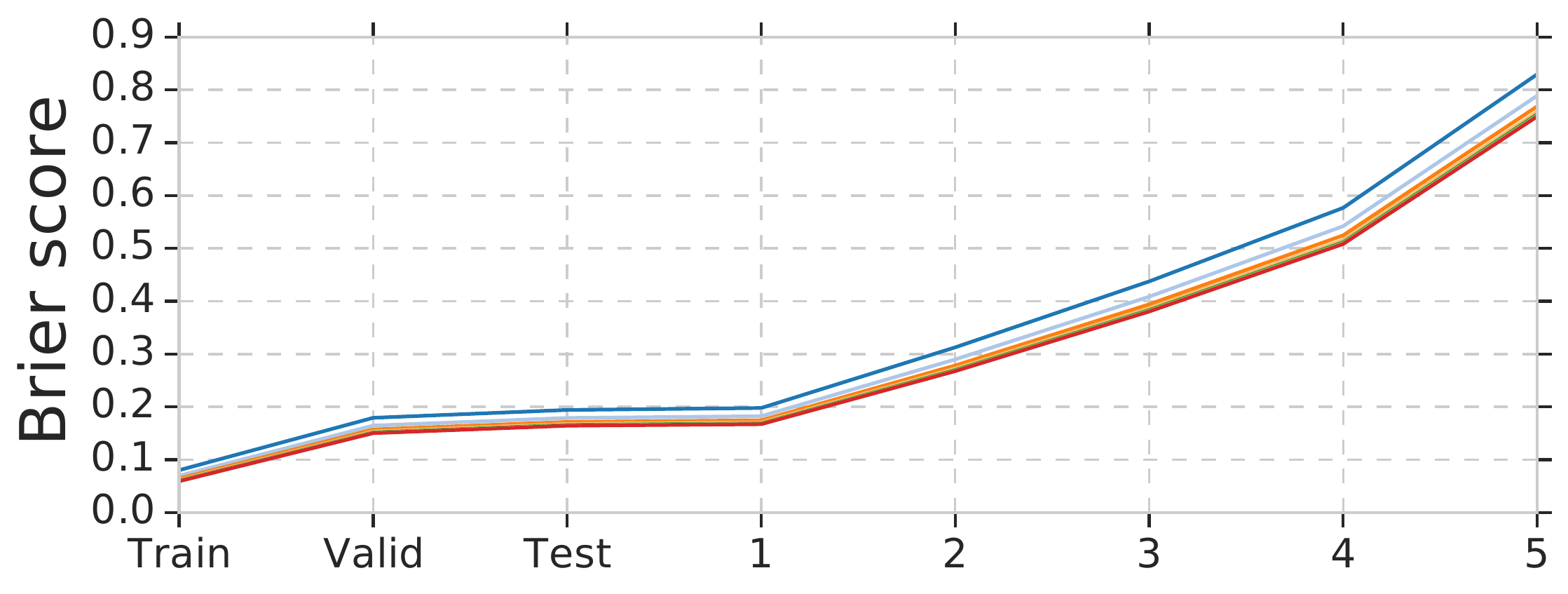}
    }\end{subfigure}
     \begin{subfigure}[LL-SVI]{
      \includegraphics[width=0.475\linewidth]{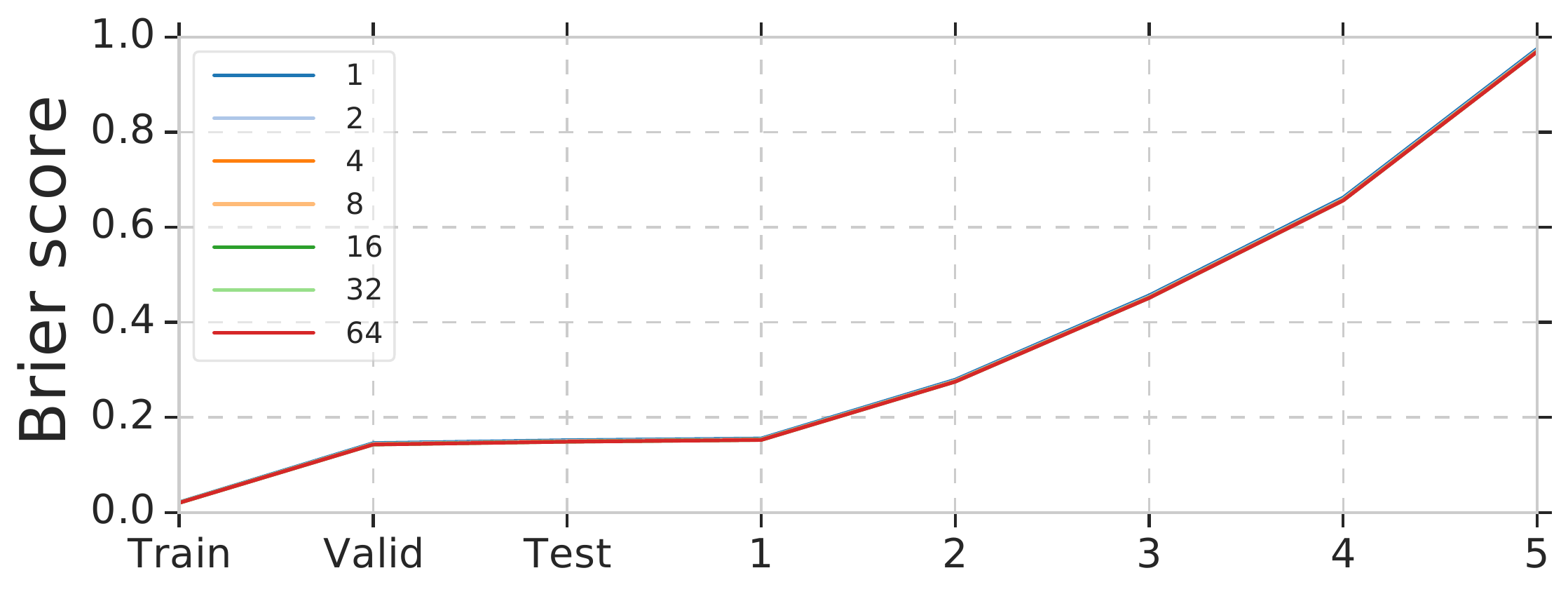}
    }\end{subfigure}
    \caption{Effect of Dropout and SVI sample sizes on CIFAR-10 Brier scores under increasing 
    Gaussian blur. See Section~\ref{sec:cifar_imagenet_results} for full results on CIFAR-10.
    }
    \label{fig:cifar_nsamples}
\end{figure}

Figure~\ref{fig:cifar_nsamples} shows the effect of the number of sample sizes used by Dropout, SVI (and last-layer variants) on the quality of predictive uncertainty, as measured by the Brier score. Increasing the number of samples has little effect on last-layer variants, whereas increasing the number of samples improves the performance for SVI and Dropout, with diminishing returns beyond size 5.

\begin{figure}[h]
    \centering
    \begin{subfigure}[Brier Score]{
      \includegraphics[width=0.31\linewidth]{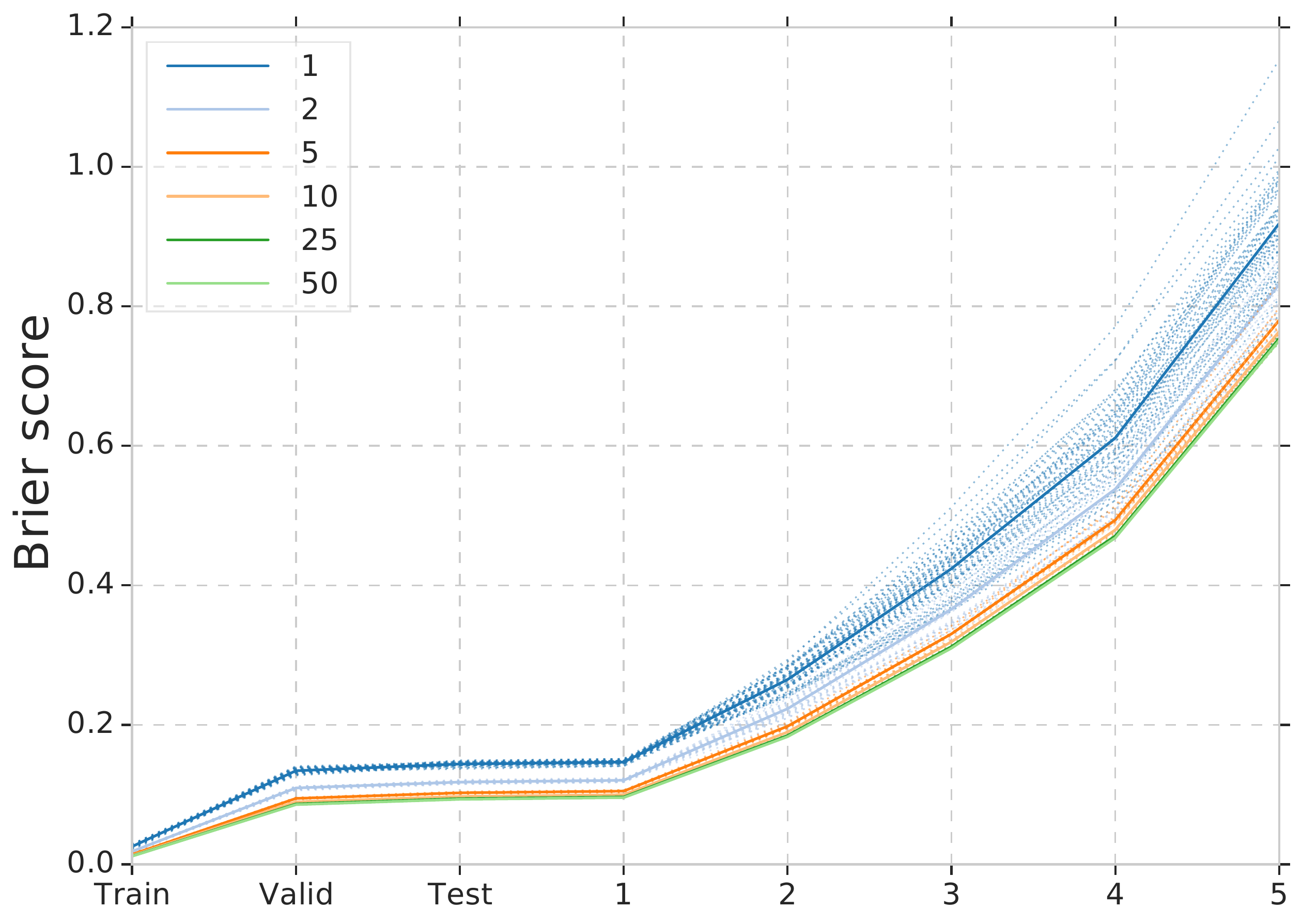}
    }\end{subfigure}
    \begin{subfigure}[Brier Reliability]{
      \includegraphics[width=0.31\linewidth]{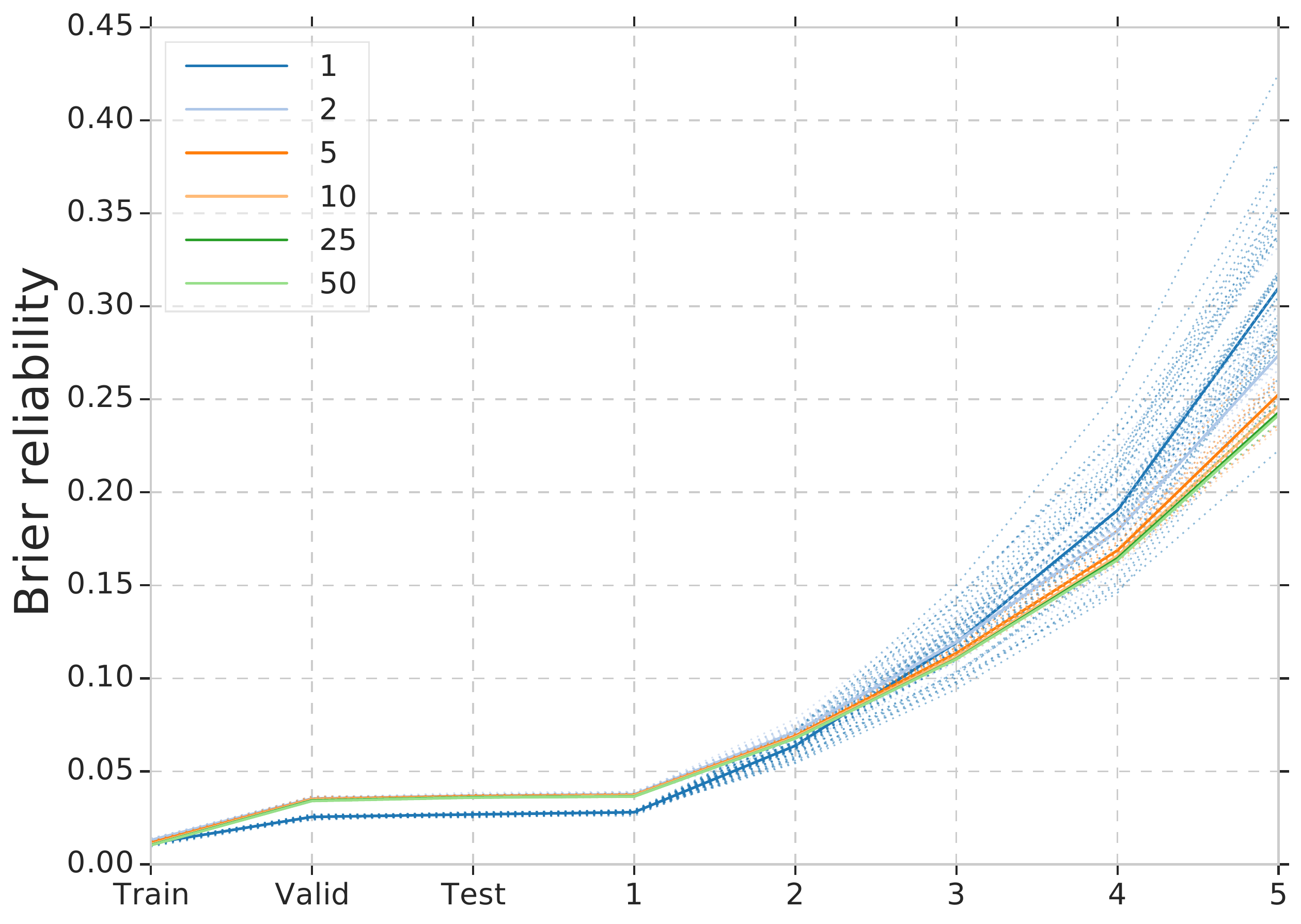}
    }\end{subfigure}
     \begin{subfigure}[Brier Resolution]{
      \includegraphics[width=0.31\linewidth]{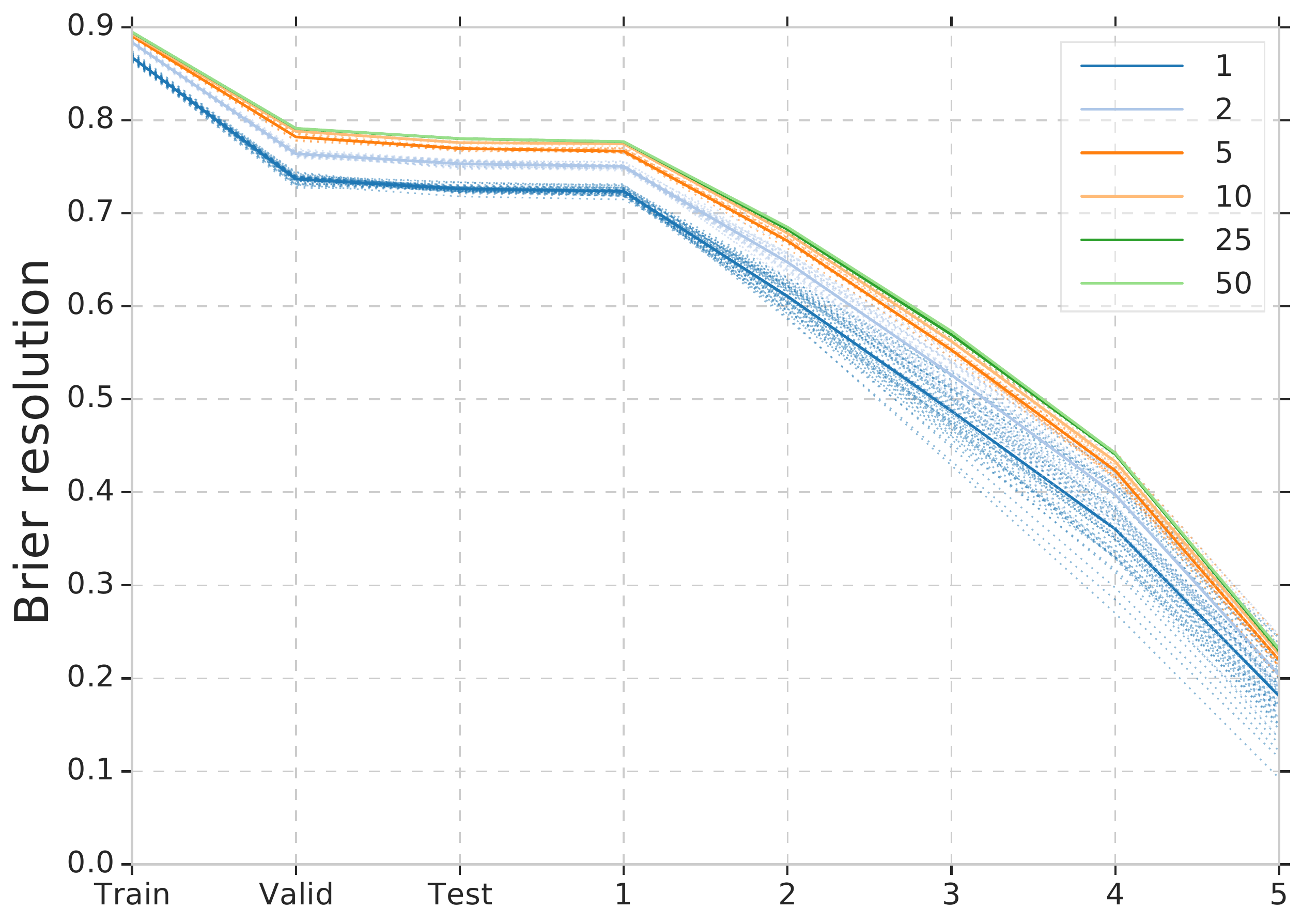}
    }\end{subfigure} \\ 
    \begin{subfigure}[Brier Score]{
      \includegraphics[width=0.31\linewidth]{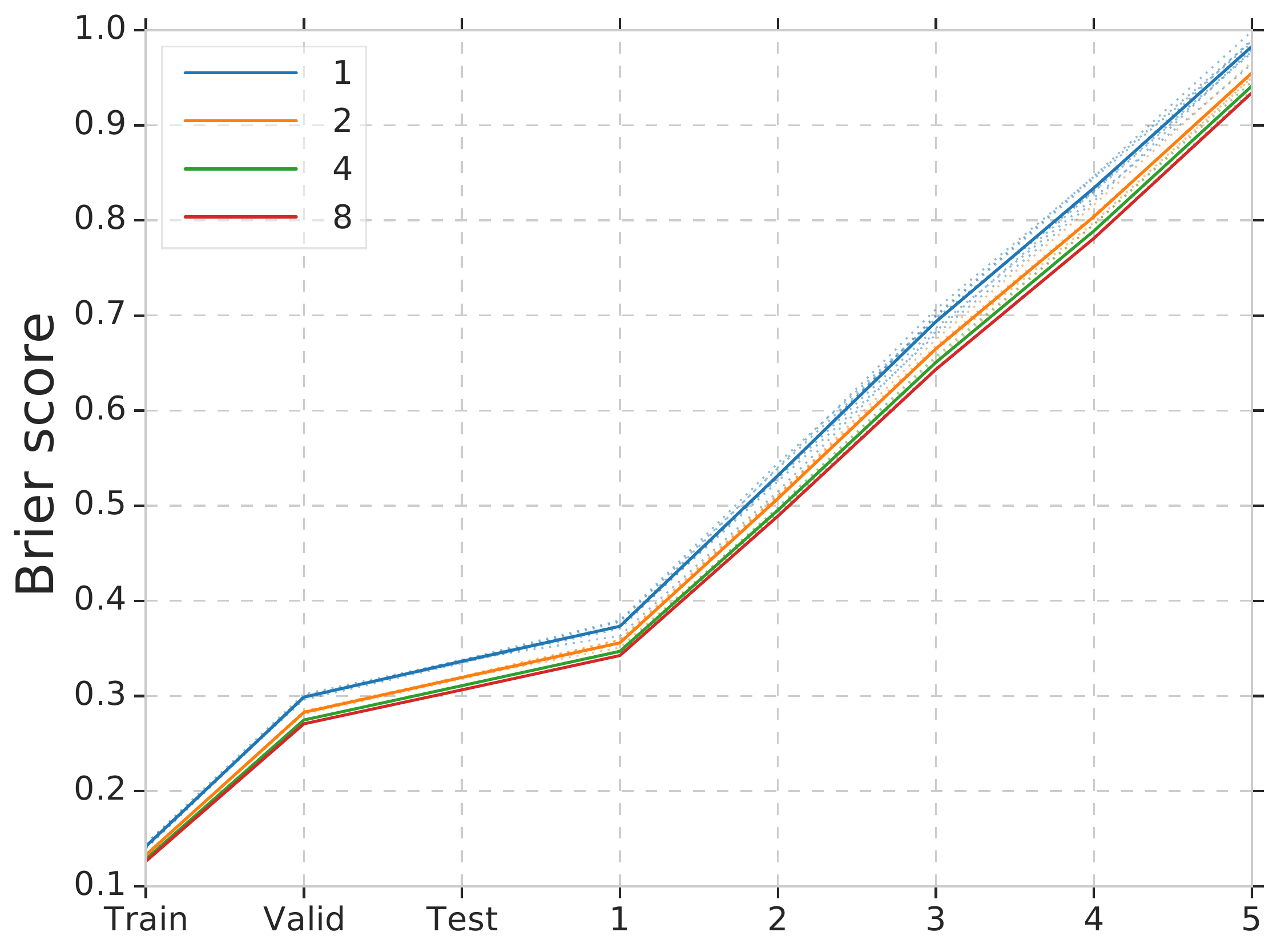}
    }\end{subfigure}
    \begin{subfigure}[Brier Reliability]{
      \includegraphics[width=0.31\linewidth]{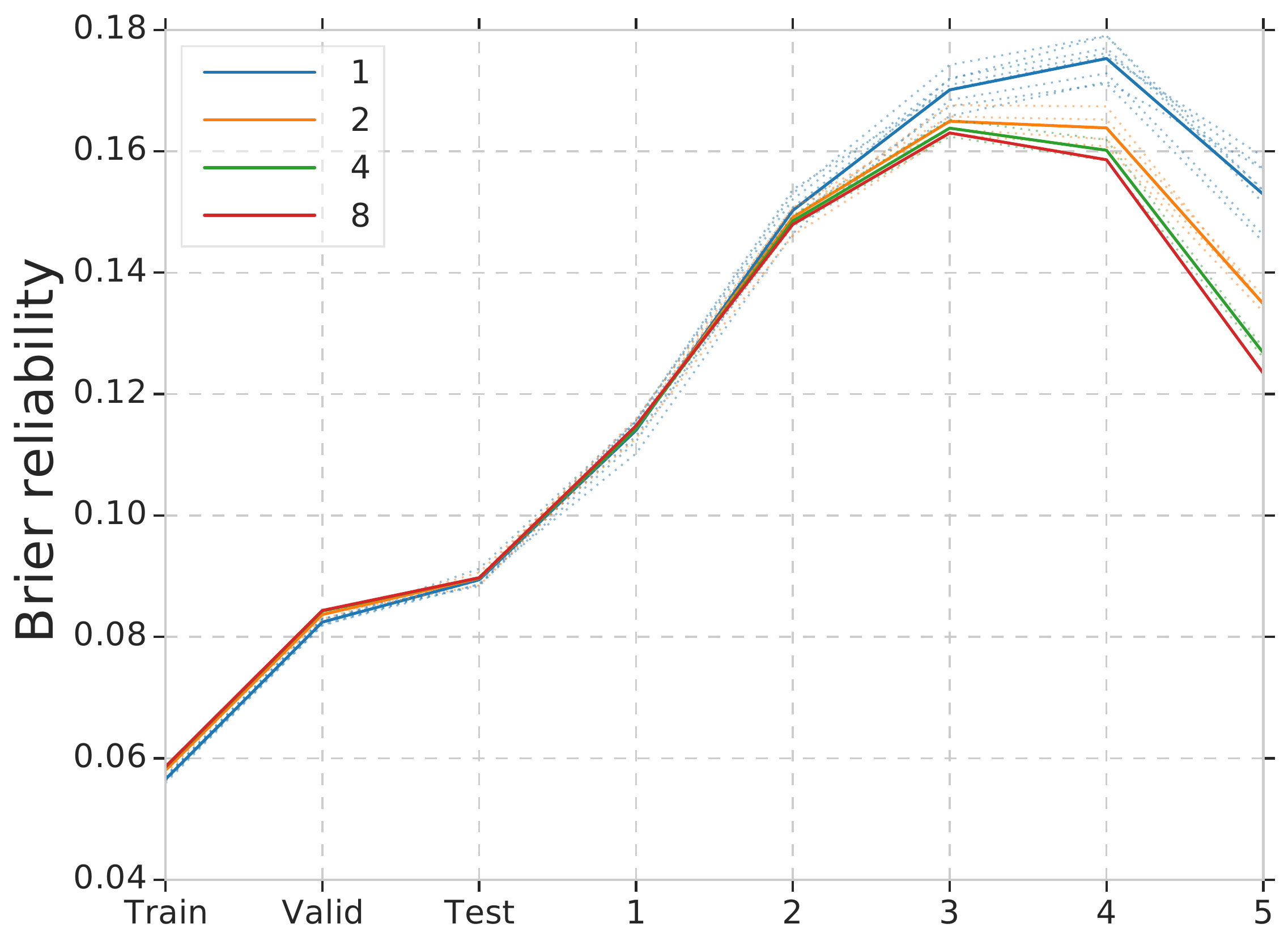}
    }\end{subfigure}
     \begin{subfigure}[Brier Resolution]{
      \includegraphics[width=0.31\linewidth]{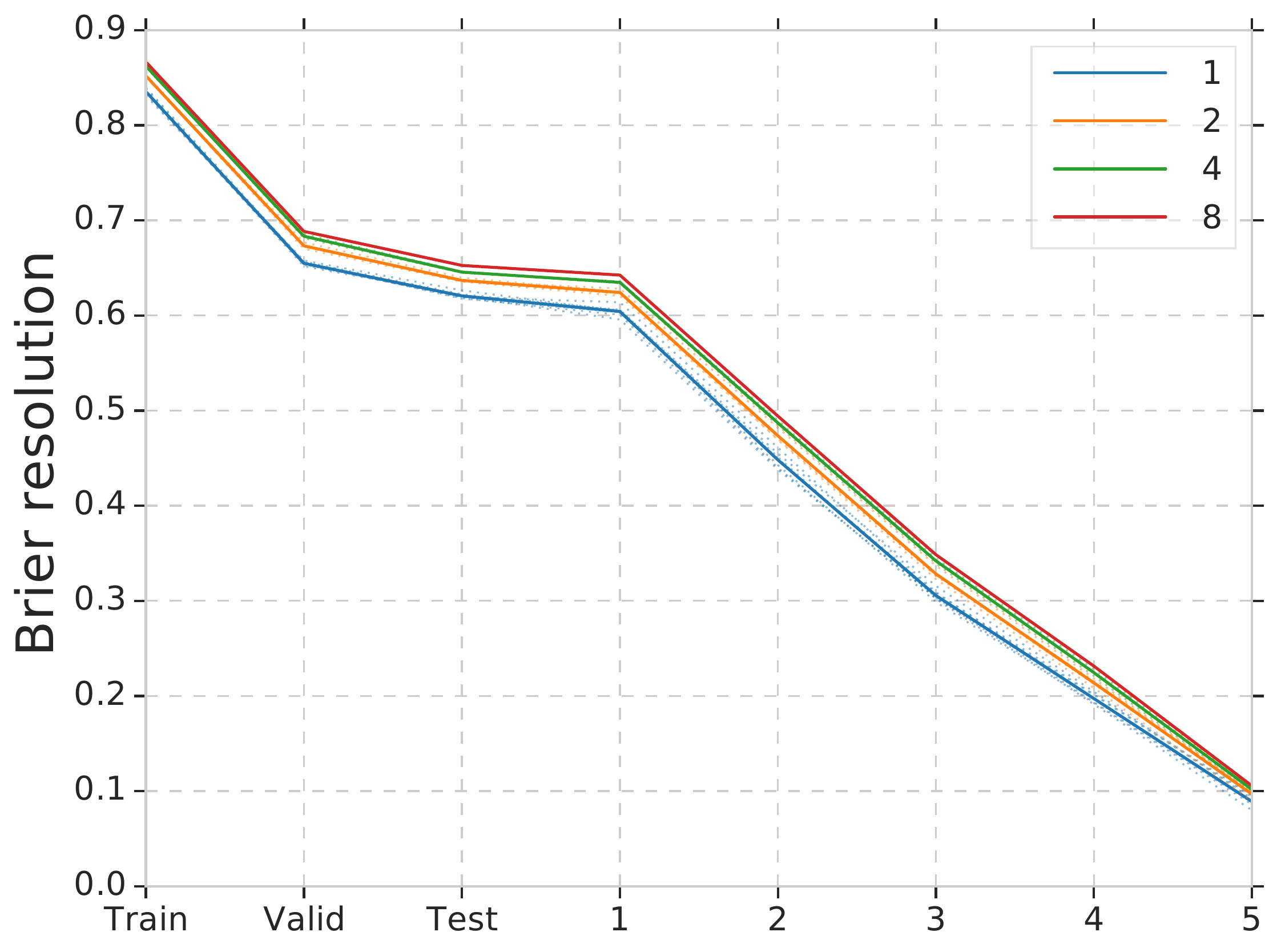}
    }\end{subfigure}
    \caption{Effect of the ensemble size on CIFAR-10 (top row) and ImageNet (bottom row) Brier scores under increasing Gaussian-blur shift.
    We additionally show the Brier score components: Reliability (lower means better calibration) and Resolution (higher values indicate better predictive quality).
    Note that the scales for Reliability are significantly smaller than the other plots.
    }
    \label{fig:ensemble_size}
\end{figure}

Figure~\ref{fig:ensemble_size} shows the effect of ensemble size on CIFAR-10 (top) and ImageNet (bottom). Similar to SVI and Dropout, we see that increasing the number of models in the ensemble improves performance with diminishing returns beyond size 5. 
As mentioned earlier, the Brier score can be further decomposed into  $\mathrm{BS} = \mathrm{calibration} + \mathrm{refinement} = \mathrm{reliability} + \mathrm{uncertainty} - \mathrm{resolution}$ where
$\mathrm{reliability} \downarrow$ measures calibration as the average violation of long-term true label frequencies, and $\mathrm{refinement}=\mathrm{uncertainty} - \mathrm{resolution}$, where  
$\mathrm{uncertainty}$ is the marginal uncertainty over labels (independent of predictions) and $\mathrm{resolution} \uparrow$ measures the deviation of individual predictions from the marginal. 

\section{Variational Gaussian Process Results}
\label{sec:vgp}
\begin{figure}[h]
    \centering
    \begin{subfigure}[Brier Score]{
      \includegraphics[width=0.85\linewidth]{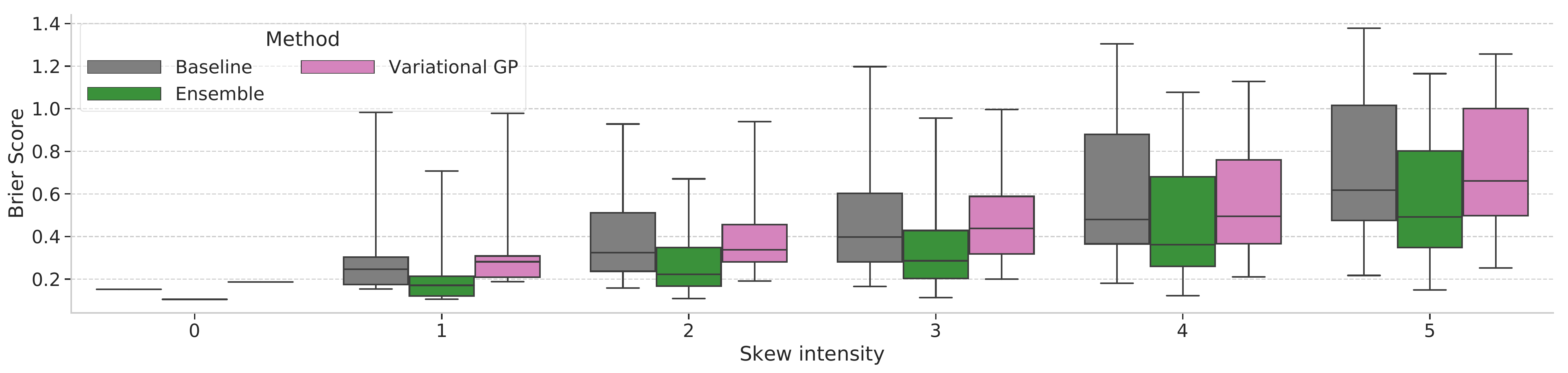}
    }\end{subfigure} \\
    \begin{subfigure}[Accuracy]{
      \includegraphics[width=0.85\linewidth]{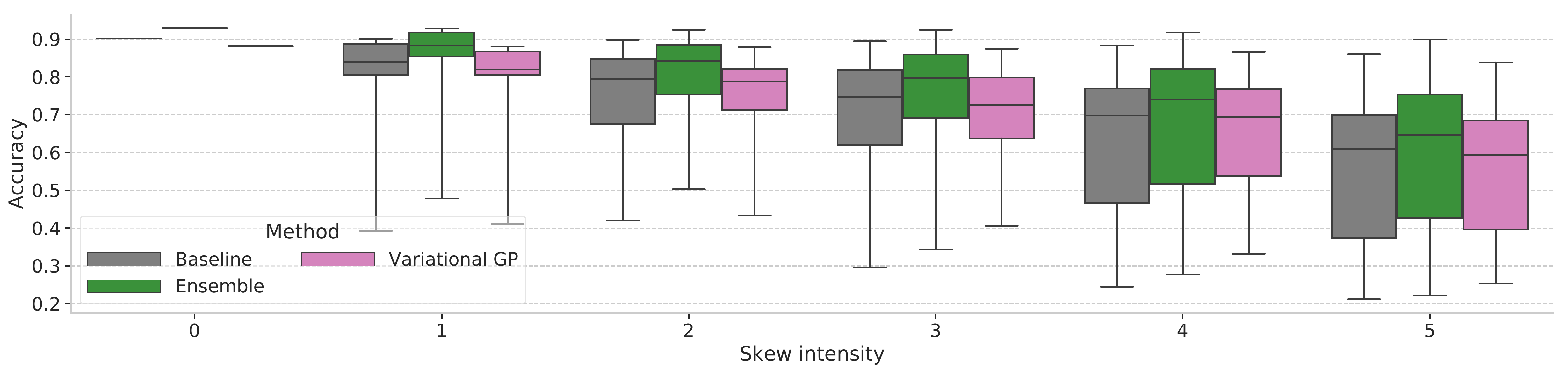}
    }\end{subfigure} \\
     \begin{subfigure}[ECE]{
      \includegraphics[width=0.85\linewidth]{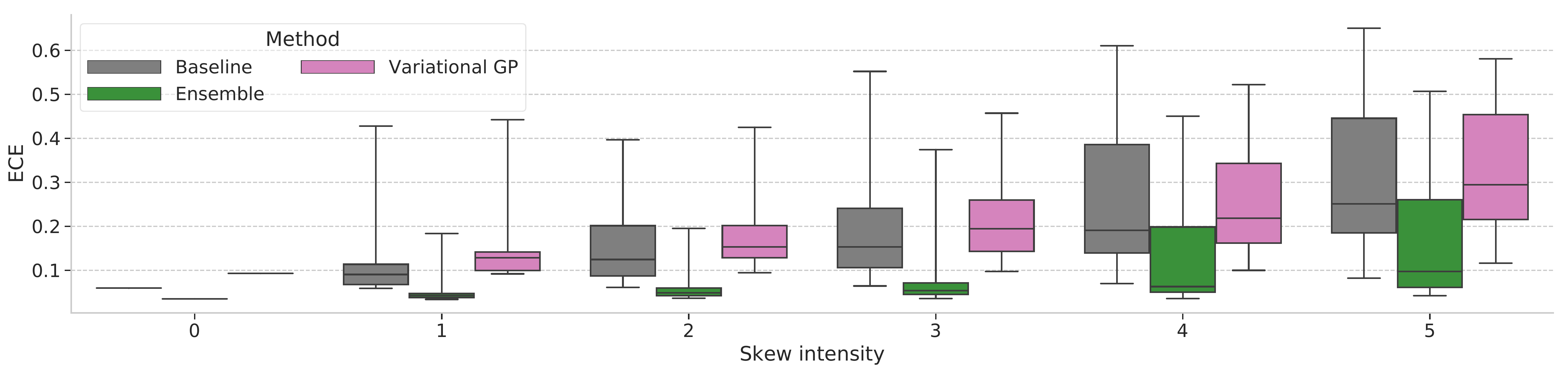}
    }\end{subfigure} \\
    \caption{Uncertainty metrics across shift levels on CIFAR-10, where level 0 is the test set, using a last layer Variational Gaussian Process. See \textbf{Appendix~\ref{sec:gaussian_processes}} for experiment details.}
    \label{fig:vgp_skew}
\end{figure}
\clearpage

\section{OOD detection for genomic sequences}
\label{sec:gene}

We studied the set of methods for detecting OOD genomic sequence, as a challenging realistic problem for OOD detection proposed by \citet{ren2019likelihood}. Classifiers are trained on 10 in-distribution bacteria classes, and tested for OOD detection of 60 OOD bacteria classes. The model architecture is the same as that in \cite{ren2019likelihood}: a convolutional neural networks with 1000 filters of length 20, followed by a global max pooling layer, a dense layer of 1000 units, and a last dense layer that outputs class prediction logits. For the dropout method, we add a dropout layer each after the max pooling layer and the dense layer respectively. For the LL-Dropout method, only a dropout layer after the dense layer is added. We use the dropout rate of 0.2. For the LL-SVI method, we replace the last dense layer with a stochastic variational inference dense layer. The classification accuracy for in-distribution is around 0.8 for the various types of classifiers.   

\begin{figure}[h]%
    \centering%
    \begin{subfigure}[Confidence vs Accuracy]{
      \includegraphics[width=0.5\linewidth]{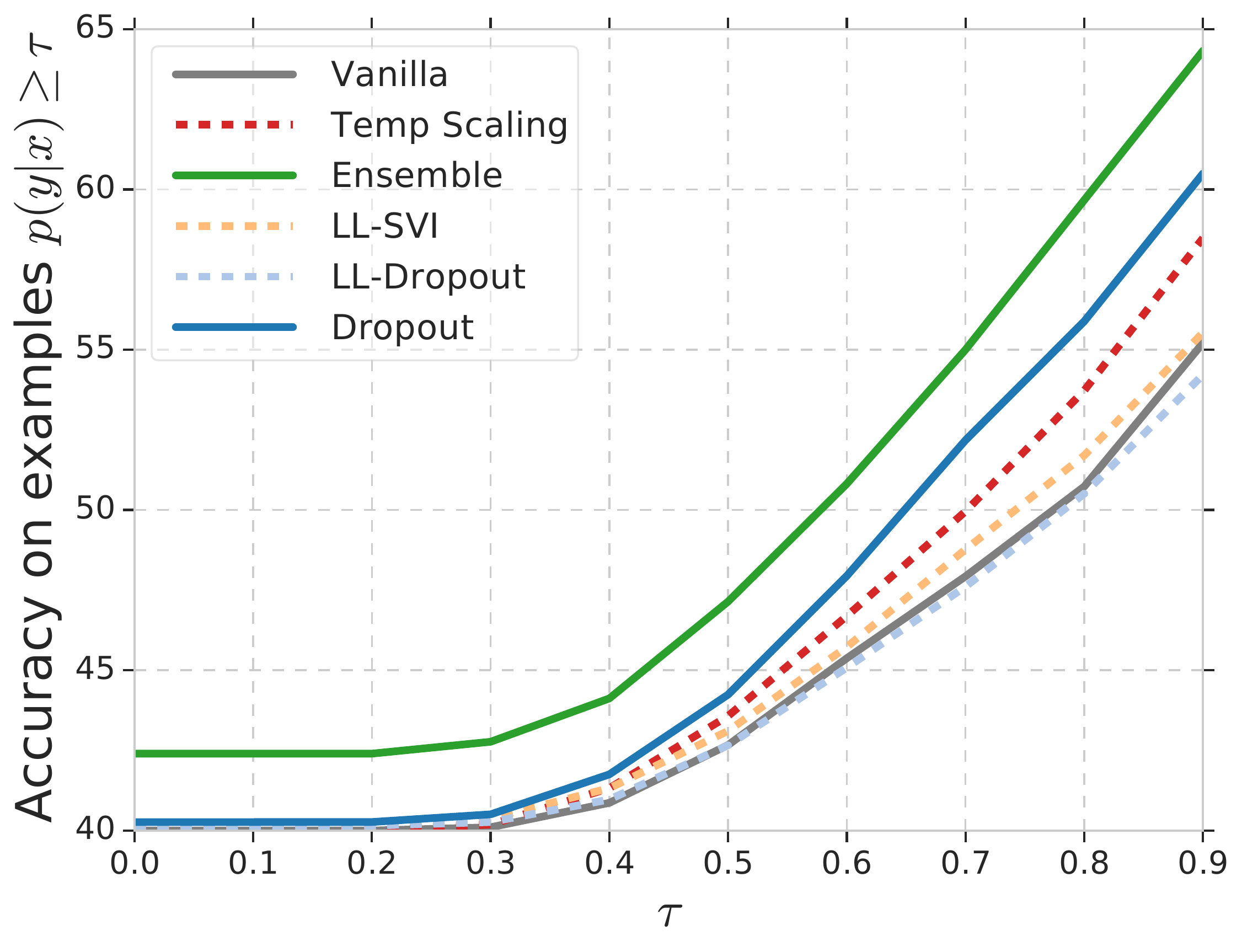}%
    }\end{subfigure}%
    \begin{subfigure}[Confidence vs Count]{
      \includegraphics[width=0.5\linewidth]{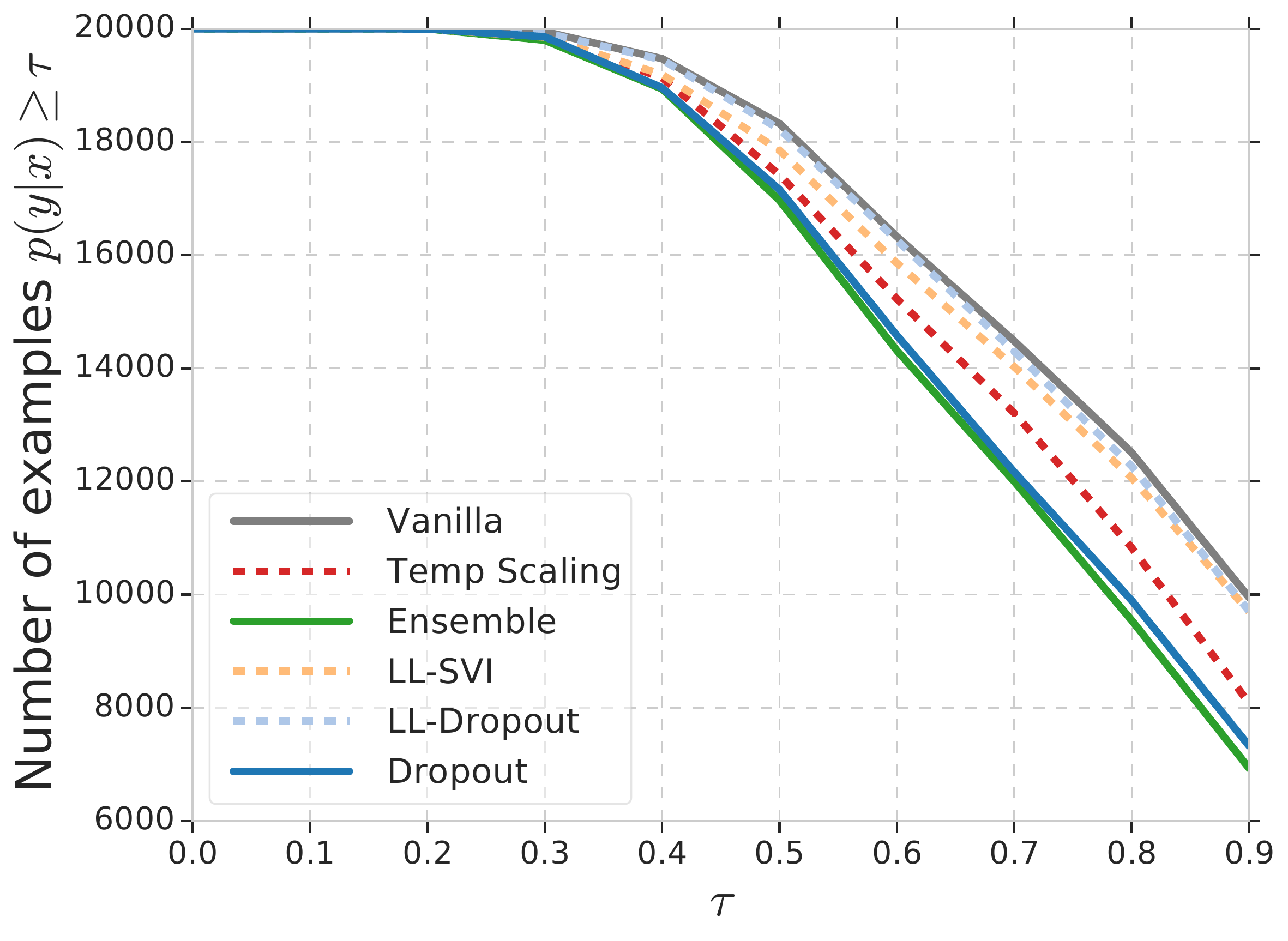}%
    }\end{subfigure}%
    \myvspace{-0.5em}%
    \caption{Confidence score vs accuracy and count respectively when evaluated for in-distribution and OOD genomic sequences.}%
    \myvspace{-0.5em}%
    \label{fig:gene_conf}%
\end{figure}

Figure \ref{fig:gene_conf} shows the confidence vs (a) accuracy and (b) count when the test data consists of a mix of in-distribution and OOD data. Ensembles significantly outperform all other methods, and achieve better trade-off between accuracy versus confidence. Dropout performs better than Temp Scaling, and they both perform better than LL-Dropout, LL-SVI, and the Vanilla method. Note that the accuracy on examples $p(y|\vx)\geq0.9$ for the best method is still below 65\%, suggesting 
that this realistic genomic sequences dataset is a  challenging problem to benchmark future methods.

 \myvspace{-1em}%

\section{Tables of Metrics}
\label{sec:tables}
The tables below report quartiles of Brier score, negative log-likelihood, and ECE for each model and dataset where quartiles are computed over all corrupted variants of the dataset.

\subsection{CIFAR-10}

\begin{tabular}{l|rrrrrrr}
Method & Vanilla & Temp. Scaling & Ensembles & Dropout & LL-Dropout & SVI & LL-SVI \\
\midrule
Brier Score (25th) & 0.243 & 0.227 & 0.165 & 0.215 & 0.259 & 0.250 & 0.246 \\
Brier Score (50th) & 0.425 & 0.392 & 0.299 & 0.349 & 0.416 & 0.363 & 0.431 \\
Brier Score (75th) & 0.747 & 0.670 & 0.572 & 0.633 & 0.728 & 0.604 & 0.732 \\
\midrule
NLL (25th) & 2.356 & 1.685 & 1.543 & 1.684 & 2.275 & 1.628 & 2.352 \\
NLL (50th) & 1.120 & 0.871 & 0.653 & 0.771 & 1.086 & 0.823 & 1.158 \\
NLL (75th) & 0.578 & 0.473 & 0.342 & 0.446 & 0.626 & 0.533 & 0.591 \\
\midrule
ECE (25th) & 0.057 & 0.022 & 0.031 & 0.021 & 0.069 & 0.029 & 0.058 \\
ECE (50th) & 0.127 & 0.049 & 0.037 & 0.034 & 0.136 & 0.064 & 0.135 \\
ECE (75th) & 0.288 & 0.180 & 0.110 & 0.174 & 0.292 & 0.187 & 0.275 \\
\end{tabular}

\subsection{ImageNet}
\begin{tabular}{l|rrrrrr}
Method & Vanilla & Temp. Scaling & Ensembles & Dropout & LL-Dropout & LL-SVI \\
\midrule
Brier Score (25th) & 0.553 & 0.551 & 0.503 & 0.577 & 0.550 & 0.590 \\
Brier Score (50th) & 0.733 & 0.726 & 0.667 & 0.754 & 0.723 & 0.766 \\
Brier Score (75th) & 0.914 & 0.899 & 0.835 & 0.922 & 0.896 & 0.938 \\
\midrule
NLL (25th) & 1.859 & 1.848 & 1.621 & 1.957 & 1.830 & 2.218 \\
NLL (50th) & 2.912 & 2.837 & 2.446 & 3.046 & 2.858 & 3.504 \\
NLL (75th) & 4.305 & 4.186 & 3.661 & 4.567 & 4.208 & 5.199 \\
\midrule
ECE (25th) & 0.057 & 0.031 & 0.022 & 0.017 & 0.034 & 0.065 \\
ECE (50th) & 0.102 & 0.072 & 0.032 & 0.043 & 0.071 & 0.106 \\
ECE (75th) & 0.164 & 0.129 & 0.053 & 0.109 & 0.123 & 0.148 \\
\end{tabular}

\subsection{Criteo}
\begin{tabular}{l|rrrrrrr}
Method & Vanilla & Temp. Scaling & Ensembles & Dropout & LL-Dropout & SVI & LL-SVI \\
\midrule
Brier Score (25th) & 0.353 & 0.355 & 0.336 & 0.350 & 0.353 & 0.512 & 0.361 \\
Brier Score (50th) & 0.385 & 0.391 & 0.366 & 0.373 & 0.379 & 0.512 & 0.396 \\
Brier Score (75th) & 0.409 & 0.416 & 0.395 & 0.393 & 0.403 & 0.512 & 0.421 \\
\midrule
NLL (25th) & 0.581 & 0.594 & 0.508 & 0.532 & 0.542 & 7.479 & 0.554 \\
NLL (50th) & 0.788 & 0.829 & 0.552 & 0.577 & 0.600 & 7.479 & 0.633 \\
NLL (75th) & 0.986 & 1.047 & 0.608 & 0.624 & 0.664 & 7.479 & 0.711 \\
\midrule
ECE (25th) & 0.041 & 0.055 & 0.044 & 0.043 & 0.052 & 0.254 & 0.066 \\
ECE (50th) & 0.097 & 0.113 & 0.100 & 0.085 & 0.100 & 0.254 & 0.127 \\
ECE (75th) & 0.135 & 0.149 & 0.141 & 0.116 & 0.136 & 0.254 & 0.162 \\
\end{tabular}

\end{document}